\documentclass{article} 
\PassOptionsToPackage{numbers, compress}{natbib}

\usepackage{iclr2025_conference,times}

\usepackage{amsmath,amsfonts,bm}

\def\eqref#1{equation~\ref{#1}}

\def\1{\bm{1}}

\DeclareMathAlphabet{\mathsfit}{\encodingdefault}{\sfdefault}{m}{sl}
\SetMathAlphabet{\mathsfit}{bold}{\encodingdefault}{\sfdefault}{bx}{n}

\usepackage{hyperref}
\usepackage{url}
\usepackage[utf8]{inputenc} 
\usepackage[T1]{fontenc}    
\usepackage{hyperref}       
\usepackage{url}            
\usepackage{booktabs}       
\usepackage{amsfonts}       
\usepackage{nicefrac}       
\usepackage{microtype}      
\usepackage{xcolor}         
\usepackage{amsmath}
\usepackage{amssymb}
\usepackage{amsthm}
\usepackage{algorithm}
\usepackage{algpseudocode}
\usepackage{float}
\usepackage{multirow}
\usepackage{graphicx}
\usepackage{booktabs}
\usepackage{tabularx}
\usepackage{multirow}
\usepackage{xspace} 
\usepackage{multicol}
\usepackage{textcomp}
\usepackage{colortbl}  
\usepackage{xcolor}  
\usepackage{array}
\usepackage{subcaption}
\usepackage{xcolor,colortbl}
\usepackage{authblk}

\newcolumntype{C}[1]{>{\centering\arraybackslash}p{#1}}

\newcommand{\graytext}[1]{\textcolor{gray}{ #1}}

\title{Mitigating Spurious Negative Pairs for Robust Industrial Anomaly Detection}

\author[1]{Hossein Mirzaei}
\author[2]{Mojtaba Nafez}
\author[2]{Jafar Habibi}
\author[3]{Mohammad Sabokrou}
\author[2]{Mohammad Hossein Rohban}

\affil[1]{École Polytechnique Fédérale de Lausanne (EPFL), Switzerland}
\affil[2]{Sharif University of Technology, Iran}
\affil[3]{Okinawa Institute of Science and Technology, Japan}

\affil[ ]{\texttt{hossein.mirzaeisadeghlou@epfl.ch}} 
\affil[ ]{\texttt{\{mojtaba.nafez77, jhabibi, rohban\}@sharif.edu}}
\affil[ ]{\texttt{mohammad.sabokrou@oist.jp}}

\iclrfinalcopy 
\begin{document}

\maketitle

\maketitle
\begin{abstract}

Despite significant progress in Anomaly Detection (AD), the robustness of existing detection methods against adversarial attacks remains a challenge, compromising their reliability in critical real-world applications such as autonomous driving. This issue primarily arises from the AD setup, which assumes that training data is limited to a group of unlabeled normal samples, making the detectors vulnerable to adversarial anomaly samples during testing. Additionally, implementing adversarial training as a safeguard encounters difficulties, such as formulating an effective objective function without access to labels. An ideal objective function for adversarial training in AD should promote strong perturbations both within and between the normal and anomaly groups to maximize margin between normal and anomaly distribution. To address these issues, we first propose crafting a pseudo-anomaly group derived from normal group samples. Then, we demonstrate that adversarial training with contrastive loss could serve as an ideal objective function, as it creates both inter- and intra-group perturbations. However, we notice that spurious negative pairs compromise the conventional contrastive loss to achieve robust AD. Spurious negative pairs are those that should be closely mapped but are erroneously separated. These pairs introduce noise and misguide the direction of inter-group adversarial perturbations. To overcome the effect of spurious negative pairs, we define opposite pairs and adversarially pull them apart to strengthen inter-group perturbations. Experimental results demonstrate our superior performance in both clean and adversarial scenarios, with a \textbf{26.1\%} improvement in robust detection across various challenging benchmark datasets. The implementation of our work is available at: \url{https://github.com/rohban-lab/COBRA}.

\end{abstract}
\section{Introduction}\label{sec:intro}

Anomaly detection (AD), also referred to as one-class classification, aims to identify whether an input sample at the time of inference belongs to the normal\footnote{In this study, the term ‘\textit{normal distribution}’ refers to inlier samples, not to Gaussian distribution.} or anomaly group.  In AD setup, the training data consists only of normal samples, and any additional information, such as labels, is unavailable \cite{bendale2015towards,perera2021one}. Recently, a plethora of literature has emerged to address the problem of AD on images, demonstrating near-perfect performance on standard AD benchmarks \cite{ruff2018deep, tack2020csi, bergman2020deep, reiss2021panda, bergmann2019mvtec, krizhevsky2009learning}. Nevertheless, these methods demonstrate a lack of robustness, especially when faced with adversarial attacks, as they encounter substantial performance deterioration when faced with such scenarios \cite{azizmalayeri2022your,lo2022adversarially,chen2020robust,shao2020open,shao2022open,bethune2023robust,goodge2021robustness,chen2021atom}. This is due to the absence of anomaly samples in the training data, which results in insufficient exposure to adversarial perturbations on anomalous patterns during training. This shortcoming would make the model vulnerable to adversarial attacks on anomaly samples during inference \cite{chen2020robust,chen2021atom}.

Numerous defense strategies have been developed to enhance the robustness of deep neural networks, with adversarial training emerging as a potential solution \cite{bai2021recent,madry2017towards}. However, its application to AD is not straightforward, as it is primarily developed for multi-class and labeled setups.

Motivated by the aforementioned issues, we propose generating pseudo-anomaly group samples by applying hard augmentations to facilitate practical adversarial training in an anomaly detection (AD) setup. This process involves shifting normal training data to ensure that the shifted samples do not belong to the normal group, measured by their likelihood using a novel thresholding approach \cite{glodek2013ensemble}. We refer to the relationship between a normal sample and its transformed version as \textit{opposite pairs}.

Given the availability of two groups—crafted anomaly and normal samples—during training, defining a loss function to incorporate them into the adversarial training presents a new challenge. Since the objective of test-time adversarial attacks is to manipulate normal samples to be confused with anomalies and vice versa, the optimal objective function should maximize the margin between the distributions of normal and anomaly samples in the learned embedding space while also achieving compact representations for each group. This can be adversarially accomplished by crafting strong intra- and inter-group perturbations \cite{chen2021large,cheng2023ml,guo2021recent}.

It has been demonstrated that Contrastive Learning (CL) \cite{chen2020simple, he2020momentum} is more effective for AD compared to existing objective functions \cite{guo2024recontrast, reiss2021mean, tack2020csi}. One can propose adversarial training with CL to develop a robust AD method. However, we noticed that adversarial training with the CL loss function falls short of achieving robust AD (see Table \ref{tab:loss_function_ablaiton}). The CL objective aims to bring positive pairs closer together and push negative pairs further apart. Positive pairs are constructed by applying light transformations to each instance, while any two instances in the training data are treated as negative pairs. We refer to negative pairs within the same group (normal-normal or anomaly-anomaly) as \textit{spurious negative pairs}. These spurious negative pairs weaken the effectiveness of adversarial CL by misdirecting inter-group perturbations, thereby reduces the margin between groups \cite{chen2021large}. 
 
 To address this, we propose COBRA (anomaly-aware \textbf{CO}ntrastive-\textbf{B}ased approach for \textbf{R}obust \textbf{A}D), a new method that mitigates the effect of spurious negative pairs to learn effective perturbations. COBRA strategically utilizes opposite pairs, exclusively formed between normal and anomaly groups, ensuring they do not intersect with spurious negatives. This approach strengthens inter-group perturbations by emphasizing these opposite pairs in the loss function, thereby increasing the margin between groups. During training, the model adversarially targets positive pairs to push them together and opposite pairs to pull them apart. This simulates a wide range of adversarial perturbations covering inter- and intra-set variations, resulting in a robust anomaly detector.

\textbf{Contribution.} COBRA introduces a simple yet effective approach to generate anomaly samples and a novel loss function to establish a robust detection boundary. We evaluate COBRA in both adversarial and clean settings, where test samples are benign. In the adversarial scenario, we employ numerous strong attacks for robustness evaluation, including PGD-1000 \cite{madry2017towards}, AutoAttack \cite{croce2020reliable}, and Adaptive AutoAttack \cite{liu2022practical}. The results show that COBRA, without using any additional datasets or pretrained models, significantly outperforms existing methods in adversarial settings, achieving a 26.1\% improvement in AUROC and competitive results in standard settings. Our experiments span various datasets, including large and real-world datasets such as Autonomous Driving \cite{Cityscapes}, ImageNet \cite{deng2009imagenet}, MVTecAD \cite{bergmann2019mvtec}, and ISIC \cite{codella2019skin}, demonstrating COBRA's practical applicability. Additionally, we conducted ablation studies to examine the impact of various COBRA components, specifically our pseudo-anomaly generation strategy and the introduced adversarial training method.

\section{  Preliminaries}

\textbf{Anomaly Detection.} Outlier detection is categorized into different areas, such as AD and Out-of-Distribution (OOD) detection, depending on the availability of normal set samples' labels. An AD method decides whether $x$ belongs to the normal or anomaly set by assigning an anomaly score $A(x;f)$ using model f. Samples with an anomaly score higher than a pre-assumed threshold are predicted as anomalies, and vice versa \cite{yang2021generalized,ruff2021unifying}. It is important to note that extending OOD detection methods to an AD setup is not feasible, as they rely on labeled normal data for feature extraction. This highlights the need for robust AD methods.\cite{azizmalayeri2022your,chen2021atom,kong2021opengan,han2022adbench}.

\noindent\textbf{Adversarial Robustness of Anomaly Detectors.} \ \ An adversarial attack is a malicious attempt to perturb a data sample \(x\) with an associated label \(y\) into a new sample \(x^*\) that maximizes the loss function \(\ell(x^*; y)\). Additionally, an upper limit of \(\epsilon\) confines the \(l_p\) norm of the adversarial noise to prevent semantic alterations. Specifically, an adversarial example \(x^*\) must satisfy the following equations: 
$x^* = \arg \max_{x^{\prime}: \|x - x^{\prime}\|_p \leq \epsilon} \ell(x^{\prime}; y)$
A prevalent and effective attack method is the Projected Gradient Descent (PGD) technique \cite{madry2017towards}, which entails iteratively maximizing the loss function by advancing towards the gradient sign of \(\ell(x^*; y)\), employing a designated step size \(\alpha\). 
To adapt adversarial attacks for AD, instead of maximizing the loss value, we aim to increase \(A(x, f)\) if \(x\) belongs to normal group and decrease it otherwise. The formulation of the attack would be:
$x_0^*=x, \quad  x_{t+1}^* =x_t^*+ y \cdot\alpha \cdot \operatorname{sign}\left(\nabla_x A(x_t^*;f_\theta)  \right),  \  x^*  = x^{*}_k.$ 
 Here $k$ is the number of attack steps, \(y=+1\) for normal samples and \(y=-1\) for anomaly samples. The same setting is applied to other attacks in our study.

\noindent \textbf{Auxiliary Anomaly Sample Crafting.}\ \
CSI \cite{tack2020csi} and CPAD \cite{li2021cutpaste} propose using fixed hard augmentation to create auxiliary samples. Specifically, CSI relies on Rotation, while CPAD considers CutPaste as a pseudo-anomaly. The GOE \cite{kirchheim2022outlier} method employs a pretrained GAN on ImageNet-1K to craft anomalies by targeting low-density areas. FITYM \cite{mirzaei2022fake} employed an underdeveloped diffusion  as a generator. Dream-OOD \cite{du2023dream} uses both image and text domains to learn visual representations of normal instances in an embedding space of a pretrained stable diffusion \cite{rombach2022high} model trained on 5 billion data (e.g. LAION \cite{schuhmann2022laion}). On the other hand, VOS \cite{du2022vos} generates anomaly embeddings instead of image data. Details about each mentioned method can be found in \ref{Appendix_Details_Related_Work}.

\section{Method}\label{method}

\textbf{Motivation.} \ \ Adversarial training is one of the most promising approaches to enhance the robustness of deep neural networks. However, applying this technique to AD poses a significant challenge, as only a single concept class—the normal distribution—is available during training. A common approach to address this limitation is to incorporate an auxiliary anomaly dataset to improve robustness \cite{azizmalayeri2022your,chen2021atom,chen2020robust,mirzaeirodeo}. However, leveraging such datasets is both costly and challenging, primarily due to the need for preprocessing and filtering out normal samples, which could otherwise provide misleading information to the detector. Moreover, the use of additional anomaly data can bias the model towards specific anomaly samples, reducing its generalizability to unseen anomalies \cite{ming2022poem}.\\

To overcome these limitations, we propose a simple yet effective method to craft anomaly samples directly from the normal data, thus eliminating the need for external anomaly datasets. Our approach involves applying hard augmentations (e.g., severe distortions) to normal samples, effectively pushing them towards the anomaly distribution. Importantly, prior work has demonstrated that the most effective anomalies for training are those that are closely related to the normal distribution, often referred to as "near anomaly samples" \cite{ming2022poem,mirzaei2022fake,chen2021atom}. Our method satisfies this proximity, as the crafted anomalies maintain stylistic similarities with the normal samples due to their generation through augmentations. To ensure that the crafted anomalies are sufficiently distinct from the normal distribution, we introduce a thresholding mechanism to filter out false anomalies. Implementing this mechanism requires a model that accurately captures the distribution of normal data. However, in the AD setup, the training data is limited to a single semantic class (e.g., images of cars) without any supplementary information, posing a challenge for building such a model. To overcome this, we employ a self-supervised approach to extract meaningful representations from the normal data. Inspired by representative studies in AD \cite{golan2018deep,hendrycks2019using,tack2020csi}, which demonstrate that using a $k-$class classifier to predict data transformations is an effective method for representation learning in one-class classification, we adopt this approach. We leverage the embeddings learned by the classifier to compute the likelihood of test samples and define a threshold for filtering out false anomalies.

Subsequently, we explore potential objective functions for adversarial training, focusing on CL given its recent success in AD tasks \cite{guo2024recontrast}.
However, we observe that employing standard CL in adversarial training may not yield optimal results. This stems from the fact that, when training on a dataset containing both normal and crafted anomaly samples, CL forms positive and negative pairs in a way that may compromise the margin between the normal and anomaly distributions. Specifically, CL seeks to uniformly repel negative pairs from each other, defined as all pairs except those that are augmentations of each other \cite{chen2020self}. Consequently, the negative pairs include normal-normal, anomaly-anomaly, and normal-anomaly pairs. Increasing the distance between normal-normal and anomaly-anomaly pairs may inadvertently reduce the separation between normal and anomaly sets, thus undermining robust detection performance. In other words, standard CL does not effectively enhance the inter-set margin needed to improve robustness. To address this, we design a novel objective function that explicitly maximizes the margin between the normal and anomaly groups to improve inter-group perturbation. \\ \textbf{Outline.} \ \ Existing AD methods experience dramatic performance decrease under adversarial attack. To address this, we propose COBRA, a novel method that integrates a distribution-aware transformation for generating psudo-anomaly samples, coupled with a novel objective function for adversarial training. In the subsequent sections, we will delve into each component in detail, outlining the mechanisms and advantages of our approach.

\subsection{Distribution Aware Hard Transformation} \label{Distribution_A} 
\noindent \textbf{Anomaly Crafting Strategy.}\ \ Previous works have demonstrated the effectiveness of leveraging an auxiliary random dataset as an additional source of anomaly data during the training phase for AD \cite{hendrycks2018deep,tao2023nonparametric,du2022vos,du2023dream,zhang2017mixup,mirzaei2022fake,kirchheim2022outlier}. However, this technique significantly depends on the diversity and distribution distance of the auxiliary dataset used for training. This limitation significantly hinders the use of this technique in areas like medical imaging, where real anomalies are scarce and difficult to obtain. Moreover, they lack any threshold for dropping incorrectly crafted anomalies (those that still belong to the normal group).  Addressing this limitation, our approach introduces a novel method that employs a series of hard transformations $\textit{T}=\{T_i\}_{i=1}^{k}$ to generate anomaly samples from normal data. We strategically distort normal images and by using a predetermined threshold ensures that the synthetically created samples significantly diverge from the normal distribution. 
We used a set of hard transformations including  Jigsaw \cite{10.1007/978-3-319-46466-4_5}, Random Erasing \cite{zhong2020random}, CutPaste \cite{ghiasi2020simple}, Rotation, Extreme Blurring, Intense Random Cropping, Noise Injection \cite{Akbiyik2019DataAI}, and Extreme Cropping, Mixup \cite{zhang2018mixup}, Cutout \cite{devries2017cutout}, CutMix \cite{yun2019cutmix}, Elastic transform and etc. Each one has been shown to be harmful for preserving semantics in previous studies \cite{tack2020csi,sohn2020learning,park2020novelty,de2021contrastive,kalantidis2020hard,li2021cutpaste,sinha2021negative,NEURIPS2020_f7cade80,Miyai_2023_WACV,Zhang_2024_WACV,DBLP:conf/ijcai/ChenXLQZTZM21}. For more details, please see Appendix \ref{Appendix_Hard_Transformation}.

\textbf{Threshold Computing.}  First, we train a transformation predictor model that captures the distribution of normal samples through the classification of various augmentations. To achieve this, we create a synthetic dataset with \(k\) classes, denoted as \(\{D_{\text{train}}^{T_i}\}_{i=1}^{k}\), by applying \(k\) different hard transformations \(T_i\) to each of the \(n\) samples in the normal training set \(D_{\text{train}}\). Then, we train a \(k\)-class classifier on this synthetic dataset to leverage its knowledge for threshold computation.  Using the classifier as a feature extractor, denoted as \(C\), we extract embeddings of the normal training samples to create an embedding set:
$ e_{\text{train}} = \left\{ C(D_{\text{train}}^i) \right\}_{i=1}^{n}.$ Next, we fit a Gaussian Mixture Model (GMM) to the training data embeddings \(e_{\text{train}}\), as this is a well-established approach in the literature \cite{cohen2021transformaly,du2022vos}. The likelihood for each sample is computed, and the \textbf{p-value} for test samples is calculated based on the empirical distribution of likelihoods from the normal training samples. The threshold \(\lambda\) is set at a default significance level of 0.05, such that samples with p-values below this threshold are considered anomalies. An ablation study on the significance level, as well as an analysis of \(C\), are provided in Appendix \ref{Appendix_Distribution_Aware_Hard_Transformation} and Appendix \ref{Appendix_Additional_Ablation}, respectively.

\noindent \textbf{Opposite Pairs with Pseudo-Anomaly Samples.}\ \ \label{Opposite Pairs with Anomalies} For each normal sample, we randomly select a subset of transformations \(T\), containing at least two transformation. These transformations are applied in a randomized sequence to the sample, producing \(x' = T_{i_m}(\ldots T_{i_1}(x))\) where \(m < k\). We then get its embedding and calculate the likelihood \(pr(C(x'))\). Finally, this likelihood is compared against the computed threshold \(\lambda\). Samples exceeding this threshold iteratively repeat this process until deemed an anomaly. We represent our proposed strategy for anomaly crafting with the notation \(\Upsilon(x)\). Before each step of training, given a batch of normal samples denoted by \(\mathcal{B}_{\text{normal}} = \{x^j\}_{j=1}^{b}\), we create a batch of anomaly samples \(\mathcal{B}_{\text{p-anomaly}} = \{x^j\}_{j=b+1}^{2b}\). Where \(x^{b+i} = \Upsilon(x^i)\) signifies an anomaly sample derived from applying a transformation \(\Upsilon\) to a normal sample \(x^i\), and \((x^i, x^{b+i})\) are considered as symmetrically opposite pairs. During training, the notation \(\Upsilon(x)\) has some minor differences, where \(\Upsilon(x^{b+i})\) is considered as \(x^{i}\).

 \begin{figure}[t]
  \centering
  \includegraphics[width=1.0\linewidth]{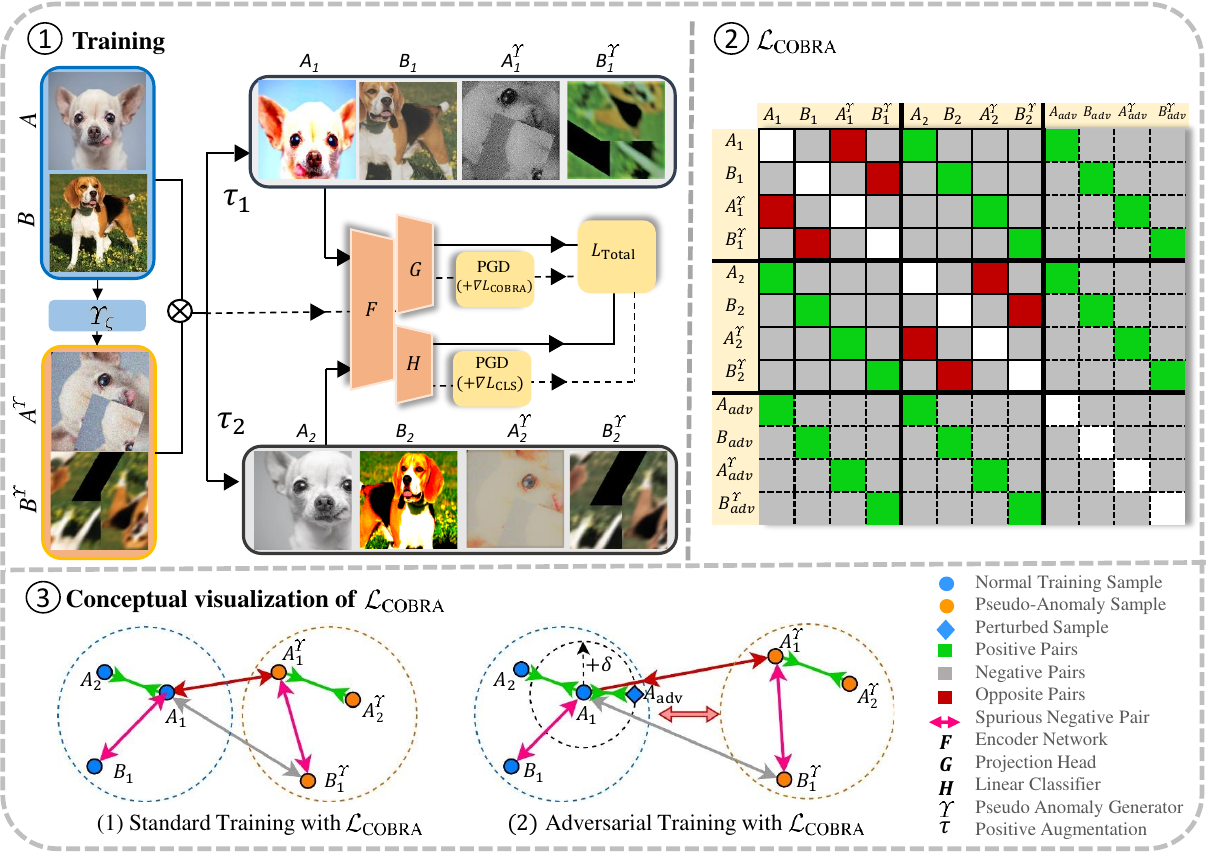}
  \caption{\textbf{\textcircled{1}}: Given a training batch that includes normal group samples \(A\) and \(B\), we create a anomaly group using our proposed transformation \( {\Upsilon}_{\lambda}\). These samples are paired as opposite pairs (e.g., \(A\) and \(A^{\Upsilon}\)) and subjected to \(\tau_1\) and \(\tau_2\) to form batches of positive pairs. Adversarial training is performed with a loss function combining \(L_{\text{CLS}}\) and \(L_{\text{COBRA}}\), where \(L_{\text{COBRA}}\) treats adversarial examples as positive pairs for the corresponding sample.  
  \textbf{\textcircled{2}, \textcircled{3}}: The illustrations demonstrate how \(L_{\text{COBRA}}\) enhances adversarial training by explicitly increasing similarities within positive pairs and decreasing similarity for opposite pairs, thus creating strong inter- and intra-group perturbations. Targeting opposite pairs instead of all negatives diminishes the effect of spurious negative pairs (e.g., \((A_1, B_1)\)), leading to stronger inter-group perturbations and enlarging the margin between distributions for normal and anomaly groups.  A detailed algorithmic of COBRA is provided in \ref{appendix_algorithm_model}.
 }
  \label{fig:Pipline_Figure}
\end{figure}
\subsection{Adversarial Training with Anomaly-Aware CL} \label{Contrastive Based Adversarial Training}
\textbf{Conventional Contrastive Loss.}\ \ In the conventional CL paradigm, each instance \(x\) within a batch transforms into two positive views, \((x_1, x_2)\), via a random selection of positive augmentations \(\tau_1, \tau_2\) from a predefined set \(\mathcal{T}\). The set of positive pairs corresponding to a sample \(x\) is denoted as \(\text{P}(x_1) = \{x_2\}\) and \(\text{P}(x_2) = \{x_1\}\). CL then defines negative pairs \(N(x)\) for sample \(x\) as the other samples' augmented views. By denoting the current batch as \(B\), we achieve: 
$N(x) = \left\{ \tau_1(x') : x' \in B \setminus \{x\} \right\} \cup \left\{ \tau_2(x') : x' \in B \setminus \{x\} \right\}$.
These views are processed through a target network to obtain projected features, symbolized as \(z = \mathcal{G}(\mathcal{F}(x))\), where \(\mathcal{F}\) signifies the feature encoder and \(\mathcal{G}\) the projection head. For simplification, \(f(.)\) substitutes \(\mathcal{G}(\mathcal{F}(.))\). The conventional NT-Xent \cite{chen2020simple} loss is articulated as:
\begin{align}
\mathcal{L}_{\text{CL}}(x) = - \sum_{i=1}^{2} \sum_{x_j \in \text{P}(x_i)}\log \frac{\exp(\text{sim}(f(x_i), f(x_j))/t)}{\sum\limits_{x_k \in \text{P}(x_i) \cup \text{N}(x)} \exp(\text{sim}(f(x_i), f(x_k))/t)}.
\end{align}
where $\text{sim}(\cdot, \cdot)$ denotes the cosine similarity function, $t$ is the temperature parameter, $\text{P}(x_i)$ is the set of positive pairs for $x_i$, and $\text{N}(x_i)$ is the set of negative pairs for $x_i$.

\noindent \textbf{Addressing Spurious Negative Pairs with $\mathcal{L}_{\text{COBRA}}$.} \ CL aims to pull positive pairs closer to each other and push negative pairs away from each other. Consider defining the current batch of samples for CL as the concatenation of two groups: normal and anomaly samples, \(\mathcal{B} = \{\mathcal{B}_{\text{normal}} \cup \mathcal{B}_{\text{p-anomaly}}\}\). Due to the definition of negative pairs in CL, each sample in \(\mathcal{B}\) includes both inter-group and intra-group relations as negative pairs. Intra-group negative pairs, i.e., normal-normal and anomaly-anomaly pairs, are considered spurious negative pairs. Distancing spurious negative pairs is counterproductive to our objective, which is to maximize the discriminative margin between normal and anomaly groups. Specifically, in the scenario of adversarial training for robust AD with \(\mathcal{L}_{\text{CL}}\), spurious negative pairs misdirect inter-group adversarial perturbation. As a result, we aim to precisely target those negative pairs that definitively belong to separate groups—what we refer to as opposite pairs. By focusing on these pairs, we aim to induce stronger perturbations that significantly enhance the discriminative margin between the normal and anomaly groups by proposing $\mathcal{L}_{\text{COBRA}}$.
\begin{align}
  \mathcal{L}_{\text{COBRA}}(x) = -\sum_{i=1}^{2} \sum_{x_j \in \text{P}(x_i)} \log \frac{\exp(\text{sim}(f(x_i), f(x_j))/t) - \exp(\text{sim}(f(x_i), f(\Upsilon(x)))/t)}{\sum\limits_{x_k \in \text{P}(x_i) \cup \text{N}(x)} \exp(\text{sim}(f(x_i), f(x_k))/t)},
\end{align} 

Note that $\Upsilon(x)$ could also be replaced by $\Upsilon(x_i)$, as both $x$ and $x_i$ are positive pairs and share similar semantics. Applying a hard transformation to either would result in a comparable hard transformation. The intuition behind $\mathcal{L}_{\text{COBRA}}$ is that the representations of the corresponding positive views $x_1$ and $x_2$ should be similar (analogous to the $\mathcal{L}_{\text{CL}}$ loss function), leading to compact representations for each group. Meanwhile, the representation of $x$ should be distinctly different from its counterpart representation $\Upsilon(x)$, resulting in a high margin between the two groups.
A conceptual visualization of $\mathcal{L}_{\text{COBRA}}$ is provided in Figure \ref{fig:Pipline_Figure}. It is important to highlight that the limitations of CL in AD are apparent in adversarial scenarios. This stems from the fact that adversarial training requires a higher degree of data complexity \cite{schmidt2018adversarially,stutz2019disentangling} compared to clean settings, necessitating a broad range of strong perturbations to achieve robust anomaly detection. One can say \(\mathcal{L}_{\text{COBRA}}\) consists of two terms, \(\mathcal{L}_{\text{CL}}\) and \(\mathcal{L}_{\text{Opposite}}\), where\begin{align}
\mathcal{L}_{\text{Opposite}} (x) = - \sum_{i=1}^{m} \sum_{x_j \in \text{P}(x_i)}\log \frac{- \exp(\text{sim}(f(x_i), f(\Upsilon(x))/t)}{\sum\limits_{x_k \in \text{P}(x_i) \cup \text{N}(x_i)} \exp(\text{sim}(f(x_i), f(x_k))/t)},
\end{align}To enhance our model's ability to distinguish between normal and anomaly groups, we employ a fully connected layer followed by softmax activation for binary classification, denoted as $\mathcal{H}$, following the $\mathcal{F}$. For the classification loss, $\mathcal{L}_{\text{CLS}}(x)$, we define a label $y$ corresponding to the batch $\mathcal{B}$, where ‘0’ is assigned as the label for normal samples and ‘1’ for pseudo anomaly samples. Our final loss function, $\mathcal{L}_{\text{COBRA}}$, is thus formulated as:
$\quad
\mathcal{L} = \sum_{i=1}^{2b} \mathcal{L}_{\text{COBRA}}(\mathcal{F}, \mathcal{G}; \mathcal{B}^i ) + \mathcal{L}_{\text{CLS}}(\mathcal{F}, \mathcal{H}; \mathcal{B}^i , y^i)
$

\noindent \textbf{Adversarial Training Step.} \ \  For $\mathcal{L}_{\text{COBRA}}$, given an input sample $x$ from the batch $\mathcal{B}$, an adversarial example $x_{adv}$ is generated by introducing a perturbation $\delta^*$, optimized to maximize our final loss: $ \delta^* = \underset{\|\delta\|_{\infty} \leq \epsilon}{\mathrm{argmax}}\, \mathcal{L}(x + \delta, y), \quad x_{adv} = x + \delta^*
$. Then, adversarial examples are used in the  training process alongside the original examples. Specifically, for $\mathcal{L}_{\text{COBRA}}$, we consider them as another positive view of each sample and aim to align each sample with its perturbed version, i.e., $P(x_i) \leftarrow P(x_i) \cup \{ x_{adv}\}$. The adversarial training objective as a min-max problem, optimizing the model parameters $\theta$ to minimize the expected loss over both clean and adversarial examples:
$$
\min_{\theta} \mathbb{E}_{(x, y) \in \mathcal{B} } \left[ \max_{\|\delta\|_\infty \leq \epsilon} \mathcal{L}(x + \delta, y; \theta) \right].
$$
Morever, the stability of the $ \mathcal{L}_{\text{COBRA}}$ can be observed in both clean and adversarial training scenarios, as illustrated in the Appendix \ref{Appendix_COBRA_Loss_Stability}.

\noindent\textbf{Anomaly Score for Evaluation.}\ \ For evaluating anomalies, we leverage the representation learned by $\mathcal{F}$ to compute the anomaly score, based on the similarity between test samples and normal training samples in the embedding space. The anomaly score $A(X)$ for a test sample $x$ is defined as: 
$ - \max_{x^i \in D_{\text{train}}} \left\{ sim (f(\mathbf{x}), f(\mathbf{x}^i)  \right\},
$ This scoring mechanism takes advantage of the contrastive training framework, ensuring that normal test samples exhibit higher similarity scores in comparison to anomaly test samples. Consequently, the anomaly score for anomaly test samples will be notably higher than for normal test samples, enabling robust AD. Alternative anomaly scores have been explored in the appendix \ref{Appendix_Additional_Ablation}.

\begin{table*}[t] 
    \caption{
    Performance of AD methods on MVTecAD dataset under clean evaluation and PGD-1000 adversarial attack with \(\epsilon=\frac{2}{255}\), measured by AUROC (\%). The best results are emphasized in bold format in each row. The table cells denote results in the ‘\graytext{Clean 
 /} PGD-1000’ format. 
}
    \footnotesize{$^*$These works incorporated adversarial training into their proposed AD methods.}

    \resizebox{ \textwidth}{!}{\begin{tabular}{l*{8}{c}>{\columncolor[gray]{0.95}}c}

    \specialrule{1.5pt}{\aboverulesep}{\belowrulesep}
    \multicolumn{1}{c}{Category} & &\multicolumn{7}{c}{Method }  
    \\ \cmidrule(lr){1-1} \cmidrule(lr){2-10}  


    &\multirow{2}{*}{CSI}&\multirow{2}{*}{\  Transformaly}&\multirow{2}{*}{PatchCore} &\multirow{2}{*}{ReContrast} & \multirow{2}{*}{DRÆM} & \multirow{2}{*}{PrincipaLS$^*$}&\multirow{2}{*}{OCSDF$^*$}&\multirow{2}{*}{ZARND$^*$}& \cellcolor{gray!0} COBRA    \\
     & & & & &  & &  &  & \cellcolor{gray!0} (Ours)  \\
    \specialrule{1.5pt}{\aboverulesep}{\belowrulesep}

   \multirow{1}{*}{Carpet} 
     & \textcolor{gray}{\  50.2 /} 11.1 & \textcolor{gray}{\  95.5 /} 0.0 & \textcolor{gray}{\  98.7 /} 18.4 & \textcolor{gray}{\  99.8 /} 9.4 & \textcolor{gray}{\  97.0 }/ 0.0 & \textcolor{gray}{\  54.8 /} 33.6 & \textcolor{gray}{\  56.1 /} 12.6 &  \textcolor{gray}{\  \textbf{85.9} /} 66.6 & \textcolor{gray}{\  60.7 /} \textbf{84.9} \\
    
    \cmidrule(lr){1-1}\cmidrule(lr){2-10}

    \multirow{1}{*}{Grid} 
   & \textcolor{gray}{\  71.2 /} 8.3 & \textcolor{gray}{\  84.2 /} 7.8 & \textcolor{gray}{\  98.2 /} 11.7 & \textcolor{gray}{\  100.0 /} 19.8 & \textcolor{gray}{\  99.9 /} 2.7 &\textcolor{gray}{\  72.1 /} 30.4 & \textcolor{gray}{\  61.7 /} 17.3 & \textcolor{gray}{\  75.7 /} 31.1 & \textcolor{gray}{\  \textbf{100.0} /} \textbf{99.5} \\
   
    \cmidrule(lr){1-1}\cmidrule(lr){2-10}

    \multirow{1}{*}{Leather} 
    & \textcolor{gray}{\  70.9 /} 0.4 & \textcolor{gray}{\  99.9 /} 4.1 & \textcolor{gray}{\  100.0 /} 10.5 & \textcolor{gray}{\  100.0 /} 3.4 & \textcolor{gray}{\  \textbf{100.0} /} 0.0 & \textcolor{gray}{\  73.2 /} 26.5 & \textcolor{gray}{\  61.4 /} 13.7 & \textcolor{gray}{\  65.0 /} 14.1 & \textcolor{gray}{\  97.4 /} \textbf{91.7} \\
   
    \cmidrule(lr){1-1}\cmidrule(lr){2-10}

    \multirow{1}{*}{Tile} 
   & \textcolor{gray}{\  67.8 /} 7.2 & \textcolor{gray}{\  97.1 /} 2.0 & \textcolor{gray}{\  98.7 /} 4.6 & \textcolor{gray}{\  99.8 /} 2.4 & \textcolor{gray}{\  99.6 /} 0.0 & \textcolor{gray}{\  58.7 /} 26.3 &\textcolor{gray}{\  54.3 / } 10.1 & \textcolor{gray}{\  \textbf{53.6} /} 3.9 & \textcolor{gray}{\  98.8 /} \textbf{78.2} \\
   
    \cmidrule(lr){1-1}\cmidrule(lr){2-10}

    \multirow{1}{*}{Wood} 
    & \textcolor{gray}{\  71.3 /} 6.0 & \textcolor{gray}{\  98.5 /} 0.0 & \textcolor{gray}{\  99.2 /} 3.8 & \textcolor{gray}{\  99.0 /} 1.6 & \textcolor{gray}{\  99.1 /} 1.8 & \textcolor{gray}{\  67.3 /} 31.2 & \textcolor{gray}{\  63.9 /} 3.7 & \textcolor{gray}{\  \textbf{58.4} /} 17.5 & \textcolor{gray}{\  96.4 /} \textbf{73.7} \\
   
    \cmidrule(lr){1-1}\cmidrule(lr){2-10}

    \multirow{1}{*}{Bottle} 
      & \textcolor{gray}{\  69.4 /} 1.2 & \textcolor{gray}{\  99.4 /} 5.1 & \textcolor{gray}{\  100.0} / 9.4 & \textcolor{gray}{\  100.0 /} 6.8 & \textcolor{gray}{\  99.2 /} 2.1& \textcolor{gray}{\  72.1/} 29.4 & \textcolor{gray}{\  59.8 /} 9.1 & \textcolor{gray}{\  79.9/} 54.9 & \textcolor{gray}{\  \textbf{100.0} /} \textbf{88.8} \\

    \cmidrule(lr){1-1}\cmidrule(lr){2-10}

    \multirow{1}{*}{Cable} 
      & \textcolor{gray}{\  66.5 /} 7.9 & \textcolor{gray}{\  81.5 /} 0.1 & \textcolor{gray}{\  99.5 /} 4.3 & \textcolor{gray}{\  99.8 /} 3.3 & \textcolor{gray}{\  91.8 /} 1.9 & \textcolor{gray}{\  63.9 /} 26.2 & \textcolor{gray}{\  61.6 /} 6.3 & \textcolor{gray}{\  \textbf{68.5} /} 30.0 & \textcolor{gray}{\  92.4 /} \textbf{74.8} \\

    \cmidrule(lr){1-1}\cmidrule(lr){2-10}

    \multirow{1}{*}{Capsule} 
      & \textcolor{gray}{\  51.6 /} 6.8 & \textcolor{gray}{\  76.0 /} 0.0 & \textcolor{gray}{\  98.1 /} 3.1 & \textcolor{gray}{\  97.7 /} 2.7 & \textcolor{gray}{\  98.5 /} 0.0 & \textcolor{gray}{\  56.8 /} 18.4 & \textcolor{gray}{\  51.9 /} 1.9 & \textcolor{gray}{\  \textbf{69.3} /} 26.2 & \textcolor{gray}{\  75.9 /} \textbf{55.8} \\

    \cmidrule(lr){1-1}\cmidrule(lr){2-10}

    \multirow{1}{*}{HazelNut} 
       & \textcolor{gray}{\  66.7 /} 0.0 & \textcolor{gray}{\  89.8 /} 0.4 & \textcolor{gray}{\  100.0 /} 7.8 & \textcolor{gray}{\  100.0 /} 4.1 & \textcolor{gray}{\  \textbf{100.0} /} 0.8 & \textcolor{gray}{\  64.8 /} 21.7  &\textcolor{gray}{\  54.2 /} 4.7 & \textcolor{gray}{\  73.2 /}24.0 & \textcolor{gray}{\  96.3 /} \textbf{74.7} \\

    \cmidrule(lr){1-1}\cmidrule(lr){2-10}

    \multirow{1}{*}{MetalNut} 
        & \textcolor{gray}{\  65.8 /} 0.7 & \textcolor{gray}{\  90.9 /} 6.2 & \textcolor{gray}{\  100.0 /} 4.8 & \textcolor{gray}{\  \textbf{100.0} /} 3.7 & \textcolor{gray}{\  98.7 /} 0.6 & \textcolor{gray}{\  61.6 /} 19.4 & \textcolor{gray}{\  59.5 /} 3.5 & \textcolor{gray}{\  43.1 /} 1.3 & \textcolor{gray}{\  96.8 /} \textbf{78.1} \\

    \cmidrule(lr){1-1}\cmidrule(lr){2-10}

    \multirow{1}{*}{Pill} 
       & \textcolor{gray}{\  48.3 /} 3.1 & \textcolor{gray}{\  83.7 /} 2.0 & \textcolor{gray}{\  96.6 /} 2.0 & \textcolor{gray}{\  98.6 /} 1.8 & \textcolor{gray}{\  98.9 /} 0.0 & \textcolor{gray}{\  52.5 /} 9.4 & \textcolor{gray}{\  57.4 /} 0.7 & \textcolor{gray}{\  \textbf{84.0} /} 42.9 & \textcolor{gray}{\  57.7 /} \textbf{53.2} \\

    \cmidrule(lr){1-1}\cmidrule(lr){2-10}

    \multirow{1}{*}{Screw} 
        & \textcolor{gray}{\  51.7 /} 0.0 & \textcolor{gray}{\  73.3 /} 0.4 & \textcolor{gray}{\  \textbf{98.1} /} 0.0 & /\textcolor{gray}{\  98.0 /} 3.8 & \textcolor{gray}{\  93.9 /} 0.0& \textcolor{gray}{\  57.6 /} 3.7 & \textcolor{gray}{\  55.0 /} 0.6 & \textcolor{gray}{\  84.7 /} 21.4 & \textcolor{gray}{\  74.2 /} \textbf{36.4} \\

    \cmidrule(lr){1-1}\cmidrule(lr){2-10}

    \multirow{1}{*}{Toothbrush} 
      & \textcolor{gray}{\  75.3 /} 1.3 & \textcolor{gray}{\  90.8 /} 0.9 & \textcolor{gray}{\  100.0 /} 6.9 & \textcolor{gray}{\  100.0 /} 6.7 & \textcolor{gray}{\  100.0 /} 0.0 & \textcolor{gray}{\  70.8 /} 28.2 & \textcolor{gray}{\  60.1 /} 6.4 & \textcolor{gray}{\  65.9  /} 22.3 & \textcolor{gray}{\  \textbf{100.0} /} \textbf{75.5} \\
      
    \cmidrule(lr){1-1}\cmidrule(lr){2-10}

    \multirow{1}{*}{Transistor} 
      & \textcolor{gray}{\  61.7 /} 9.8 & \textcolor{gray}{\  76.4 /} 2.5 & \textcolor{gray}{\  \textbf{100.0} /} 7.8 & \textcolor{gray}{\  99.7 /} 8.1 & \textcolor{gray}{\  93.1 /} 5.3& \textcolor{gray}{\  60.1 /} 26.3 & \textcolor{gray}{\  58.7 /} 4.5 & \textcolor{gray}{\  86.5 /} 46.3 & \textcolor{gray}{\  91.0 /} \textbf{69.2} \\

    \cmidrule(lr){1-1}\cmidrule(lr){2-10}

    \multirow{1}{*}{Zipper} 
        & \textcolor{gray}{\  68.2 /} 5.4 & \textcolor{gray}{\  90.7 /} 0.8 & \textcolor{gray}{\  99.4 /} 13.6 & \textcolor{gray}{\  99.5 /} 13.7 & \textcolor{gray}{\  \textbf{100.0} /} 4.7& \textcolor{gray}{\  67.9 /} 35.7/ & \textcolor{gray}{\  65.6 /} 9.8 & \textcolor{gray}{\  82.2 /}48.6 & \textcolor{gray}{\  99.2 /} \textbf{92.5} \\

    \specialrule{1.5pt}{\aboverulesep}{\belowrulesep}

    \rowcolor{gray!11}
    \multirow{1}{*}{\textbf{Average}}
       & \textcolor{gray}{\  63.8 /} 4.6 & \textcolor{gray}{\  88.5 /} 2.2 & \textcolor{gray}{\  99.1 /} 7.2 & \textcolor{gray}{\  \textbf{99.5} /} 6.1 & \textcolor{gray}{\  98.0 /} 1.3 & \textcolor{gray}{\  63.6 /} 24.4 & \textcolor{gray}{\  58.7 /} 7.0 & \textcolor{gray}{\  71.6/} 30.1 & \textcolor{gray}{\  89.1 /} \textbf{75.1} \\

    \specialrule{1.5pt}{\aboverulesep}{\belowrulesep}
    
\end{tabular}}

\label{Table1:Novelty_Detection_MVtechad}	
\end{table*}

\begin{table*}[h]
    \caption{  Performance of AD methods on various datasets under clean evaluation and PGD-1000. For the experiments across \textbf{all tables}, adversarial attacks were considered, using $\epsilon=\frac{4}{255}$ for low-resolution images and $\epsilon=\frac{2}{255}$ for high-resolution images, measured by AUROC (\%). The table cells denote results in the ‘\graytext{Clean /} PGD’ format. Experiments performed in the one-class AD setup.
    }
    \footnotesize{\scriptsize $^*$These works incorporated adversarial training into their proposed AD methods.}
    

    \resizebox{ \textwidth}{!}{\begin{tabular}{cl*{9}{c}>{\columncolor[gray]{0.95}}c}


    \specialrule{1.5pt}{\aboverulesep}{\belowrulesep}
    & \multicolumn{1}{c}{Dataset} & &\multicolumn{7}{c}{Method}
    \\ \cmidrule(lr){2-2}  \cmidrule(lr){3-12}

    & &\multirow{2}{*}{DeepSVDD}&\multirow{2}{*}{CSI}&\multirow{2}{*}{MSAD} &\multirow{2}{*}{Transformaly}&\multirow{2}{*}{PatchCore} & \multirow{2}{*}{PrincipaLS$^*$} & \multirow{2}{*}{OCSDF$^*$} &\multirow{2}{*}{APAE$^*$} & \multirow{2}{*}{ZARND$^*$} & \cellcolor{gray!0} COBRA    \\
    & & & & & & &  &  &   &   & \cellcolor{gray!0} (\textit{Ours})  \\
    \specialrule{1.5pt}{\aboverulesep}{\belowrulesep}

    \multirow{8}{*}{\centering \rotatebox[origin=c]{90}{ \textbf{Low Res} }}
    &\multirow{1}{*}{\centering CIFAR10} 
     & \graytext{64.8 /} 8.7 & \graytext{94.3 /} 10.6 & \graytext{97.2 /} 4.8 & \graytext{\textbf{98.3} /} 3.7 & \graytext{68.3 /} 3.9 & \graytext{58.3 /} 33.2 & \graytext{58.7 /} 31.3 & \graytext{56.3 /} 2.2 & \graytext{89.7 /} 56.0 &  \graytext{83.7 /} \textbf{62.3} \\
    
    \cmidrule(lr){2-12} 

    &\multirow{1}{*}{CIFAR100}
    & \graytext{67.0 /} 3.6 & \graytext{89.6 /} 11.9 & \graytext{96.4 /} 8.4 & \graytext{\textbf{97.3} /} 9.4 & \graytext{66.8 /} 4.3 & \graytext{51.9 /} 26.2 & \graytext{50.2 /} 23.5 & \graytext{53.1 /} 4.1 & \graytext{88.4 /} 47.6 & \graytext{76.9 /} \textbf{51.7} \\
    
    \cmidrule(lr){2-12}

    &\multirow{1}{*}{MNIST} 
    & \graytext{94.8 /} 8.2 & \graytext{93.8 /} 3.4 & \graytext{96.0 /} 3.2 & \graytext{94.8 /} 7.9 & \graytext{83.2 /} 2.6 & \graytext{97.8 /} 83.1 & \graytext{96.1 /} 68.9 & \graytext{93.4 /} 34.7 & \graytext{\textbf{99.0} /} 91.2 & \graytext{92.8 /} \textbf{96.4} \\
    
    \cmidrule(lr){2-12}

    &\multirow{1}{*}{FMnist} 
    &  \graytext{94.5 /} 7.9 & \graytext{92.7 /} 5.8 & \graytext{94.2 /} 6.6 & \graytext{94.4 /} 7.4 & \graytext{77.4 /} 5.5 & \graytext{92.5 /} 69.2 & \graytext{91.8 /} 64.9 & \graytext{88.3 /} 19.5 & \graytext{\textbf{95.0} /} 82.3 & \graytext{93.1 /} \textbf{89.6} \\
    
    \cmidrule(lr){2-12}

    &\multirow{1}{*}{SVHN} 
    & \graytext{60.3 /} 1.5& \graytext{\textbf{96.8} /} 3.1 & \graytext{58.3 /} 0.2 & \graytext{56.9 /} 0.9 & \graytext{52.1 /} 2.1 & \graytext{63.0 /} 11.2 & \graytext{58.1 /} 9.7 & \graytext{52.6 /} 1.4 & \graytext{53.5 /} 9.6 & \graytext{89.3 /} \textbf{58.2}\\
    
    \specialrule{1.5pt}{\aboverulesep}{\belowrulesep}

    \multirow{6}{*}{\centering \rotatebox[origin=c]{90}{\textbf{High Res}}}
    &\multirow{1}{*}{ImageNet }
    &  \graytext{56.4 /} 4.0 & \graytext{91.6 /} 5.6 & \graytext{98.9 /} 2.6 & \graytext{\textbf{99.0} /} 2.9 & \graytext{67.6 /} 2.5 & \graytext{56.2 /} 28.3 & \graytext{55.3 /} 25.8 & \graytext{58.3 /} 2.1 & \graytext{96.4 /} 27.4 & \graytext{85.2 /} \textbf{57.0}\\


    
   \cmidrule(lr){2-12}

    &\multirow{1}{*}{VisA} 
    & \graytext{53.6 /} 1.8 & \graytext{62.5 /} 0.3 & \graytext{84.1 /}  4.6 & \graytext{85.5 /} 0.0 & \graytext{\textbf{95.1} /} 2.7 & \graytext{57.3 /} 16.1 & \graytext{53.0 /} 13.9 & \graytext{67.2 /} 9.1 & \graytext{71.8 /} 24.9 & \graytext{75.2 /} \textbf{73.8} \\

    \cmidrule(lr){2-12}

    &\multirow{1}{*}{CityScapes}
    & \graytext{59.7 /} 2.7 & \graytext{68.9 /} 0.1 & \graytext{86.5 /} 2.9 & \graytext{87.4 /} 4.5 & \graytext{\textbf{76.2} /} 6.1 & \graytext{60.3 /} 24.2 & \graytext{59.6 /} 20.1 & \graytext{63.0 /} 3.6 & \graytext{75.9 /} 28.6 & \graytext{81.7} \textbf{56.2} \\
    
    \cmidrule(lr){2-12}

    &\multirow{1}{*}{DAGM} 
    & \graytext{57.3 /} 2.7 & \graytext{74.5 /} 1.6 & \graytext{73.8 /} 0.0 & \graytext{81.4 /} 0.5 & \graytext{\textbf{93.6} /} 1.9 & \graytext{59.2 /} 24.8 & \graytext{57.6 /} 20.3 & \graytext{54.5 /} 13.8 & \graytext{64.5 /} 17.2 & \graytext{82.4 /} \textbf{56.8} \\
    
    \cmidrule(lr){2-12}

   &\multirow{1}{*}{ISIC2018} 
    &\graytext{64.1 /} 0.3 & \graytext{71.2 /} 0.0 & \graytext{76.7 /} 3.4 & \graytext{\textbf{86.6} /} 3.9 & \graytext{78.9 /} 0.0 &\graytext{61.7 /} 26.5 & \graytext{64.0 /} 18.6 & \graytext{67.2 /} 8.5 & \graytext{70.2/} 14.6 & \graytext{81.3 /} \textbf{56.1} \\

    \specialrule{1.5pt}{\aboverulesep}{\belowrulesep}

    \rowcolor{gray!11}
    &\multirow{1}{*}{\textbf{Average}} 
    & \graytext{67.3 /} 4.1 & \graytext{83.6 /} 4.2 & \graytext{86.2 /} 3.7 & \graytext{\textbf{88.1} /} 4.1 & \graytext{75.9 /} 3.1 & \graytext{65.8 /} 34.3 & \graytext{64.5 /} 29.7 & \graytext{65.4 /} 9.9 & \graytext{80.4 /}39.7 & 
    \graytext{84.1 /} \textbf{65.8} \\
    \specialrule{1.5pt}{\aboverulesep}{\belowrulesep}
    \end{tabular}}
\label{Table1:Novelty_Detection_main_table}
\end{table*}

\begin{table*}[h]
    \caption{    Performance of AD methods under clean evaluation and PGD-1000, measured by AUROC (\%). The table cells denote results in the ‘\graytext{Clean /} PGD’ format. Experiments performed in the unlabeled multi-class AD setup. }
    \label{Table:Novelty_Detection_multi_class_unlabel}
    \footnotesize{\scriptsize $^*$These works incorporated adversarial training into their proposed AD methods.}
    
    \resizebox{\linewidth}{!}{\begin{tabular}{ll*{7}{C{2cm}}} 

        \specialrule{1.5pt}{\aboverulesep}{\belowrulesep}

        \multicolumn{1}{c}{In} & \multicolumn{1}{c}{Out} & \multicolumn{7}{c}{Method} \\
        \cmidrule(lr){3-9} 
    
        &&\multicolumn{1}{c}{MSAD}&\multicolumn{1}{c}{Transformaly}&\multicolumn{1}{c}{PrincipaLS$^*$}&\multicolumn{1}{c}{OCSDF$^*$}&\multicolumn{1}{c}{APAE$^*$}& \multicolumn{1}{c}{ZARND$^*$} &\multicolumn{1}{c}{\cellcolor{gray!0} COBRA (\textit{Ours})}\\
    
        \specialrule{1.5pt}{\aboverulesep}{\belowrulesep}

        \multirow{5}{*}{CIFAR10} 
        &CIFAR100 & \graytext{76.9 /} 0.4 & \graytext{88.7 /} 0.0 & \graytext{54.8 /} 14.6 & \graytext{51.0 /} 12.8 & \graytext{53.6 /} 1.2 & \graytext{76.6  /} 34.1 & \cellcolor{gray!11} \graytext{76.0 /} \textbf{63.3} \\
        &SVHN & \graytext{94.6 /} 0.0 & \graytext{98.2 /} 1.2 & \graytext{72.1 /} 23.6 & \graytext{67.7 /} 18.8 & \graytext{60.8 /} 2.1 & \graytext{84.3 /} 42.7 & \cellcolor{gray!11} \graytext{98.5 /} \textbf{78.6}  \\
        &MNIST & \graytext{99.3 /} 1.6 & \graytext{99.4 /} 3.6 & \graytext{82.5 /} 42.7 & \graytext{74.2 /} 37.4 & \graytext{71.3 /} 15.3 & \graytext{99.4/} 82.2 & \cellcolor{gray!11} \graytext{80.8 /} \textbf{85.8}  \\
        &FMnist & \graytext{99.2 /} 3.8 & \graytext{99.1 /} 3.7 & \graytext{78.3 /} 38.5 & \graytext{64.5 /} 33.7 & \graytext{59.4 /} 9.4 & \graytext{98.2/}  67.3  & \cellcolor{gray!11} \graytext{82.8 /} \textbf{75.7} \\
        &ImageNet & \graytext{83.7 /} 0.0 & \graytext{92.8 /} 0.8 & \graytext{55.3 /} 12.3 & \graytext{52.8 /} 10.2 &\graytext{56.1 /} 0.3 & \graytext{71.5 /} 28.4 & \cellcolor{gray!11} \graytext{85.5 /} \textbf{53.1} \\
        
        \specialrule{1.5pt}{\aboverulesep}{\belowrulesep}

        \multirow{5}{*}{CIFAR100} 
        &CIFAR10& \graytext{61.4 /} 0.0 & \graytext{82.5 /} 0.3 &  \graytext{47.6 /} 8.1 & \graytext{51.1 /} 6.3 & \graytext{50.5 /} 0.7 & \graytext{64.6  /} 21.2 & \cellcolor{gray!11} \graytext{48.7 /} \textbf{27.5} \\
        &SVHN& \graytext{86.6 /} 2.7 & \graytext{94.7 /} 2.6 & \graytext{66.3 /} 13.2 & \graytext{58.7 /} 9.2 & \graytext{58.1 /} 1.1 & \graytext{70.0/} 26.8 & \cellcolor{gray!11} \graytext{93.2 /} \textbf{49.4} \\
        &MNIST& \graytext{97.4 /} 3.5 & \graytext{98.8 /} 0.8 & \graytext{80.4 /} 30.4 & \graytext{76.4 /} 28.9 & \graytext{74.7 /} 11.8 & \graytext{87.0  /} 30.4 & \cellcolor{gray!11} \graytext{77.8 /} \textbf{54.1} \\
        &FMnist & \graytext{96.5 /} 0.9 & \graytext{98.4 /} 5.2 & \graytext{72.7 /} 18.7& \graytext{62.8 /} 14.3 & \graytext{60.9 /} 9.7 & \cellcolor{gray!11} \graytext{97.3 /} \textbf{76.3} & \graytext{58.2 /}  32.9  \\
        &ImageNet & \graytext{71.6 /} 1.6 & \graytext{80.4 /} 2.0 & \graytext{51.6 /} 6.3 & \graytext{48.9 /} 5.4 & \graytext{52.7 /} 0.1 & \graytext{71.6 /} 21.8 & \cellcolor{gray!11} \graytext{69.1 /} \textbf{32.3} \\

        \specialrule{1.5pt}{\aboverulesep}{\belowrulesep}
    \end{tabular}}
 \centering
\end{table*}

\begin{table}[h]
\centering
\small 
\centering
\caption{ Performance of COBRA on various datasets under clean evaluation and
several adversarial attack, measured by AUROC (\%). Experiments performed in the one-class AD setup.}

\label{tab:advance_attack_ablation}
\renewcommand{\arraystretch}{0.5} 
    \resizebox{1\linewidth}{!}{\begin{tabular}{l*{7}{C{1.9cm}}} 

    \specialrule{1.5pt}{\aboverulesep}{\belowrulesep}

    \multicolumn{1}{c}{Dataset} & \multicolumn{6}{c}{Attack} \\
    
    \cmidrule(lr){1-1} \cmidrule(lr){2-8}
    
     & Clean & BlackBox & FGSM & CAA & $A^3$   & AutoAttack & PGD-1000  \\
    \specialrule{1.5pt}{\aboverulesep}{\belowrulesep}

    CIFAR10 & 83.7 & 81.8 & 70.2 & 64.5 & 60.7 &65.9& 62.3 \\    
    
    \cmidrule(lr){1-1} \cmidrule(lr){2-8}
    CIFAR100 & 76.9& 74.6& 64.5& 53.0 & 50.1& 54.8 & 51.7 \\  

    \cmidrule(lr){1-1} \cmidrule(lr){2-8}
    FMnist & 93.1& 92.9& 90.7 & 91.6 & 90.8 & 87.4 & 89.6  \\  
    
    \cmidrule(lr){1-1} \cmidrule(lr){2-8}
    ImageNet & 85.2 & 82.0& 71.4 & 53.6 & 61.8 & 59.4 & 57.0 \\ 

    \cmidrule(lr){1-1} \cmidrule(lr){2-8}
    MVTecAD & 89.1& 83.4 & 79.8 & 76.3 & 74.8 & 77.0 & 75.1 \\
    
    \cmidrule(lr){1-1} \cmidrule(lr){2-8}
    VisA & 75.2 & 74.6 & 74.0 & 73.5 & 71.6 & 74.9 &  73.8 \\


    \specialrule{1.5pt}{\aboverulesep}{\belowrulesep}
\end{tabular}}
\end{table}


 \begin{figure}[h]
  \centering
  \includegraphics[width=1.0\linewidth]{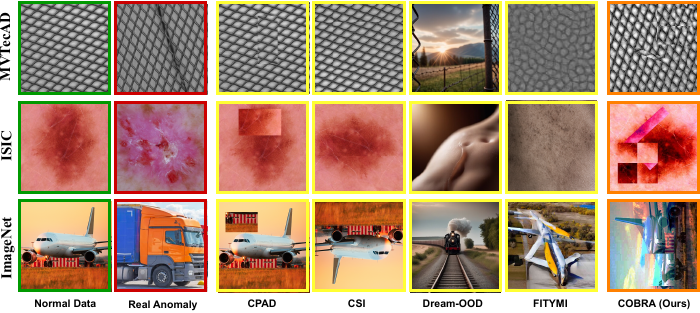}
  \caption{ The figure highlights the challenges CPAD and CSI face in generating inncorrect anomalies due to the absence of a threshold. Techniques such as FITYMI and Dream-OOD, which generate anomalies from the embedding space of a pretrained model, typically lead to a loss of pixel-level detail and show biases towards the dataset used for pre-training (e.g., ImageNet). Such biases decrease their effectiveness on datasets not seen during pre-training, such as medical imaging datasets like ISIC. In contrast, COBRA, by adapting to the distribution of the normal dataset, efficiently crafts informative anomalies in the pixel space and utilizes a thresholding method to filter out incorrect anomalies, all \textbf{without} the need for any additional datasets.}
  \label{fig:different-anomaly-generator-data-comparison}
\end{figure}

\begin{table}[h]
\centering
\small 
\centering
 \caption{Comparison of performance of COBRA with replacing alternative anomaly synthesis methods, in terms of effectiveness for the robust AD task. \textit{None} corresponds to a scenario where we neglect any pseudo-anomaly and adapt COBRA for that setting.  }

\label{tab:pseudo_anomaly_strategy}
\renewcommand{\arraystretch}{0.5} 
    \resizebox{1\linewidth}{!}{\begin{tabular}{l*{8}{C{1.9cm}}} 

    \specialrule{1.5pt}{\aboverulesep}{\belowrulesep}

    \multicolumn{1}{c}{Dataset} &  \multicolumn{8}{c}{Anomaly Craft Strategy} \\
    
    \cmidrule(lr){1-1}\cmidrule(lr){2-9}
    
     & \textit{None} &   CPAD   & CSI  & GOE & VOS &FITYMI &  Dream-OOD     & Ours  \\
    \specialrule{1.5pt}{\aboverulesep}{\belowrulesep}

    MVTecAD & \graytext{57.6 /} 10.8 & \graytext{86.2 /} 70.8 & \graytext{ 61.4 /} 12.9 & \graytext{58.1 /} 25.3 & \graytext{53.4 /} 15.8 & \graytext{65.0 /} 38.7 & \graytext{68.6 /} 36.4  & \graytext{\textbf{89.1} /} \textbf{75.1}  \\    
    
    \cmidrule(lr){1-1} \cmidrule(lr){2-9}
    ImageNet & \graytext{72.8 /} 32.5 & \graytext{67.5 /} 43.4 & \graytext{82.7 /} 58.2 & \graytext{81.9 /} 60.4 & \graytext{72.4 /} 56.5 & \graytext{68.9 /} 47.2  & \graytext{\textbf{87.3} /} \textbf{64.1}  & \graytext{85.2 /} 57.0  \\
    
    \cmidrule(lr){1-1} \cmidrule(lr){2-9}
    CIFAR10 & \graytext{78.6 /} 50.3 & \graytext{69.4 /} 53.8 & \graytext{82.9 /} 60.7 & \graytext{\textbf{84.2} /} 58.8 & \graytext{79.3 /} 53.1  & \graytext{76.9 /} 50.6  & \graytext{75.2 /} 57.8  & \graytext{83.7 /} \textbf{62.3}  \\    

    \cmidrule(lr){1-1} \cmidrule(lr){2-9}
    FMnist& \graytext{82.4 /} 71.7 & \graytext{86.3 /} 78.5 & \graytext{89.5 /} 82.6 & \graytext{73.9 /} 64.1 & \graytext{68.2 /} 61.9  & \graytext{71.7 /} 62.0  & \graytext{76.4 /} 68.5 & \graytext{\textbf{93.1} /} \textbf{89.6}  \\

    \rowcolor{gray!11}
    \specialrule{1.5pt}{\aboverulesep}{\belowrulesep}
    \textbf{Average} & \graytext{72.8 /} 41.3 & \graytext{77.3 /} 61.6 & \graytext{79.0 /} 53.1 & \graytext{74.5 /} 52.1 & \graytext{68.3 /} 47.3 & \graytext{70.6 /} 50.3  & \graytext{76.9 /} 56.7  & \graytext{\textbf{87.8} /} \textbf{71.0}  \\  
    \specialrule{1.5pt}{\aboverulesep}{\belowrulesep}
\end{tabular}}
\end{table}

\begin{table}[h]
\small 
\centering
\caption{Comparison of the performance of COBRA with alternative loss functions versus \(\mathcal{L}_{\text{COBRA}}\), in terms of effectiveness for the robust AD task. }
\label{tab:loss_function_ablaiton}
\renewcommand{\arraystretch}{0.5} 
    \resizebox{1\linewidth}{!}{\begin{tabular}{l*{6}{C{2.5cm}}} 

    \specialrule{1.5pt}{\aboverulesep}{\belowrulesep}

    \multicolumn{1}{c}{Dataset} &  \multicolumn{6}{c}{Loss Function} \\
    
    \cmidrule(lr){1-1}\cmidrule(lr){2-7}
    
     &   \textbf{$\mathcal{L}_{\text{CLS}}$}  &\textbf{$\mathcal{L}_{\text{CL}}$} & \textbf{$\mathcal{L}_{\text{SupCL}}$}  & \textbf{$\mathcal{L}_{\text{Opposite}}$} &   \textbf{$\mathcal{L}_{\text{COBRA}}$}  &    {\scriptsize \textbf{$\mathcal{L}_{\text{COBRA}} + \mathcal{L}_{\text{CLS}}$}}   \\
    \specialrule{1.5pt}{\aboverulesep}{\belowrulesep}

    MVTecAD & \graytext{62.6 /} 40.4 & \graytext{76.4 /} 58.7 & \graytext{80.4 /} 60.5  & \graytext{64.0 /} 53.6  & \graytext{83.7 /} 68.2  & \graytext{\textbf{89.1} /} \textbf{75.1}  \\    
    
    \cmidrule(lr){1-1} \cmidrule(lr){2-7}
    ImageNet & \graytext{59.5 /} 45.1& \graytext{68.3 /} 47.6 & \graytext{74.6 /} 46.3  & \graytext{57.4 /} 45.8  & \graytext{82.9 /} 54.3  & \graytext{\textbf{85.2} /} \textbf{57.0}  \\     
    
    \cmidrule(lr){1-1} \cmidrule(lr){2-7}
    CIFAR10  & \graytext{62.6 /} 49.6 & \graytext{67.9 /} 54.3 & \graytext{74.2 /} 53.8  & \graytext{65.9 /} 52.4  & \graytext{78.5 /} 61.0  & \graytext{\textbf{83.7} /} \textbf{62.3}  \\    

    \cmidrule(lr){1-1} \cmidrule(lr){2-7}
    FMnist & \graytext{82.0 /} 78.4 & \graytext{88.3 /} 82.5 & \graytext{91.5 /} 83.2  & \graytext{80.4 /} 73.5 & \graytext{92.8 /} 87.6  & \graytext{\textbf{93.1} /} \textbf{89.6}  \\   

    \rowcolor{gray!11}
    \specialrule{1.5pt}{\aboverulesep}{\belowrulesep}
    \textbf{Average} & \graytext{66.7 /} 53.3 & \graytext{75.2 /} 60.8 & \graytext{80.4 /} 60.9 & \graytext{66.9 /} 56.3 & \graytext{84.5 /} 67.7  & \graytext{\textbf{87.7} /} \textbf{71.0}  \\  
    \specialrule{1.5pt}{\aboverulesep}{\belowrulesep}
\end{tabular}}
\end{table}

 \begin{figure}[h]
  \centering
  \begin{minipage}{\textwidth}
  \centering
  \includegraphics[width=1.0\linewidth]{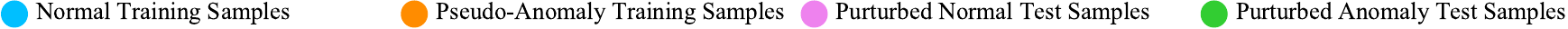}  
 \subfloat[CL]{\includegraphics[width=0.22\linewidth]{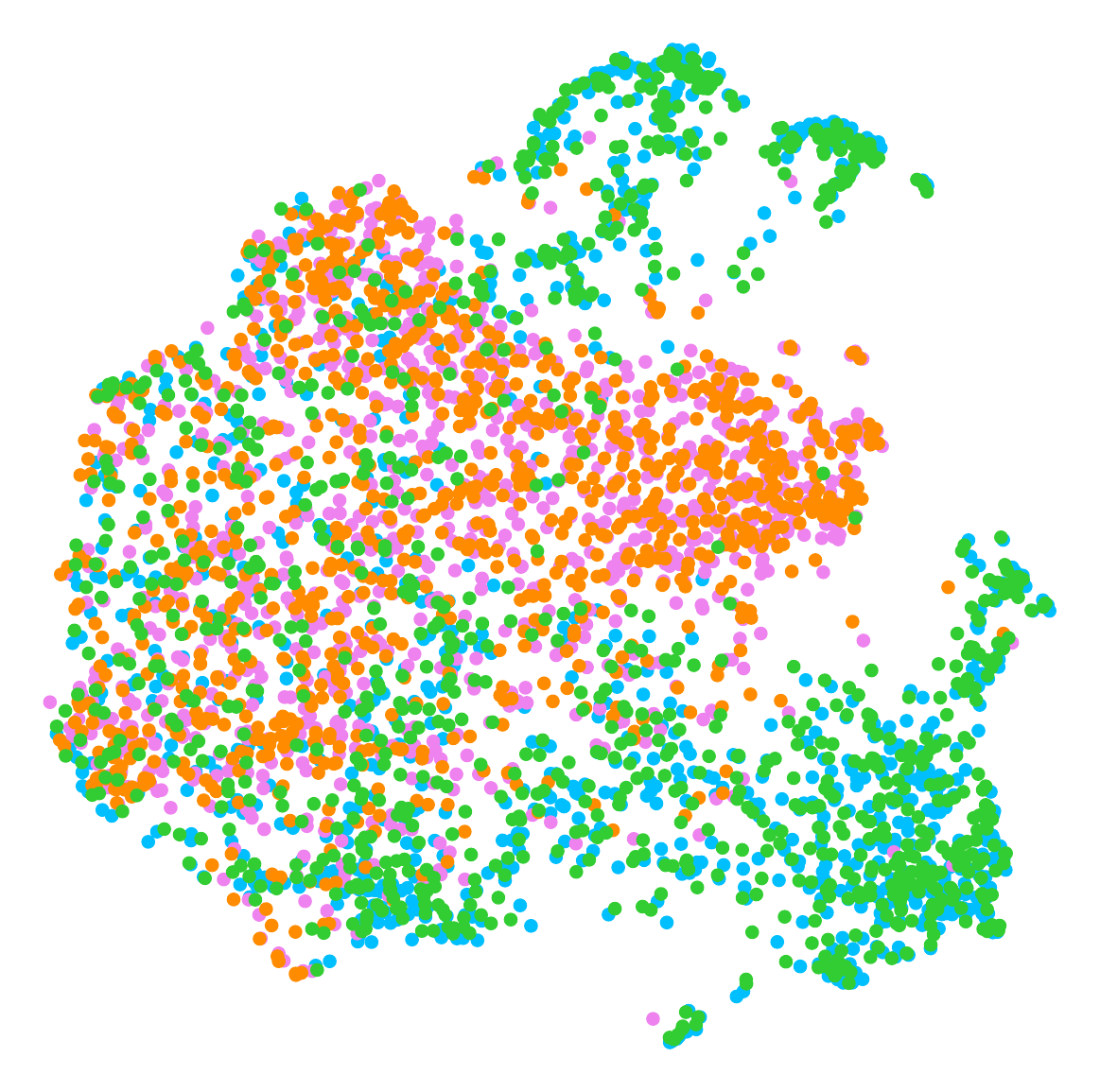}\vspace{5pt}\label{fig:mvtecad}}
  \hfill
   \subfloat[CLS]{\includegraphics[width=0.22\linewidth]{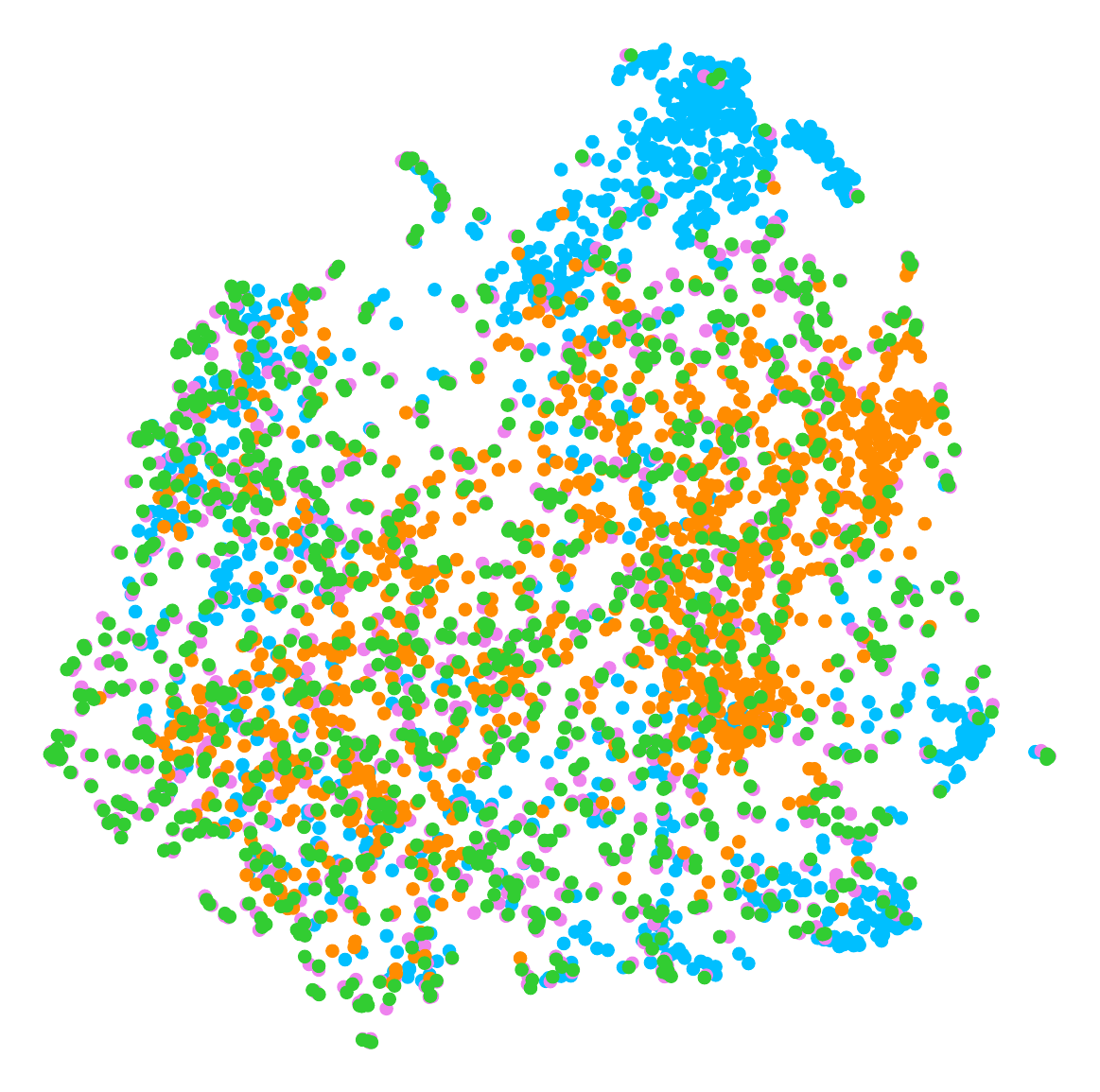}\vspace{5pt}\label{fig:cifar-10}}
  \hfill
  \subfloat[SupCLR]{\includegraphics[width=0.22\linewidth]{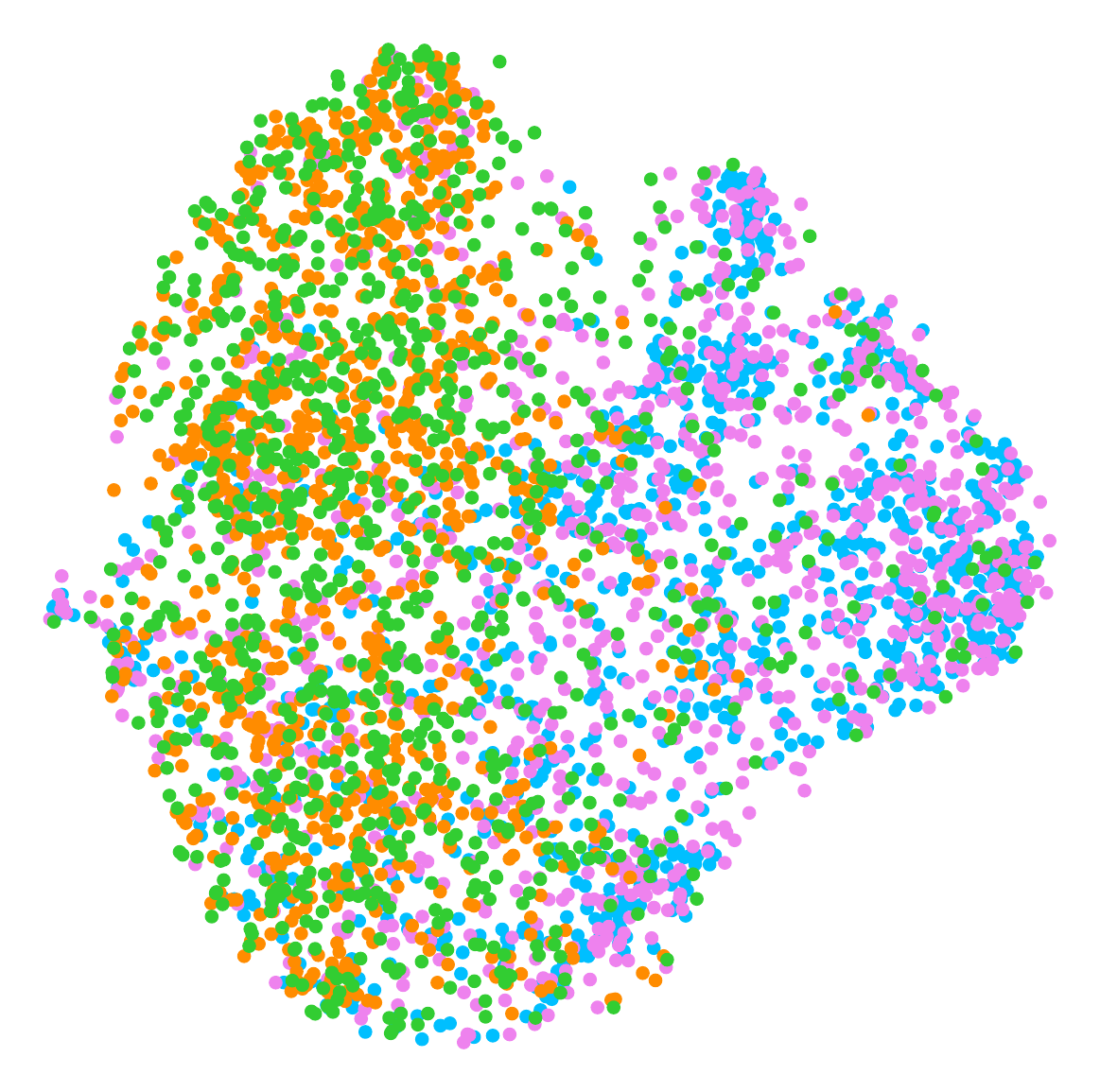}\vspace{5pt}\label{fig:mnist}}
  \hfill
   \subfloat[COBRA]{\includegraphics[width=0.22\linewidth]{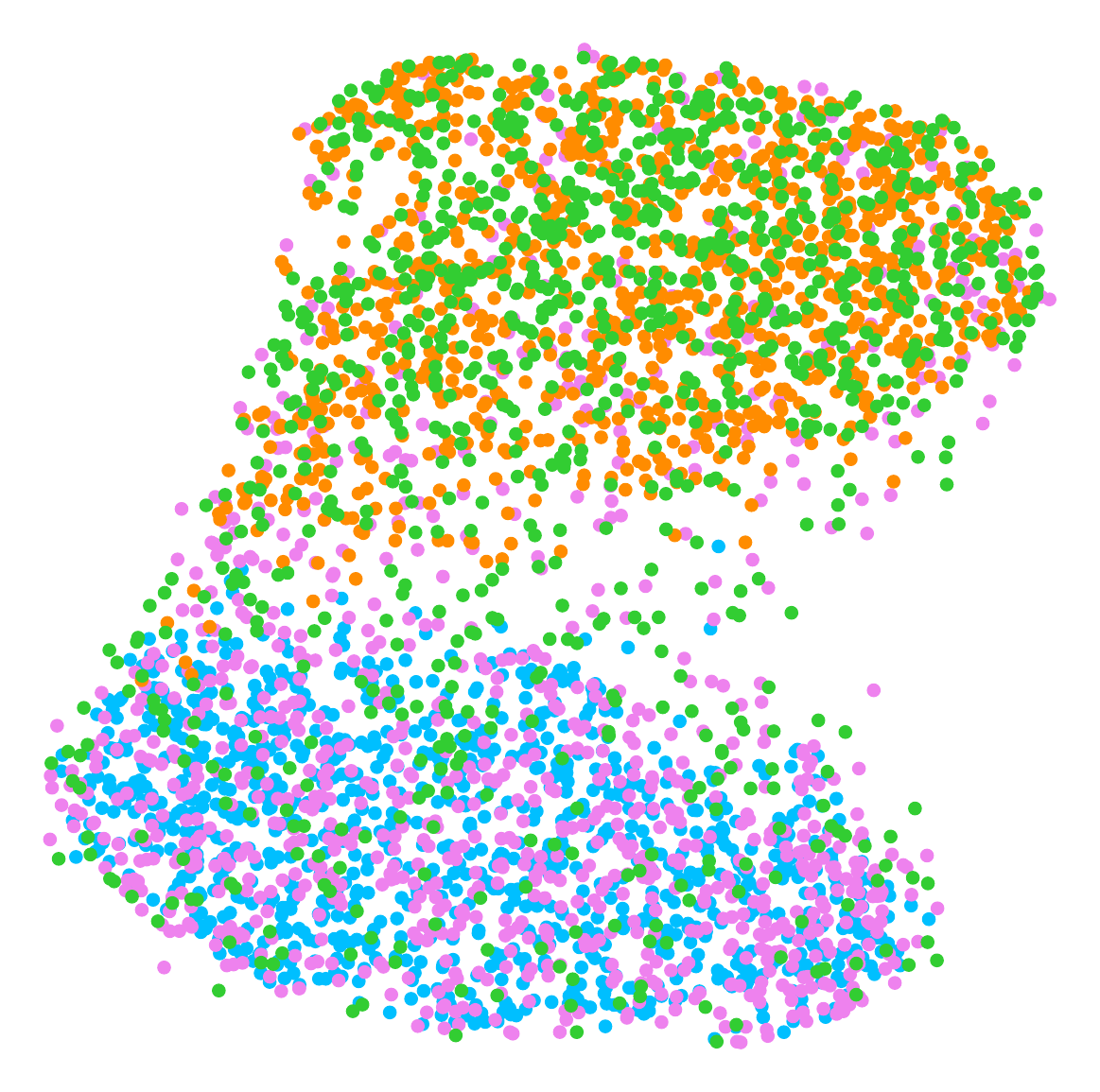}\vspace{5pt}\label{fig:svhn}}
  
  \caption{UMAP visualization of features extracted by the encoder $f$, trained with various loss functions on the CIFAR-10 dataset, is presented in a one-class setup with the 'Automobile' class designated as the normal set. For this particular experiment, all elements except the loss function remain constant to ensure a fair comparison. }
  \label{fig:loss_adv_umap_visualization}
  \end{minipage}
\end{figure}

\section{Experiments}
In this section, we verify the effectiveness of COBRA in robust AD with several benchmark datasets, encompassing those that are large-scale and real-world. Evaluation is conducted to assess existing AD methods, including both clean and adversarially trained methods, as well as our own method, under both clean and various adversarial attack scenarios. Table \ref{Table1:Novelty_Detection_MVtechad} provides a comparative analysis in a one-class setting on the MVTecAD dataset, a challenging real-world benchmark in AD. Additional comparisons in one-class settings across other benchmarks are detailed in Table \ref{Table1:Novelty_Detection_main_table}, while Table \ref{Table:Novelty_Detection_multi_class_unlabel} showcases our method's superiority in unlabeled multi-class setting. Moreover, we demonstrate our method's robustness by evaluating it against several attacks presented in Table \ref{tab:advance_attack_ablation}. Additional details on the adaptation of attacks and supplementary evaluation metrics can be found in \ref{Appendix_Adv__Adaptation} and \ref{appendix_Supplementary_Metrics}.

\noindent\textbf{Experimental Setup.}\label{Experimental setup}\ \
Our experiments were conducted in two categories: one-class and unlabeled multi-class anomaly detection (AD). In the one-class setup, considering a dataset \(D\) with \(M\) classes, experiments were conducted by treating each class in turn as the normal set and the other \(M-1\) classes as the anomaly set. This process was repeated for each class, and performance was averaged across all classes to report the overall detection performance. In the unlabeled multi-class setup, this setting incorporates another dataset \(D'\), considering one dataset as the normal set and another as the anomaly set. We compared COBRA with PANDA \cite{reiss2021panda}, Transformaly \cite{cohen2021transformaly}, Patchcore \cite{roth2021total}, CSI \cite{tack2020csi}, MSAD \cite{reiss2021mean}, ReContrast \cite{guo2024recontrast}, and Draem \cite{zavrtanik2021draem}, as well as methods specifically proposed for robust AD, including ZARND \cite{ZARND}, PrincipaLS \cite{lo2022adversarially}, OCSDF \cite{bethune2023robust}, and APAE \cite{goodge2021robustness}. Details about each mentioned method can be found in Appendix \ref{Appendix_Details_Related_Work}.

\noindent\textbf{ Evaluation Details.}\ \ To evaluate the methods' adversarial robustness, both normal and anomalous test samples will be subjected to end-to-end adversarial attacks targeting the methods' anomaly scores. We set the value of \(\epsilon\) to \(\frac{4}{255}\) for low-resolution datasets and to \(\frac{2}{255}\) for high-resolution datasets. For the PGD attack, we set the number of steps \(N\) to 1000, initializing the attack from 10 different random starting points for each trial to enhance the attack's effectiveness and coverage. Furthermore, to highlight COBRA's robust performance, we considered additional strong attacks, including AutoAttack (AA), Adaptive AutoAttack (\(A^3\)), and black-box attacks. Furthermore, to highlight COBRA's robust performance, we considered an additional range of simple to strong attacks, including black-box attacks \cite{guo2019simple}, FGSM attacks \cite{goodfellow2014explaining}, CAA \cite{Mao2020CompositeAA}, AutoAttack (AA), and Adaptive AutoAttack ($A^3$). Other methods' performance under AutoAttack can be found in Appendix \ref{Appendix_AutoAttack}.  Additionally, details on the model's evaluation under both $\ell_\infty$ and $\ell_2$ PGD attacks across varying epsilon values are presented in Appendix \ref{appendix:linf_diverse_epsilon}.

\textbf{ Implementation Details \& Datasets} \label{implementaion_detail}  \ \ For obtaining the threshold $\lambda$, we utilized a from-scratch ResNet-18 as \(C\) and trained on the created dataset for 100 epochs. For adversarial training, we use PGD-10 step and \(\epsilon=\frac{4}{255}\). We employ ResNet-18 as the foundational encoder network, accompanied by an auxiliary head comprising a 2-layer multi-layer perceptron with a 128-dimensional embedding dimension. More details about the implementation can be found in Appendix \ref{Appendix_Implementation_Details}. COBRA is evaluated using challenging datasets that includes both high- and low-resolution images. The high-resolution dataset comprises MVTecAD \cite{bergmann2019mvtec}, VisA \cite{zou2022spot}, CityScapes \cite{Cityscapes}, ImageNet \cite{deng2009imagenet}, ISIC2018 \cite{codella2019skin}, and DAGM \cite{wieler_matthias_2007_8086136}, while the low-resolution dataset includes SVHN \cite{goodfellow2013multi}, FMNIST \cite{xiao2017fashion}, CIFAR10, CIFAR100, and MNIST. Further details can be found in Appendix \ref{Appendix_Dataset_Details}.    

\textbf{Analyzing Results.} 
 The results presented underscore COBRA's effectiveness as an robust AD method. Remarkably, COBRA enhances the average robust detection performance across various datasets by up to \textbf{26.1\%}, \textbf{without} relying on pre-trained models or extra datasets. This demonstrates COBRA's real-world applicability by enhancing robust performance on the MVTecAD dataset from \textbf{30.1\%} to \textbf{75.1\%}. COBRA's versatility is further highlighted by its general applicability to different AD scenarios, including one-class and unlabeled multi-class setups. Notably, in open-world applications where robustness is vital, a slight drop in clean performance is considered a worthwhile trade-off for enhanced robustness. Our results align with this perspective, achieving an average of \textbf{84.1\%} in clean and \textbf{65.8\%} in adversarial settings across various datasets. This performance surpasses methods like Transformaly \cite{cohen2021transformaly}, which, while achieving 88.2\% in clean settings, significantly falls to 4.1\% in adversarial scenarios. Furthermore, we replaced our adversarial training with clean training in the COBRA Pipeline. As expected, and in line with findings reported in the literature \cite{tsipras2018robustness}, this resulted in decreased robust detection performance. However, it improved clean detection performance from an AUROC of 84.1\% to \textbf{90.7\%}.

\section{Ablation Study} \label{ablation_sec}

\noindent\textbf{Pseudo-anomaly Generating Strategy.} \ \ In order to demonstrate the superiority of our strategy for pseudo-anomaly sample crafting, with other modules fixed, we replaced ours with alternative methods. We provided a brief description of alternative methods in Section \ref{Distribution_A}. The results, which are presented in Table \ref{tab:pseudo_anomaly_strategy}, along with a visualization comparison of samples in Figure \ref{fig:different-anomaly-generator-data-comparison}, show the superiority of our effective synthesizer method. Notably, our strategy, without using any extra data, outperforms Dream-OOD with billions of sample complexity by a margin of 15\%. In Appendix \ref{appendix_Quality_of_Generated_Abnormal}, we further evaluate the quality of our generated data.\\
\noindent\textbf{Adversarial Training Objective Function.} \ \ \label{loss_ablation_section}
We replaced our proposed loss function with various alternatives, such as classification (CLS), CL and Supervised CL. The results, detailed in Table \ref{tab:loss_function_ablaiton}, reveal that the COBRA loss function, by generating challenging intra- and inter-group adversarial examples during training, surpasses other alternatives significantly. $L_{\text{COBRA}}$ outperforms CLS by increasing normal distribution compactness provided by intra-group perturbations, and outperforms CL  and SupCL by considering opposite pairs for increasing margins, as illustrated in Figure \ref{fig:loss_adv_umap_visualization}. As the results indicate, our perfomanc outperforms other loss functions by 11\%. Additional ablation studies, experimental results including error bars, limitations, and qualitative visualizations are provided in the Appendix.
\section{Conclusion}

In conclusion, our work introduces COBRA, a novel and effective approach for enhancing AD methods' robustness against adversarial attacks. By leveraging a novel loss function inspired by contrastive learning and strategically crafting informative anomaly samples, COBRA achieves superior detection performance under both clean and adversarial evaluation conditions. We verify COBRA through comprehensive ablation experiments on its different components. Moreover, our extensive experiments across multiple challenging datasets, as well as under various strong attacks, confirm our method's effectiveness, setting a new benchmark for future research in reliable AD.

\clearpage
{
\small
\bibliographystyle{unsrtnat}
\bibliography{COBRA_ICLR2025}

\begin{thebibliography}{100}
\providecommand{\natexlab}[1]{#1}
\providecommand{\url}[1]{\texttt{#1}}
\expandafter\ifx\csname urlstyle\endcsname\relax
  \providecommand{\doi}[1]{doi: #1}\else
  \providecommand{\doi}{doi: \begingroup \urlstyle{rm}\Url}\fi

\bibitem[Bendale and Boult(2015)]{bendale2015towards}
Abhijit Bendale and Terrance Boult.
\newblock Towards open world recognition.
\newblock In \emph{Proceedings of the IEEE Conference on Computer Vision and Pattern Recognition (CVPR)}, pages 1893--1902, 2015.

\bibitem[Perera et~al.(2021)Perera, Oza, and Patel]{perera2021one}
Pramuditha Perera, Poojan Oza, and Vishal~M Patel.
\newblock One-class classification: A survey.
\newblock \emph{arXiv preprint arXiv:2101.03064}, 2021.

\bibitem[Ruff et~al.(2018)Ruff, Vandermeulen, Goernitz, Deecke, Siddiqui, Binder, M{\"u}ller, and Kloft]{ruff2018deep}
Lukas Ruff, Robert Vandermeulen, Nico Goernitz, Lucas Deecke, Shoaib~Ahmed Siddiqui, Alexander Binder, Emmanuel M{\"u}ller, and Marius Kloft.
\newblock Deep one-class classification.
\newblock In \emph{International conference on machine learning}, pages 4393--4402. PMLR, 2018.

\bibitem[Tack et~al.(2020)Tack, Mo, Jeong, and Shin]{tack2020csi}
Jihoon Tack, Sangwoo Mo, Jongheon Jeong, and Jinwoo Shin.
\newblock Csi: Novelty detection via contrastive learning on distributionally shifted instances.
\newblock \emph{Advances in neural information processing systems}, 33:\penalty0 11839--11852, 2020.

\bibitem[Bergman et~al.(2020)Bergman, Cohen, and Hoshen]{bergman2020deep}
Liron Bergman, Niv Cohen, and Yedid Hoshen.
\newblock Deep nearest neighbor anomaly detection.
\newblock \emph{arXiv preprint arXiv:2002.10445}, 2020.

\bibitem[Reiss et~al.(2021)Reiss, Cohen, Bergman, and Hoshen]{reiss2021panda}
Tal Reiss, Niv Cohen, Liron Bergman, and Yedid Hoshen.
\newblock Panda: Adapting pretrained features for anomaly detection and segmentation.
\newblock In \emph{Proceedings of the IEEE/CVF Conference on Computer Vision and Pattern Recognition}, pages 2806--2814, 2021.

\bibitem[Bergmann et~al.(2019)Bergmann, Fauser, Sattlegger, and Steger]{bergmann2019mvtec}
Paul Bergmann, Michael Fauser, David Sattlegger, and Carsten Steger.
\newblock Mvtec ad--a comprehensive real-world dataset for unsupervised anomaly detection.
\newblock In \emph{Proceedings of the IEEE/CVF conference on computer vision and pattern recognition}, pages 9592--9600, 2019.

\bibitem[Krizhevsky et~al.(2009)Krizhevsky, Hinton, et~al.]{krizhevsky2009learning}
Alex Krizhevsky, Geoffrey Hinton, et~al.
\newblock Learning multiple layers of features from tiny images.
\newblock 2009.

\bibitem[Azizmalayeri et~al.(2022)Azizmalayeri, Soltani~Moakhar, Zarei, Zohrabi, Manzuri, and Rohban]{azizmalayeri2022your}
Mohammad Azizmalayeri, Arshia Soltani~Moakhar, Arman Zarei, Reihaneh Zohrabi, Mohammad Manzuri, and Mohammad~Hossein Rohban.
\newblock Your out-of-distribution detection method is not robust!
\newblock \emph{Advances in Neural Information Processing Systems}, 35:\penalty0 4887--4901, 2022.

\bibitem[Lo et~al.(2022)Lo, Oza, and Patel]{lo2022adversarially}
Shao-Yuan Lo, Poojan Oza, and Vishal~M Patel.
\newblock Adversarially robust one-class novelty detection.
\newblock \emph{IEEE Transactions on Pattern Analysis and Machine Intelligence}, 2022.

\bibitem[Chen et~al.(2020{\natexlab{a}})Chen, Li, Wu, Liang, and Jha]{chen2020robust}
Jiefeng Chen, Yixuan Li, Xi~Wu, Yingyu Liang, and Somesh Jha.
\newblock Robust out-of-distribution detection for neural networks.
\newblock \emph{arXiv preprint arXiv:2003.09711}, 2020{\natexlab{a}}.

\bibitem[Shao et~al.(2020)Shao, Perera, Yuen, and Patel]{shao2020open}
Rui Shao, Pramuditha Perera, Pong~C Yuen, and Vishal~M Patel.
\newblock Open-set adversarial defense.
\newblock In \emph{Computer Vision--ECCV 2020: 16th European Conference, Glasgow, UK, August 23--28, 2020, Proceedings, Part XVII 16}, pages 682--698. Springer, 2020.

\bibitem[Shao et~al.(2022)Shao, Perera, Yuen, and Patel]{shao2022open}
Rui Shao, Pramuditha Perera, Pong~C Yuen, and Vishal~M Patel.
\newblock Open-set adversarial defense with clean-adversarial mutual learning.
\newblock \emph{International Journal of Computer Vision}, 130\penalty0 (4):\penalty0 1070--1087, 2022.

\bibitem[B{\'e}thune et~al.(2023)B{\'e}thune, Novello, Boissin, Coiffier, Serrurier, Vincenot, and Troya-Galvis]{bethune2023robust}
Louis B{\'e}thune, Paul Novello, Thibaut Boissin, Guillaume Coiffier, Mathieu Serrurier, Quentin Vincenot, and Andres Troya-Galvis.
\newblock Robust one-class classification with signed distance function using 1-lipschitz neural networks.
\newblock \emph{arXiv preprint arXiv:2303.01978}, 2023.

\bibitem[Goodge et~al.(2021)Goodge, Hooi, Ng, and Ng]{goodge2021robustness}
Adam Goodge, Bryan Hooi, See~Kiong Ng, and Wee~Siong Ng.
\newblock Robustness of autoencoders for anomaly detection under adversarial impact.
\newblock In \emph{Proceedings of the Twenty-Ninth International Conference on International Joint Conferences on Artificial Intelligence}, pages 1244--1250, 2021.

\bibitem[Chen et~al.(2021{\natexlab{a}})Chen, Li, Wu, Liang, and Jha]{chen2021atom}
Jiefeng Chen, Yixuan Li, Xi~Wu, Yingyu Liang, and Somesh Jha.
\newblock Atom: Robustifying out-of-distribution detection using outlier mining.
\newblock In \emph{Machine Learning and Knowledge Discovery in Databases. Research Track: European Conference, ECML PKDD 2021, Bilbao, Spain, September 13--17, 2021, Proceedings, Part III 21}, pages 430--445. Springer, 2021{\natexlab{a}}.

\bibitem[Bai et~al.(2021)Bai, Luo, Zhao, Wen, and Wang]{bai2021recent}
Tao Bai, Jinqi Luo, Jun Zhao, Bihan Wen, and Qian Wang.
\newblock Recent advances in adversarial training for adversarial robustness.
\newblock \emph{arXiv preprint arXiv:2102.01356}, 2021.

\bibitem[Madry et~al.(2017)Madry, Makelov, Schmidt, Tsipras, and Vladu]{madry2017towards}
Aleksander Madry, Aleksandar Makelov, Ludwig Schmidt, Dimitris Tsipras, and Adrian Vladu.
\newblock Towards deep learning models resistant to adversarial attacks.
\newblock \emph{arXiv preprint arXiv:1706.06083}, 2017.

\bibitem[Glodek et~al.(2013)Glodek, Schels, and Schwenker]{glodek2013ensemble}
Michael Glodek, Martin Schels, and Friedhelm Schwenker.
\newblock Ensemble gaussian mixture models for probability density estimation.
\newblock \emph{Computational statistics}, 28:\penalty0 127--138, 2013.

\bibitem[Chen et~al.(2021{\natexlab{b}})Chen, Niu, Gong, Li, Yang, and Sugiyama]{chen2021large}
Shuo Chen, Gang Niu, Chen Gong, Jun Li, Jian Yang, and Masashi Sugiyama.
\newblock Large-margin contrastive learning with distance polarization regularizer.
\newblock In \emph{International Conference on Machine Learning}, pages 1673--1683. PMLR, 2021{\natexlab{b}}.

\bibitem[Cheng et~al.(2023)Cheng, Cao, Ye, Zhu, Li, and Zou]{cheng2023ml}
Xuxin Cheng, Bowen Cao, Qichen Ye, Zhihong Zhu, Hongxiang Li, and Yuexian Zou.
\newblock Ml-lmcl: Mutual learning and large-margin contrastive learning for improving asr robustness in spoken language understanding.
\newblock \emph{arXiv preprint arXiv:2311.11375}, 2023.

\bibitem[Guo and Zhang(2021)]{guo2021recent}
Yiwen Guo and Changshui Zhang.
\newblock Recent advances in large margin learning.
\newblock \emph{IEEE Transactions on Pattern Analysis and Machine Intelligence}, 44\penalty0 (10):\penalty0 7167--7174, 2021.

\bibitem[Chen et~al.(2020{\natexlab{b}})Chen, Kornblith, Norouzi, and Hinton]{chen2020simple}
Ting Chen, Simon Kornblith, Mohammad Norouzi, and Geoffrey Hinton.
\newblock A simple framework for contrastive learning of visual representations.
\newblock In \emph{Proceedings of the 37th International Conference on Machine Learning (ICML)}, pages 1597--1607. PMLR, 2020{\natexlab{b}}.

\bibitem[He et~al.(2020)He, Fan, Wu, Xie, and Girshick]{he2020momentum}
Kaiming He, Haoqi Fan, Yuxin Wu, Saining Xie, and Ross Girshick.
\newblock Momentum contrast for unsupervised visual representation learning.
\newblock In \emph{Proceedings of the IEEE/CVF conference on computer vision and pattern recognition}, pages 9729--9738, 2020.

\bibitem[Guo et~al.(2024)Guo, Jia, Zhang, Li, et~al.]{guo2024recontrast}
Jia Guo, Lize Jia, Weihang Zhang, Huiqi Li, et~al.
\newblock Recontrast: Domain-specific anomaly detection via contrastive reconstruction.
\newblock \emph{Advances in Neural Information Processing Systems}, 36, 2024.

\bibitem[Reiss and Hoshen(2021)]{reiss2021mean}
Tal Reiss and Yedid Hoshen.
\newblock Mean-shifted contrastive loss for anomaly detection.
\newblock \emph{arXiv preprint arXiv:2106.03844}, 2021.

\bibitem[Croce and Hein(2020)]{croce2020reliable}
Francesco Croce and Matthias Hein.
\newblock Reliable evaluation of adversarial robustness with an ensemble of diverse parameter-free attacks.
\newblock In \emph{International conference on machine learning}, pages 2206--2216. PMLR, 2020.

\bibitem[Liu et~al.(2022)Liu, Cheng, Gao, Liu, Zhang, and Song]{liu2022practical}
Ye~Liu, Yaya Cheng, Lianli Gao, Xianglong Liu, Qilong Zhang, and Jingkuan Song.
\newblock Practical evaluation of adversarial robustness via adaptive auto attack, 2022.

\bibitem[Cordts et~al.(2016)Cordts, Omran, Ramos, Rehfeld, Enzweiler, Benenson, Franke, Roth, and Schiele]{Cityscapes}
Marius Cordts, Mohamed Omran, Sebastian Ramos, Timo Rehfeld, Markus Enzweiler, Rodrigo Benenson, Uwe Franke, Stefan Roth, and Bernt Schiele.
\newblock The cityscapes dataset for semantic urban scene understanding.
\newblock In \emph{Proc. of the IEEE Conference on Computer Vision and Pattern Recognition (CVPR)}, 2016.

\bibitem[Deng et~al.(2009)Deng, Dong, Socher, Li, Li, and Fei-Fei]{deng2009imagenet}
Jia Deng, Wei Dong, Richard Socher, Li-Jia Li, Kai Li, and Li~Fei-Fei.
\newblock Imagenet: A large-scale hierarchical image database.
\newblock In \emph{2009 IEEE Conference on Computer Vision and Pattern Recognition}, pages 248--255. IEEE, 2009.

\bibitem[Codella et~al.(2019)Codella, Rotemberg, Tschandl, Celebi, Dusza, Gutman, Helba, Kalloo, Liopyris, Marchetti, et~al.]{codella2019skin}
Noel Codella, Veronica Rotemberg, Philipp Tschandl, M~Emre Celebi, Stephen Dusza, David Gutman, Brian Helba, Aadi Kalloo, Konstantinos Liopyris, Michael Marchetti, et~al.
\newblock Skin lesion analysis toward melanoma detection 2018: A challenge hosted by the international skin imaging collaboration (isic).
\newblock \emph{arXiv preprint arXiv:1902.03368}, 2019.

\bibitem[Yang et~al.(2021)Yang, Zhou, Li, and Liu]{yang2021generalized}
Jingkang Yang, Kaiyang Zhou, Yixuan Li, and Ziwei Liu.
\newblock Generalized out-of-distribution detection: A survey.
\newblock \emph{arXiv preprint arXiv:2110.11334}, 2021.

\bibitem[Ruff et~al.(2021)Ruff, Kauffmann, Vandermeulen, Montavon, Samek, Kloft, Dietterich, and M{\"u}ller]{ruff2021unifying}
Lukas Ruff, Jacob~R Kauffmann, Robert~A Vandermeulen, Gr{\'e}goire Montavon, Wojciech Samek, Marius Kloft, Thomas~G Dietterich, and Klaus-Robert M{\"u}ller.
\newblock A unifying review of deep and shallow anomaly detection.
\newblock \emph{Proceedings of the IEEE}, 109\penalty0 (5):\penalty0 756--795, 2021.

\bibitem[Kong and Ramanan(2021)]{kong2021opengan}
Shu Kong and Deva Ramanan.
\newblock Opengan: Open-set recognition via open data generation.
\newblock In \emph{Proceedings of the IEEE/CVF International Conference on Computer Vision}, pages 813--822, 2021.

\bibitem[Han et~al.(2022)Han, Hu, Huang, Jiang, and Zhao]{han2022adbench}
Songqiao Han, Xiyang Hu, Hailiang Huang, Minqi Jiang, and Yue Zhao.
\newblock Adbench: Anomaly detection benchmark.
\newblock \emph{Advances in Neural Information Processing Systems}, 35:\penalty0 32142--32159, 2022.

\bibitem[Li et~al.(2021)Li, Sohn, Yoon, and Pfister]{li2021cutpaste}
Chun-Liang Li, Kihyuk Sohn, Jinsung Yoon, and Tomas Pfister.
\newblock Cutpaste: Self-supervised learning for anomaly detection and localization.
\newblock In \emph{Proceedings of the IEEE/CVF Conference on Computer Vision and Pattern Recognition}, pages 9664--9674, 2021.

\bibitem[Kirchheim and Ortmeier(2022)]{kirchheim2022outlier}
Konstantin Kirchheim and Frank Ortmeier.
\newblock On outlier exposure with generative models.
\newblock In \emph{NeurIPS ML Safety Workshop}, 2022.

\bibitem[Mirzaei et~al.(2022)Mirzaei, Salehi, Shahabi, Gavves, Snoek, Sabokrou, and Rohban]{mirzaei2022fake}
Hossein Mirzaei, Mohammadreza Salehi, Sajjad Shahabi, Efstratios Gavves, Cees~GM Snoek, Mohammad Sabokrou, and Mohammad~Hossein Rohban.
\newblock Fake it till you make it: Near-distribution novelty detection by score-based generative models.
\newblock \emph{arXiv preprint arXiv:2205.14297}, 2022.

\bibitem[Du et~al.(2023)Du, Sun, Zhu, and Li]{du2023dream}
Xuefeng Du, Yiyou Sun, Xiaojin Zhu, and Yixuan Li.
\newblock Dream the impossible: Outlier imagination with diffusion models.
\newblock \emph{arXiv preprint arXiv:2309.13415}, 2023.

\bibitem[Rombach et~al.(2022)Rombach, Blattmann, Lorenz, Esser, and Ommer]{rombach2022high}
Robin Rombach, Andreas Blattmann, Dominik Lorenz, Patrick Esser, and Bj{\"o}rn Ommer.
\newblock High-resolution image synthesis with latent diffusion models.
\newblock In \emph{Proceedings of the IEEE/CVF Conference on Computer Vision and Pattern Recognition}, pages 10684--10695, 2022.

\bibitem[Schuhmann et~al.(2022)Schuhmann, Beaumont, Vencu, Gordon, Wightman, Cherti, Coombes, Katta, Mullis, Wortsman, et~al.]{schuhmann2022laion}
Christoph Schuhmann, Romain Beaumont, Richard Vencu, Cade Gordon, Ross Wightman, Mehdi Cherti, Theo Coombes, Aarush Katta, Clayton Mullis, Mitchell Wortsman, et~al.
\newblock Laion-5b: An open large-scale dataset for training next generation image-text models.
\newblock \emph{arXiv preprint arXiv:2210.08402}, 2022.

\bibitem[Du et~al.(2022)Du, Wang, Cai, and Li]{du2022vos}
Xuefeng Du, Zhaoning Wang, Mu~Cai, and Yixuan Li.
\newblock Vos: Learning what you don't know by virtual outlier synthesis.
\newblock \emph{arXiv preprint arXiv:2202.01197}, 2022.

\bibitem[Mirzaei et~al.(2024{\natexlab{a}})Mirzaei, Jafari, Dehbashi, Ansari, Ghobadi, Hadi, Moakhar, Azizmalayeri, Baghshah, and Rohban]{mirzaeirodeo}
Hossein Mirzaei, Mohammad Jafari, Hamid~Reza Dehbashi, Ali Ansari, Sepehr Ghobadi, Masoud Hadi, Arshia~Soltani Moakhar, Mohammad Azizmalayeri, Mahdieh~Soleymani Baghshah, and Mohammad~Hossein Rohban.
\newblock Rodeo: Robust outlier detection via exposing adaptive out-of-distribution samples.
\newblock In \emph{Forty-first International Conference on Machine Learning}, 2024{\natexlab{a}}.

\bibitem[Ming et~al.(2022)Ming, Fan, and Li]{ming2022poem}
Yifei Ming, Ying Fan, and Yixuan Li.
\newblock Poem: Out-of-distribution detection with posterior sampling.
\newblock In \emph{International Conference on Machine Learning}, pages 15650--15665. PMLR, 2022.

\bibitem[Golan and El-Yaniv(2018)]{golan2018deep}
Izhak Golan and Ran El-Yaniv.
\newblock Deep anomaly detection using geometric transformations.
\newblock \emph{Advances in neural information processing systems}, 31, 2018.

\bibitem[Hendrycks et~al.(2019{\natexlab{a}})Hendrycks, Mazeika, Kadavath, and Song]{hendrycks2019using}
Dan Hendrycks, Mantas Mazeika, Saurav Kadavath, and Dawn Song.
\newblock Using self-supervised learning can improve model robustness and uncertainty.
\newblock \emph{Advances in Neural Information Processing Systems}, 32, 2019{\natexlab{a}}.

\bibitem[Chen et~al.(2020{\natexlab{c}})Chen, Chen, Zhou, Mao, Li, He, Xue, Zhang, and Yu]{chen2020self}
Kejiang Chen, Yuefeng Chen, Hang Zhou, Xiaofeng Mao, Yuhong Li, Yuan He, Hui Xue, Weiming Zhang, and Nenghai Yu.
\newblock Self-supervised adversarial training.
\newblock In \emph{ICASSP 2020-2020 IEEE International Conference on Acoustics, Speech and Signal Processing (ICASSP)}, pages 2218--2222. IEEE, 2020{\natexlab{c}}.

\bibitem[Hendrycks et~al.(2018)Hendrycks, Mazeika, and Dietterich]{hendrycks2018deep}
Dan Hendrycks, Mantas Mazeika, and Thomas Dietterich.
\newblock Deep anomaly detection with outlier exposure.
\newblock \emph{arXiv preprint arXiv:1812.04606}, 2018.

\bibitem[Tao et~al.(2023)Tao, Du, Zhu, and Li]{tao2023nonparametric}
Leitian Tao, Xuefeng Du, Xiaojin Zhu, and Yixuan Li.
\newblock Non-parametric outlier synthesis, 2023.

\bibitem[Zhang et~al.(2017)Zhang, Cisse, Dauphin, and Lopez-Paz]{zhang2017mixup}
Hongyi Zhang, Moustapha Cisse, Yann~N Dauphin, and David Lopez-Paz.
\newblock mixup: Beyond empirical risk minimization.
\newblock \emph{arXiv preprint arXiv:1710.09412}, 2017.

\bibitem[Noroozi and Favaro(2016)]{10.1007/978-3-319-46466-4_5}
Mehdi Noroozi and Paolo Favaro.
\newblock Unsupervised learning of visual representations by solving jigsaw puzzles.
\newblock In Bastian Leibe, Jiri Matas, Nicu Sebe, and Max Welling, editors, \emph{Computer Vision -- ECCV 2016}, pages 69--84, Cham, 2016. Springer International Publishing.
\newblock ISBN 978-3-319-46466-4.

\bibitem[Zhong et~al.(2020)Zhong, Zheng, Kang, Li, and Yang]{zhong2020random}
Zhun Zhong, Liang Zheng, Guoliang Kang, Shaozi Li, and Yi~Yang.
\newblock Random erasing data augmentation.
\newblock In \emph{Proceedings of the AAAI Conference on Artificial Intelligence (AAAI)}, 2020.

\bibitem[Ghiasi et~al.(2020)Ghiasi, Cui, Srinivas, Qian, Lin, Cubuk, Le, and Zoph]{ghiasi2020simple}
Golnaz Ghiasi, Yin Cui, Aravind Srinivas, Rui Qian, Tsung-Yi Lin, Ekin~D Cubuk, Quoc~V Le, and Barret Zoph.
\newblock Simple copy-paste is a strong data augmentation method for instance segmentation.
\newblock \emph{arXiv preprint arXiv:2012.07177}, 2020.

\bibitem[Akbiyik(2019)]{Akbiyik2019DataAI}
M.~Eren Akbiyik.
\newblock Data augmentation in training cnns: Injecting noise to images.
\newblock \emph{ArXiv}, abs/2307.06855, 2019.
\newblock URL \url{https://api.semanticscholar.org/CorpusID:214522707}.

\bibitem[Hongyi~Zhang(2018)]{zhang2018mixup}
Yann N.~Dauphin Hongyi~Zhang, Moustapha~Cisse.
\newblock mixup: Beyond empirical risk minimization.
\newblock \emph{International Conference on Learning Representations}, 2018.
\newblock URL \url{https://openreview.net/forum?id=r1Ddp1-Rb}.

\bibitem[DeVries and Taylor(2017)]{devries2017cutout}
Terrance DeVries and Graham~W Taylor.
\newblock Improved regularization of convolutional neural networks with cutout.
\newblock \emph{arXiv preprint arXiv:1708.04552}, 2017.

\bibitem[Yun et~al.(2019)Yun, Han, Oh, Chun, Choe, and Yoo]{yun2019cutmix}
Sangdoo Yun, Dongyoon Han, Seong~Joon Oh, Sanghyuk Chun, Junsuk Choe, and Youngjoon Yoo.
\newblock Cutmix: Regularization strategy to train strong classifiers with localizable features, 2019.

\bibitem[Sohn et~al.(2020)Sohn, Li, Yoon, Jin, and Pfister]{sohn2020learning}
Kihyuk Sohn, Chun-Liang Li, Jinsung Yoon, Minho Jin, and Tomas Pfister.
\newblock Learning and evaluating representations for deep one-class classification.
\newblock \emph{arXiv preprint arXiv:2011.02578}, 2020.

\bibitem[Park and Darrell(2020)]{park2020novelty}
Dong~Huk Park and Trevor Darrell.
\newblock Novelty detection with rotated contrastive predictive coding.
\newblock 2020.

\bibitem[de~Haan and L{\"o}we(2021)]{de2021contrastive}
Puck de~Haan and Sindy L{\"o}we.
\newblock Contrastive predictive coding for anomaly detection.
\newblock \emph{arXiv preprint arXiv:2107.07820}, 2021.

\bibitem[Kalantidis et~al.(2020{\natexlab{a}})Kalantidis, Sariyildiz, Pion, Weinzaepfel, and Larlus]{kalantidis2020hard}
Yannis Kalantidis, Mert~Bulent Sariyildiz, Noe Pion, Philippe Weinzaepfel, and Diane Larlus.
\newblock Hard negative mixing for contrastive learning.
\newblock \emph{Advances in Neural Information Processing Systems}, 33:\penalty0 21798--21809, 2020{\natexlab{a}}.

\bibitem[Sinha et~al.(2021)Sinha, Ayush, Song, Uzkent, Jin, and Ermon]{sinha2021negative}
Abhishek Sinha, Kumar Ayush, Jiaming Song, Burak Uzkent, Hongxia Jin, and Stefano Ermon.
\newblock Negative data augmentation.
\newblock In \emph{International Conference on Learning Representations}, 2021.
\newblock URL \url{https://openreview.net/forum?id=Ovp8dvB8IBH}.

\bibitem[Kalantidis et~al.(2020{\natexlab{b}})Kalantidis, Sariyildiz, Pion, Weinzaepfel, and Larlus]{NEURIPS2020_f7cade80}
Yannis Kalantidis, Mert~Bulent Sariyildiz, Noe Pion, Philippe Weinzaepfel, and Diane Larlus.
\newblock Hard negative mixing for contrastive learning.
\newblock In H.~Larochelle, M.~Ranzato, R.~Hadsell, M.F. Balcan, and H.~Lin, editors, \emph{Advances in Neural Information Processing Systems}, volume~33, pages 21798--21809. Curran Associates, Inc., 2020{\natexlab{b}}.
\newblock URL \url{https://proceedings.neurips.cc/paper/2020/file/f7cade80b7cc92b991cf4d2806d6bd78-Paper.pdf}.

\bibitem[Miyai et~al.(2023)Miyai, Yu, Ikami, Irie, and Aizawa]{Miyai_2023_WACV}
Atsuyuki Miyai, Qing Yu, Daiki Ikami, Go~Irie, and Kiyoharu Aizawa.
\newblock Rethinking rotation in self-supervised contrastive learning: Adaptive positive or negative data augmentation.
\newblock In \emph{Proceedings of the IEEE/CVF Winter Conference on Applications of Computer Vision (WACV)}, pages 2809--2818, January 2023.

\bibitem[Zhang et~al.(2024)Zhang, Hua, Sun, Wang, and McLoone]{Zhang_2024_WACV}
Zhaoyu Zhang, Yang Hua, Guanxiong Sun, Hui Wang, and Se\'an McLoone.
\newblock Improving the leaking of augmentations in data-efficient gans via adaptive negative data augmentation.
\newblock In \emph{Proceedings of the IEEE/CVF Winter Conference on Applications of Computer Vision (WACV)}, pages 5412--5421, January 2024.

\bibitem[Chen et~al.(2021{\natexlab{c}})Chen, Xie, Lin, Qiao, Zhou, Tan, Zhang, and Ma]{DBLP:conf/ijcai/ChenXLQZTZM21}
Chengwei Chen, Yuan Xie, Shaohui Lin, Ruizhi Qiao, Jian Zhou, Xin Tan, Yi~Zhang, and Lizhuang Ma.
\newblock Novelty detection via contrastive learning with negative data augmentation.
\newblock In Zhi{-}Hua Zhou, editor, \emph{Proceedings of the Thirtieth International Joint Conference on Artificial Intelligence, {IJCAI} 2021, Virtual Event / Montreal, Canada, 19-27 August 2021}, pages 606--614. ijcai.org, 2021{\natexlab{c}}.
\newblock \doi{10.24963/IJCAI.2021/84}.
\newblock URL \url{https://doi.org/10.24963/ijcai.2021/84}.

\bibitem[Cohen and Avidan(2021)]{cohen2021transformaly}
Matan~Jacob Cohen and Shai Avidan.
\newblock Transformaly--two (feature spaces) are better than one.
\newblock \emph{arXiv preprint arXiv:2112.04185}, 2021.

\bibitem[Schmidt et~al.(2018)Schmidt, Santurkar, Tsipras, Talwar, and Madry]{schmidt2018adversarially}
Ludwig Schmidt, Shibani Santurkar, Dimitris Tsipras, Kunal Talwar, and Aleksander Madry.
\newblock Adversarially robust generalization requires more data.
\newblock \emph{Advances in neural information processing systems}, 31, 2018.

\bibitem[Stutz et~al.(2019)Stutz, Hein, and Schiele]{stutz2019disentangling}
David Stutz, Matthias Hein, and Bernt Schiele.
\newblock Disentangling adversarial robustness and generalization.
\newblock In \emph{Proceedings of the IEEE/CVF Conference on Computer Vision and Pattern Recognition}, pages 6976--6987, 2019.

\bibitem[Roth et~al.(2021)Roth, Pemula, Zepeda, Schölkopf, Brox, and Gehler]{roth2021total}
Karsten Roth, Latha Pemula, Joaquin Zepeda, Bernhard Schölkopf, Thomas Brox, and Peter Gehler.
\newblock Towards total recall in industrial anomaly detection, 2021.

\bibitem[Zavrtanik et~al.(2021)Zavrtanik, Kristan, and Sko{\v{c}}aj]{zavrtanik2021draem}
Vitjan Zavrtanik, Matej Kristan, and Danijel Sko{\v{c}}aj.
\newblock Draem-a discriminatively trained reconstruction embedding for surface anomaly detection.
\newblock In \emph{Proceedings of the IEEE/CVF International Conference on Computer Vision}, pages 8330--8339, 2021.

\bibitem[Mirzaei et~al.(2024{\natexlab{b}})Mirzaei, Jafari, Dehbashi, Taghavi, Sabokrou, and Rohban]{ZARND}
Hossein Mirzaei, Mohammad Jafari, Hamid Dehbashi, Zeinab Taghavi, Mohammad Sabokrou, and Mohammad~Hossein Rohban.
\newblock Killing it with zero-shot: Adversarially robust novelty detection.
\newblock pages 7415--7419, 04 2024{\natexlab{b}}.
\newblock \doi{10.1109/ICASSP48485.2024.10446155}.

\bibitem[Guo et~al.(2019)Guo, Gardner, You, Wilson, and Weinberger]{guo2019simple}
Chuan Guo, Jacob Gardner, Yurong You, Andrew~Gordon Wilson, and Kilian Weinberger.
\newblock Simple black-box adversarial attacks.
\newblock In \emph{International Conference on Machine Learning}, pages 2484--2493. PMLR, 2019.

\bibitem[Goodfellow et~al.(2014)Goodfellow, Shlens, and Szegedy]{goodfellow2014explaining}
Ian~J Goodfellow, Jonathon Shlens, and Christian Szegedy.
\newblock Explaining and harnessing adversarial examples.
\newblock \emph{arXiv preprint arXiv:1412.6572}, 2014.

\bibitem[Mao et~al.(2020)Mao, Chen, Wang, Su, He, and Xue]{Mao2020CompositeAA}
Xiaofeng Mao, Yuefeng Chen, Shuhui Wang, Hang Su, Yuan He, and Hui Xue.
\newblock Composite adversarial attacks.
\newblock \emph{ArXiv}, abs/2012.05434, 2020.
\newblock URL \url{https://api.semanticscholar.org/CorpusID:228083968}.

\bibitem[Zou et~al.(2022)Zou, Jeong, Pemula, Zhang, and Dabeer]{zou2022spot}
Yang Zou, Jongheon Jeong, Latha Pemula, Dongqing Zhang, and Onkar Dabeer.
\newblock Spot-the-difference self-supervised pre-training for anomaly detection and segmentation.
\newblock In \emph{European Conference on Computer Vision}, pages 392--408. Springer, 2022.

\bibitem[Wieler et~al.(2007)Wieler, Hahn, and Hamprecht]{wieler_matthias_2007_8086136}
Matthias Wieler, Tobias Hahn, and Fred~A. Hamprecht.
\newblock {Weakly Supervised Learning for Industrial Optical Inspection}, September 2007.
\newblock URL \url{https://doi.org/10.5281/zenodo.8086136}.
\newblock {Acknowledgements The data was created by Matthias Wieler and Tobias Hahn. This work was conducted at the Robert Bosch Corporate Research department, Schwieberdingen, Germany. We also thank the participants in the contest, the participants of the symposium who helped finance the competition with their registration fees and the GNNS (German Chapter of the European Neural Network Society) for sponsoring the competition.}

\bibitem[Goodfellow et~al.(2013)Goodfellow, Bulatov, Ibarz, Arnoud, and Shet]{goodfellow2013multi}
Ian~J Goodfellow, Yaroslav Bulatov, Julian Ibarz, Sacha Arnoud, and Vinay Shet.
\newblock Multi-digit number recognition from street view imagery using deep convolutional neural networks.
\newblock \emph{arXiv preprint arXiv:1312.6082}, 2013.

\bibitem[Xiao et~al.(2017)Xiao, Rasul, and Vollgraf]{xiao2017fashion}
Han Xiao, Kashif Rasul, and Roland Vollgraf.
\newblock Fashion-mnist: a novel image dataset for benchmarking machine learning algorithms.
\newblock \emph{arXiv preprint arXiv:1708.07747}, 2017.

\bibitem[Tsipras et~al.(2018)Tsipras, Santurkar, Engstrom, Turner, and Madry]{tsipras2018robustness}
Dimitris Tsipras, Shibani Santurkar, Logan Engstrom, Alexander Turner, and Aleksander Madry.
\newblock Robustness may be at odds with accuracy.
\newblock \emph{arXiv preprint arXiv:1805.12152}, 2018.

\bibitem[Yu et~al.(2021)Yu, Zheng, Wang, Li, Wu, Zhao, and Wu]{yu2021fastflow}
Jiawei Yu, Ye~Zheng, Xiang Wang, Wei Li, Yushuang Wu, Rui Zhao, and Liwei Wu.
\newblock Fastflow: Unsupervised anomaly detection and localization via 2d normalizing flows.
\newblock \emph{arXiv preprint arXiv:2111.07677}, 2021.

\bibitem[Mirzaei et~al.(2024{\natexlab{c}})Mirzaei, Nafez, Jafari, Soltani, Azizmalayeri, Habibi, Sabokrou, and Rohban]{mirzaei2024universal}
Hossein Mirzaei, Mojtaba Nafez, Mohammad Jafari, Mohammad~Bagher Soltani, Mohammad Azizmalayeri, Jafar Habibi, Mohammad Sabokrou, and Mohammad~Hossein Rohban.
\newblock Universal novelty detection through adaptive contrastive learning.
\newblock In \emph{Proceedings of the IEEE/CVF Conference on Computer Vision and Pattern Recognition}, pages 22914--22923, 2024{\natexlab{c}}.

\bibitem[Salehi et~al.(2021)Salehi, Mirzaei, Hendrycks, Li, Rohban, and Sabokrou]{salehi2021unified}
Mohammadreza Salehi, Hossein Mirzaei, Dan Hendrycks, Yixuan Li, Mohammad~Hossein Rohban, and Mohammad Sabokrou.
\newblock A unified survey on anomaly, novelty, open-set, and out-of-distribution detection: Solutions and future challenges.
\newblock \emph{arXiv preprint arXiv:2110.14051}, 2021.

\bibitem[Mirzaei and Mathis(2024)]{mirzaei2024adversarially}
Hossein Mirzaei and Mackenzie~W Mathis.
\newblock Adversarially robust out-of-distribution detection using lyapunov-stabilized embeddings.
\newblock \emph{arXiv preprint arXiv:2410.10744}, 2024.

\bibitem[Mirzaei et~al.()Mirzaei, Ansari, Nia, Nafez, Madadi, Rezaee, Taghavi, Maleki, Shamsaie, Hajialilue, et~al.]{mirzaeiscanning}
Hossein Mirzaei, Ali Ansari, Bahar~Dibaei Nia, Mojtaba Nafez, Moein Madadi, Sepehr Rezaee, Zeinab~Sadat Taghavi, Arad Maleki, Kian Shamsaie, Mahdi Hajialilue, et~al.
\newblock Scanning trojaned models using out-of-distribution samples.
\newblock In \emph{The Thirty-eighth Annual Conference on Neural Information Processing Systems}.

\bibitem[Moakhar et~al.(2023)Moakhar, Azizmalayeri, Mirzaei, Manzuri, and Rohban]{moakhar2023seeking}
Arshia~Soltani Moakhar, Mohammad Azizmalayeri, Hossein Mirzaei, Mohammad~Taghi Manzuri, and Mohammad~Hossein Rohban.
\newblock Seeking next layer neurons' attention for error-backpropagation-like training in a multi-agent network framework.
\newblock \emph{arXiv preprint arXiv:2310.09952}, 2023.

\bibitem[Mirzaei et~al.(2024{\natexlab{d}})Mirzaei, Jafari, Dehbashi, Taghavi, Sabokrou, and Rohban]{mirzaei2024killing}
Hossein Mirzaei, Mohammad Jafari, Hamid~Reza Dehbashi, Zeinab~Sadat Taghavi, Mohammad Sabokrou, and Mohammad~Hossein Rohban.
\newblock Killing it with zero-shot: Adversarially robust novelty detection.
\newblock In \emph{ICASSP 2024-2024 IEEE International Conference on Acoustics, Speech and Signal Processing (ICASSP)}, pages 7415--7419. IEEE, 2024{\natexlab{d}}.

\bibitem[Jafari et~al.(2024)Jafari, Zhang, Zhang, and Liu]{jafari2024power}
Mohammad Jafari, Yimeng Zhang, Yihua Zhang, and Sijia Liu.
\newblock The power of few: Accelerating and enhancing data reweighting with coreset selection.
\newblock In \emph{ICASSP 2024-2024 IEEE International Conference on Acoustics, Speech and Signal Processing (ICASSP)}, pages 7100--7104. IEEE, 2024.

\bibitem[Taghavi et~al.(2023{\natexlab{a}})Taghavi, Satvaty, and Sameti]{taghavi2023change}
Zeinab~Sadat Taghavi, Ali Satvaty, and Hossein Sameti.
\newblock A change of heart: Improving speech emotion recognition through speech-to-text modality conversion.
\newblock \emph{arXiv preprint arXiv:2307.11584}, 2023{\natexlab{a}}.

\bibitem[Rahimi et~al.(2024{\natexlab{a}})Rahimi, Amirzadeh, Sohrabi, Taghavi, and Sameti]{rahimi-etal-2024-hallusafe}
Zahra Rahimi, Hamidreza Amirzadeh, Alireza Sohrabi, Zeinab Taghavi, and Hossein Sameti.
\newblock {H}allu{S}afe at {S}em{E}val-2024 task 6: An {NLI}-based approach to make {LLM}s safer by better detecting hallucinations and overgeneration mistakes.
\newblock In Atul~Kr. Ojha, A.~Seza Do{\u{g}}ru{\"o}z, Harish Tayyar~Madabushi, Giovanni Da~San~Martino, Sara Rosenthal, and Aiala Ros{\'a}, editors, \emph{Proceedings of the 18th International Workshop on Semantic Evaluation (SemEval-2024)}, pages 139--147, Mexico City, Mexico, June 2024{\natexlab{a}}. Association for Computational Linguistics.
\newblock \doi{10.18653/v1/2024.semeval-1.22}.
\newblock URL \url{https://aclanthology.org/2024.semeval-1.22/}.

\bibitem[Taghavi et~al.(2023{\natexlab{b}})Taghavi, Gooran, Dalili, Amirzadeh, Nematbakhsh, and Sameti]{taghavi2023imaginations}
Zeinab~Sadat Taghavi, Soroush Gooran, Seyed~Arshan Dalili, Hamidreza Amirzadeh, Mohammad~Jalal Nematbakhsh, and Hossein Sameti.
\newblock Imaginations of wall-e: Reconstructing experiences with an imagination-inspired module for advanced ai systems.
\newblock \emph{arXiv preprint arXiv:2308.10354}, 2023{\natexlab{b}}.

\bibitem[Taghavi et~al.(2023{\natexlab{c}})Taghavi, Naeini, Sadraei~Javaheri, Gooran, Asgari, Rabiee, and Sameti]{taghavi-etal-2023-ebhaam}
Zeinab Taghavi, Parsa~Haghighi Naeini, Mohammad~Ali Sadraei~Javaheri, Soroush Gooran, Ehsaneddin Asgari, Hamid~Reza Rabiee, and Hossein Sameti.
\newblock Ebhaam at {S}em{E}val-2023 task 1: A {CLIP}-based approach for comparing cross-modality and unimodality in visual word sense disambiguation.
\newblock In Atul~Kr. Ojha, A.~Seza Do{\u{g}}ru{\"o}z, Giovanni Da~San~Martino, Harish Tayyar~Madabushi, Ritesh Kumar, and Elisa Sartori, editors, \emph{Proceedings of the 17th International Workshop on Semantic Evaluation (SemEval-2023)}, pages 1960--1964, Toronto, Canada, July 2023{\natexlab{c}}. Association for Computational Linguistics.
\newblock \doi{10.18653/v1/2023.semeval-1.269}.
\newblock URL \url{https://aclanthology.org/2023.semeval-1.269/}.

\bibitem[Taghavi and Mirzaei(2024)]{taghavi2024backdooring}
ZeinabSadat Taghavi and Hossein Mirzaei.
\newblock Backdooring outlier detection methods: A novel attack approach.
\newblock \emph{arXiv preprint arXiv:2412.05010}, 2024.

\bibitem[Ebrahimi et~al.(2024{\natexlab{a}})Ebrahimi, Azari, Iravani, Alizadeh, Taghavi, and Sameti]{ebrahimi2024sharifa}
Seyedeh~Fatemeh Ebrahimi, Karim~Akhavan Azari, Amirmasoud Iravani, Hadi Alizadeh, Zeinab~Sadat Taghavi, and Hossein Sameti.
\newblock Sharif-str at semeval-2024 task 1: Transformer as a regression model for fine-grained scoring of textual semantic relations.
\newblock \emph{arXiv preprint arXiv:2407.12426}, 2024{\natexlab{a}}.

\bibitem[Ebrahimi et~al.(2024{\natexlab{b}})Ebrahimi, Azari, Iravani, Qazvini, Sadeghi, Taghavi, and Sameti]{ebrahimi2024sharif}
Seyedeh~Fatemeh Ebrahimi, Karim~Akhavan Azari, Amirmasoud Iravani, Arian Qazvini, Pouya Sadeghi, Zeinab~Sadat Taghavi, and Hossein Sameti.
\newblock Sharif-mgtd at semeval-2024 task 8: A transformer-based approach to detect machine generated text.
\newblock \emph{arXiv preprint arXiv:2407.11774}, 2024{\natexlab{b}}.

\bibitem[Rahimi et~al.(2024{\natexlab{b}})Rahimi, Shirzady, Taghavi, and Sameti]{rahimi-etal-2024-nimz}
Zahra Rahimi, Mohammad~Moein Shirzady, Zeinab Taghavi, and Hossein Sameti.
\newblock {NIMZ} at {S}em{E}val-2024 task 9: Evaluating methods in solving brainteasers defying commonsense.
\newblock In Atul~Kr. Ojha, A.~Seza Do{\u{g}}ru{\"o}z, Harish Tayyar~Madabushi, Giovanni Da~San~Martino, Sara Rosenthal, and Aiala Ros{\'a}, editors, \emph{Proceedings of the 18th International Workshop on Semantic Evaluation (SemEval-2024)}, pages 148--154, Mexico City, Mexico, June 2024{\natexlab{b}}. Association for Computational Linguistics.
\newblock \doi{10.18653/v1/2024.semeval-1.23}.
\newblock URL \url{https://aclanthology.org/2024.semeval-1.23/}.

\bibitem[Naeem et~al.(2020)Naeem, Oh, Uh, Choi, and Yoo]{ferjad2020icml}
Muhammad~Ferjad Naeem, Seong~Joon Oh, Youngjung Uh, Yunjey Choi, and Jaejun Yoo.
\newblock Reliable fidelity and diversity metrics for generative models.
\newblock 2020.

\bibitem[Hendrycks et~al.(2019{\natexlab{b}})Hendrycks, Mazeika, and Dietterich]{hendrycks2019oe}
Dan Hendrycks, Mantas Mazeika, and Thomas Dietterich.
\newblock Deep anomaly detection with outlier exposure.
\newblock \emph{Proceedings of the International Conference on Learning Representations}, 2019{\natexlab{b}}.

\bibitem[Zhang et~al.(2019)Zhang, Yu, Jiao, Xing, El~Ghaoui, and Jordan]{zhang2019theoretically}
Hongyang Zhang, Yaodong Yu, Jiantao Jiao, Eric Xing, Laurent El~Ghaoui, and Michael Jordan.
\newblock Theoretically principled trade-off between robustness and accuracy.
\newblock In \emph{International conference on machine learning}, pages 7472--7482. PMLR, 2019.

\bibitem[Raghunathan et~al.(2020)Raghunathan, Xie, Yang, Duchi, and Liang]{raghunathan2020understanding}
Aditi Raghunathan, Sang~Michael Xie, Fanny Yang, John Duchi, and Percy Liang.
\newblock Understanding and mitigating the tradeoff between robustness and accuracy.
\newblock \emph{arXiv preprint arXiv:2002.10716}, 2020.

\end{thebibliography}
}
\clearpage
\appendix
\section{Algorithm Block} \label{appendix_algorithm_model}
\begin{algorithm} \label{COBRA_PsudoCode}
\caption{Adversarially Robust Anomaly Detection through Spurious Negative Pair Mitigation
}
\begin{algorithmic}
\fontsize{7.5}{7.5}\selectfont
\State $\mathcal{T}$ $\gets$ $\{\text{Color Jitter, Horizontal Flip, Grayscale, ...}\}$ \Comment{Set of k light augmentations}
    \State $T \gets \{\text{Rotation, Elastic, Distortion, ...} \}$ \Comment{Set of k hard augmentations}
\\
\Function{Classifier\_GMM\_Trainer}{training\_data, $T$}
    \State $synthetic\_data$ $\gets \{ (T_i(\text{training\_data}), i) \mid T_i \in T, i \in \text{range}(T) \}$  \Comment{Create k-class cls dataset}
    \State Train a k-class classifier $C$ on $synthetic\_data$
    \State $e_{train} \gets C(\text{training\_data})$ \Comment{Obtain embeddings from the classifier}
    \State Fit GMM on $e_{train}$ 
    \State $\lambda$ $\gets$ 0.05 
    \State \Return $C$, GMM, $\lambda$
\EndFunction
\vspace{0.7em}
\Function{$\Upsilon$}{$X_{normal}$, $T$, $C$, GMM, $\lambda$}
        \While{$p\_value > \lambda$}
            \State $transforms\_seq \gets$ Sample a random sequence of transforms from $T$
            \State $x_{p-anomaly}$$\gets transforms\_seq$($x_{normal}$)\Comment{Apply hard transformations to create a pseudo-anomaly}
            \State $e_{p-anomaly}$ $\gets$ $C(x_{p-anomaly})$  \Comment{Obtain embeddings from the classifier}
            \State $p\_value$ $\gets$ GMM.P\_Value ($e_{p-anomaly}$) \Comment{Compute P-Value of embeddings}
        \EndWhile
        \State \Return $x_{p-anomaly}$
\EndFunction
\vspace{0.7em}
\Function{Pseudo\_Anomaly\_Generator}{$X_{normal}$, $T$, $C$, GMM, $\lambda$}
    \State $X_{p-anomaly}$ = $\{\}$
    \For{$x_{normal}$ \textbf{in} $X_{normal}$} \Comment{iterate over batch of normal data}
        \State $x_{p-anomaly}$ $\gets$ $\Upsilon$($X_{normal}$, $T$, $C$, GMM, $\lambda$)
    \State $X_{p-anomaly}$.Add($x_{p-anomaly}$)
    \EndFor
    \State \Return $X_{p-anomaly}$ \Comment{Return the generated pseudo-anomaly sample}
\EndFunction
\vspace{0.7em}
\Function{PGD}{x, y, $\mathcal{F}, \mathcal{G}, \mathcal{H}$, pgd\_steps, $\alpha$, $\epsilon$}
    \State $x\_adv \gets x$
    \For{step \textbf{in} pgd\_steps}
        \State grad = Compute\_Gradient( $\left[ \mathcal{L}_{\text{COBRA}}(\mathcal{F}, \mathcal{G}; x\_adv) + \mathcal{L}_{\text{CLS}}(\mathcal{F}, \mathcal{H}; x\_adv, y) \right]$, $x\_adv$) \\ \Comment{Compute the gradient of the loss with respect to the input $x\_adv$}
        \State  $x\_adv = x\_adv + \alpha * sign(grad)$ 
        \State $x\_adv = clip(x\_adv, x - \epsilon, x + \epsilon)$ \Comment{Gradient ascent and projection to valid $\epsilon$-ball}
        \State $x\_adv = clip(x\_adv, 0, 1)$
    \EndFor
    \State \Return $x\_adv$
\EndFunction
\vspace{0.7em}
\Function{Adversarial\_Training\_COBRA}{training\_data, $\mathcal{F}, \mathcal{G}, \mathcal{H}$, $C$, GMM, $\lambda$, pgd\_steps, $\alpha$, $\epsilon$}
     \For{$\mathcal{B}_{normal}$ \textbf{in} training\_data}
        \State $\mathcal{B}_{p-anomaly} \gets$ Pseudo\_Anomaly\_Generator($\mathcal{B}_{normal}$, $T$, $C$, GMM, $\lambda$)
        \State $\mathcal{B} \gets \text{Concatenate}(\mathcal{B}_{normal}, \mathcal{B}_{p-anomaly})$
        \State $Y \gets [0]\times \|\mathcal{B}_{normal}\|+[1]\times \|\mathcal{B}_{p-anomaly}\|$
        \State $\tau_1, \tau_2$ = Sample tow random requence of transforms from $\mathcal{T}$
        \State $\mathcal{B}_{adv} \gets \{\}$
        \For{$x, y$ \textbf{in} $(\mathcal{B}, Y)$}
            \State $x_1$, $x_2$ $\gets \tau_1(x), \tau_2(x)$
            \State P($x_1$), P($x_2$) $\gets$ $\{x_2\}$, $\{x_1\}$
            \State N($x$) $\gets \left\{ \tau_1(x') : x' \in \mathcal{B} \setminus \{x\} \right\} \cup \left\{ \tau_2(x') : x' \in \mathcal{B} \setminus \{x\} \right\}$ \Comment{N($x$)=N($x_1$)=N($x_2$)}
            \State $x\_adv \gets$ PGD(x, y, $\mathcal{F}, \mathcal{G}, \mathcal{H}$, pgd\_steps, $\alpha$, $\epsilon$)
    
            \State $\mathcal{B}_{adv}$.Add($x_{adv}$)
        \EndFor
        \State $\mathcal{L}$ = 0
        \For{$x, x_{adv}, y$ \textbf{in} $(\mathcal{B}, \mathcal{B}_{adv}, Y)$}  
            \State $x_1$, $x_2$ $\gets \tau_1(x), \tau_2(x)$
            \State P($x_1$), P($x_2$), P($x_{adv}$) $\gets \{x_2, x_{adv}\}, \{x_1, x_{adv}\}, \{x_1, x_2\}$ 
            \State N($x$) $\gets \left\{ \tau_1(x') : x' \in \mathcal{B} \setminus \{x\} \right\} \cup \left\{ \tau_2(x') : x' \in \mathcal{B} \setminus \{x\} \right\}$
            \State \hspace{13.1em} $\cup \left\{ x'_{adv} : x'_{adv} \in \mathcal{B}_{adv} \setminus \{x_{adv}\} \right\}$
            \State $\mathcal{L}$ += $\mathcal{L}_{\text{COBRA}}(\mathcal{F}, \mathcal{G}; x ) + \mathcal{L}_{\text{CLS}}(\mathcal{F}, \mathcal{H}; [x, x_{adv}] , [y, y])$            
        \EndFor
        \State Update Networks($\mathcal{F}, \mathcal{G}, \mathcal{H}$) using $\mathcal{L}$
        
    \EndFor
\EndFunction
\vspace{0.7em}
\Function{Main}{epochs, training\_data, $\mathcal{F}, \mathcal{G}, \mathcal{H}$, pgd\_steps, $\alpha$, $\epsilon$}
    \State $C$, GMM, $\lambda$ $\gets$ Classifier\_GMM\_Trainer(training\_data, $T$)
    \For{epoch \textbf{in} epochs}  
        \State Adversarial\_Training\_COBRA(training\_data, $\mathcal{F}, \mathcal{G}, \mathcal{H}$, $C$, GMM, $\lambda$, pgd\_steps, $\alpha$, $\epsilon$)
    \EndFor
\EndFunction
\vspace{1.0em}
\State MAIN(epochs, training\_data, $\mathcal{F}, \mathcal{G}, \mathcal{H}$, 10, $\alpha$, $\epsilon$)

\end{algorithmic}
\end{algorithm}

\section{Related Work}\label{Related_work_appendix}
\noindent\textbf{Previous AD Methods.}\ \ Recent standard AD methods can be categorized into two types: transfer learning based and CL based methods. Transfer learning-based methods utilize a trained model on a large dataset as a backbone and leverage its rich features for the AD task. This approach is evident in methods including PANDA \cite{reiss2021panda}, Transformaly \cite{cohen2021transformaly}, Patchcore \cite{roth2021total}, and Fastflow \cite{yu2021fastflow}. On the other hand, CL framework has demonstrated its superiority by extracting discriminative features, as showcased by CSI \cite{tack2020csi}, MSAD \cite{reiss2021mean}, Recontrast \cite{guo2024recontrast}, and Draem \cite{zavrtanik2021draem}. Extending transfer learning based methods to an adversarial setting is not feasible because pre-trained features, which act as a key, are not robust, necessitating a new training paradigm. This limitation inspired us to adopt CL for adversarial training, which also aligns with the unlabeled nature of AD.  There have been a few efforts to propose robust AD methods, including PrincipaLS \cite{lo2022adversarially}, OCSDF \cite{bethune2023robust}, and APAE \cite{goodge2021robustness}. However, their results on even tiny datasets are less than random detection. Details about each mentioned method can be found in Appendix \ref{Appendix_Details_Related_Work}.

\section{Details of Related Work} \label{Appendix_Details_Related_Work}
\subsection{Previous AD methods}
There has been some efforts to develope a robust AD method. PrincipaLS  employs a novel latent space manipulation technique to adjust the representations of data point  with optimizing a robustness criterion designed to minimize the model's sensitivity to adversarial perturbations. OCSDF leverages the Signed Distance Function to delineate the boundary of a data distribution. Through the employment of 1-Lipschitz neural networks, it adeptly approximates normality scores, thus enhancing robustness to  adversarial perturbations. APAE introduces the  approximate projection autoencoder   as a defense mechanism, integrating gradient descent on latent embeddings and feature-weighting normalization to enhance detection robustness. it worth noting we exclude robust OOD detection from our experiments, where their method have been develpoed by relying on labels and could not extend to AD setup. ZARND a introduces a robust method enhancing anomaly detection by integrating robust features from pretrained models with nearest-neighbor algorithms. The approach significantly improves robustness against adversarial attacks, which traditionally degrade ND performance. By leveraging features from adversarially robust models and employing k-Nearest Neighbors (k-NN) for anomaly scoring \cite{mirzaei2022fake,mirzaei2024universal,salehi2021unified,mirzaei2024adversarially,mirzaeiscanning,moakhar2023seeking,mirzaei2024killing,jafari2024power,taghavi2023change,rahimi-etal-2024-hallusafe,taghavi2023imaginations,taghavi-etal-2023-ebhaam,taghavi2024backdooring,ebrahimi2024sharifa,ebrahimi2024sharif,rahimi-etal-2024-nimz}.

\subsection{Auxiliary Anomaly Sample Crafting}
Previous studies on anomaly generation often struggle with producing samples that are either too similar to normal instances (distant anomalies) or inadvertently create samples that still belong to the normal category. Conversely, many studies have underscored the benefit of using related auxiliary anomaly samples to enhance detector performance. In light of this, we have proposed COBRA, which, unlike its counterparts such as Dream-OOD, does not require an extra dataset and performs well in an unsupervised setting where labels for normal samples are not available.

COBRA aligns with the distribution of the normal dataset, effectively generating informative pseudo-anomalies within the pixel space. It employs a thresholding technique to sift out inaccuracies, all while obviating the need for supplementary datasets.

\section{Distribution Aware Hard Transformation}
\label{Appendix_Distribution_Aware_Hard_Transformation}

In this section, we elaborate on the practical implementation details of our proposed pseudo-anomaly crafting strategy. As discussed in the main text of the paper, we set the hyperparameter $\beta$ to $0.05$. To assess the robustness of our model with respect to this parameter, we conducted extensive experiments over a broader range, specifically $[0.02, 0.20]$. The stability of our model's performance across this range is demonstrated in the experimental results (refer to Table~\ref{tab:appendix_beta_ablation}). These findings underscore the robustness of our approach in handling varying hyperparameter settings without significant performance degradation.

\begin{table}[h]
\centering
\small 
\centering
\caption{Performance of the model for different values of the hyperparameter $\beta$ used in pseudo-anomaly crafting. Results are shown in each table cell as 'Clean/\graytext{PGD-1000}'. (Evaluations are done on CIFAR-10 and CIFAR-100 datasets under a PGD-1000 attack with $\epsilon = \frac{4}{255}$, and on other high-resolution datasets with $\epsilon = \frac{2}{255}$).}
\label{tab:appendix_beta_ablation}
\renewcommand{\arraystretch}{0.5} 
    \resizebox{1\linewidth}{!}{\begin{tabular}{*{8}{C{1.9cm}}} 

    \specialrule{1.5pt}{\aboverulesep}{\belowrulesep}

    \multicolumn{1}{c}{\textbf{\(\beta\)}} &  \multicolumn{7}{c}{Datasets} \\
    
    \cmidrule(lr){1-1}\cmidrule(lr){2-8}
    
     & cifar10 &  cifar100  & MVTecAD  & CityScapes & VisA & ISIC2018 & DAGM \\
    \specialrule{1.5pt}{\aboverulesep}{\belowrulesep}

    0.02 & \graytext{82.7 /} 61.8 & \graytext{75.4 /} 52.3 & \graytext{88.0 /} 74.1 & \graytext{79.9 /} 57.5 & \graytext{73.7 /} 70.1 & \graytext{82.3 /} 55.6 & \graytext{80.4 /} 55.1 \\
    \cmidrule(lr){1-1} \cmidrule(lr){2-8}

    0.05 & \graytext{83.7 /} 62.3 & \graytext{76.9 /} 51.7 & \graytext{89.1 /} 75.1 & \graytext{81.7 /} 56.2 & \graytext{75.2 /} 73.8 & \graytext{81.3 /} 56.1 & \graytext{82.4 /} 56.8 \\
    \cmidrule(lr){1-1} \cmidrule(lr){2-8}

    0.08 & \graytext{82.3 /} 62.1 & \graytext{74.2 /} 50.8 & \graytext{87.3 /} 73.9 & \graytext{81.3 /} 55.9 & \graytext{76.1 /} 73.0 & \graytext{81.0 /} 57.0 & \graytext{82.1 /} 55.7 \\
    \cmidrule(lr){1-1} \cmidrule(lr){2-8}

    0.11 & \graytext{81.9 /} 61.3 & \graytext{74.8 /} 51.0 & \graytext{88.1 /} 74.4 & \graytext{80.8 /} 55.1 & \graytext{74.2 /} 72.7 & \graytext{80.6 /} 55.2 & \graytext{81.3 /} 57.0 \\
    \cmidrule(lr){1-1} \cmidrule(lr){2-8}

    0.14 & \graytext{80.8 /} 60.9 & \graytext{77.2 /} 51.8 & \graytext{89.0 /} 74.5 & \graytext{80.6 /} 56.2 & \graytext{75.8 /} 72.6 & \graytext{79.8 /} 56.4 & \graytext{79.5 /} 54.6 \\
    \cmidrule(lr){1-1} \cmidrule(lr){2-8}

    0.17 & \graytext{79.5 /} 59.3 & \graytext{75.8 /} 50.4 & \graytext{86.9 /} 75.0 & \graytext{79.6 /} 56.7 & \graytext{74.2 /} 72.9 & \graytext{79.3 /} 54.1 & \graytext{79.0 /} 53.7 \\
    \cmidrule(lr){1-1} \cmidrule(lr){2-8}

    0.20 & \graytext{80.1 /} 58.9 & \graytext{76.3 /} 49.9 & \graytext{89.2 /} 74.2 & \graytext{80.0 /} 54.2 & \graytext{73.9 /} 72.1 & \graytext{80.3 /} 55.3 & \graytext{81.9 /} 54.7 \\

    \specialrule{1.5pt}{\aboverulesep}{\belowrulesep}
\end{tabular}}
\end{table}

\subsection{Hard Transformation}
\label{Appendix_Hard_Transformation}

We employed a series of hard transformations to create the $T$ set. The transformations and their respective hyperparameters are as follows:

\begin{itemize}
    \item \textbf{Jigsaw}: Images were divided into a 2x2 grid, and the tiles were randomly permuted. This transformation disrupts the spatial continuity of image features.
    \item \textbf{Random Erasing}: A random rectangular region in the image was erased, with a size proportional to the image area and an aspect ratio randomly chosen between 10\% and 50\%.
    \item \textbf{CutPaste}: A square region, with side length varying from 10\% to 50\% of the image width, was cut and pasted into a different location within the same image.
    \item \textbf{Rotation}: Images were rotated by a random angle within $\pm$90 degrees to introduce a moderate level of distortion.
    \item \textbf{Extreme Blurring}: Applied a Gaussian blur with a kernel size of up to 5\% of the image width and a high variance ($\sigma=2.5$), resulting in significant blurring.
    \item \textbf{Intense Random Cropping}: Random sections, sized between 50\% and 80\% of the original image size, were cropped to challenge the model with incomplete patterns.
    \item \textbf{Noise Injection}: Gaussian noise with a mean of 0 and a standard deviation of 0.1 was added to the images.
    \item \textbf{Extreme Cropping}: Cropped the image to retain only 40\% to 60\% of its original size, focusing on the center to ensure a significant deviation from the original distribution.
    \item \textbf{Mixup}: To create composite images that merge features from both source images in a more challenging manner, we combined pairs of images using a blending coefficient $\alpha$, drawn from a Beta distribution Beta($\alpha$, $\alpha$) with $\alpha$ set to 0.1. This lower value of $\alpha$ results in a higher variance of the mixing coefficients, producing images that are significantly more blended than with a higher $\alpha$ value. This aggressive mixing ensures the crafting of pseudo-anomaly samples.
    \item \textbf{Cutout}: Square regions with a side length of 25\% of the image width were filled with a constant value to simulate occlusion.
    \item \textbf{CutMix}: Portions from one image were cut and pasted onto another, with the cut region's size approximately 20\% of the image area.
\end{itemize}

\noindent\textbf{Quality of Generated Abnormal Data.}\ \ \label{appendix_Quality_of_Generated_Abnormal}
In Table \ref{tab:pseudo_anomaly_strategy} of our paper, we evaluated our crafting strategy by comparing it with alternative methods. Additionally, Figure \ref{fig:different-anomaly-generator-data-comparison} presents samples that visually compare our results with others. To address any concerns further, we employed the FID metric, which measures the distance from a normal distribution, as well as the DC metric \cite{ferjad2020icml} to assess diversity. A lower FID indicates that the crafted anomalies are more similar to normal instances, while a higher DC suggests greater diversity among the samples. Table \ref{DC_FID_Compare} compares the quality of anomalies generated by our method with those from other alternatives, using the FID and DC metrics.
\begin{table}[h]
\centering
\caption{Assessment of Generated Anomaly Data Quality Compared to Alternative Methods}
\resizebox{0.8\linewidth}{!}{\begin{tabular}{lcccccc}
\toprule
\textbf{Dataset} & \textbf{Metric} & \textbf{GOE} & \textbf{FITYMI} & \textbf{Dream-OOD} & \textbf{Dream-OOD} & \textbf{Ours} \\
\midrule
\multirow{2}{*}{MVTecAD} & FID ↓ & 245 & 163 & 227 & 227 & 129 \\
 & DC ↑ & 0.53 & 0.29 & 0.75 & 0.75 & 0.84 \\
\midrule
\multirow{2}{*}{ImageNet} & FID ↓ & 118 & 145 & 98 & 98 & 107 \\
 & DC ↑ & 0.64 & 0.38 & 0.87 & 0.87 & 0.58 \\
\midrule
\multirow{2}{*}{CIFAR10} & FID ↓ & 76 & 86 & 72 & 72 & 54 \\
 & DC ↑ & 0.45 & 0.36 & 0.56 & 0.56 & 0.72 \\
\midrule
\multirow{2}{*}{FMNIST} & FID ↓ & 42 & 37 & 64 & 64 & 32 \\
 & DC ↑ & 0.31 & 0.17 & 0.45 & 0.45 & 0.63 \\
\midrule
\multirow{2}{*}{\textbf{Average}} & FID ↓ & 120.3 & 107.8 & 115.3 & 115.3 & \textbf{80.3} \\
 & DC ↑ & 0.48 & 0.3 & 0.65 & 0.65 & \textbf{0.69} \\
\bottomrule
\end{tabular}}
\label{DC_FID_Compare}
\end{table}

\subsection{Positive Transformation}
Consistent with the self-supervised learning literature \cite{chen2020simple,he2020momentum}, we employ mild transformations that generate samples with minimal visual differences while fully preserving semantics. These include \textit{color jitter, random grayscale conversion, and random cropping (cropping the image to retain 80\% to 100\% of its original size)}. These transformations are designed to subtly alter the appearance of images without changing their underlying content, thereby enabling the model to learn robust features that are invariant to minor perturbations.

\section{Additional Ablation Study} \label{Appendix_Additional_Ablation}

Here, we conduct further experiments to thoroughly evaluate COBRA across diverse settings, highlighting its key components and their substantial impact. \\ \\
 
\noindent \textbf{Clean Training}
In this scenario, we skipped adversarial training and instead trained COBRA with standard training, keeping all other components fixed. The results indicate that clean performance increased from \textbf{84.1} to \textbf{90.7}, demonstrating COBRA's superiority in various scenarios of training and evaluating. The results are presented in Table \ref{fig:clean_training_cobra}.
\\ \\ 

\begin{table}[!t]
\caption{Performance of COBRA trained with standard training (without adversarial training) across various datasets.}
\resizebox{\linewidth}{!}{\begin{tabular}{*{11}{C{1.8cm}}} 
    \hline\noalign{\smallskip}
    \multicolumn{1}{c}{Evaluation setting} &\multicolumn{10}{c}{Datasets} 
    \\ \cmidrule(lr){2-11}   
    & CIFAR10 & CIFAR100 & MNIST & FMnist & SVHN & \scriptsize ImagenNet30 & MVTechAD & VisA & DAGM & ISIC2018 \\
    \cmidrule(lr){1-1}\cmidrule(lr){2-11}\noalign{\smallskip}
    Clean & 94.3 & 93.7 & 95.1 & 93.4 & 96.0 & 92.7 & 95.2 & 78.1 & 90.8 & 83.4 \\
    \cmidrule(lr){1-1}\cmidrule(lr){2-11}\noalign{\smallskip}
    Adversarial & 1.7 & 4.6 & 1.0 & 3.8 & 2.5 & 4.9 & 5.8 & 2.3 & 4.8 & 4.2 \\
\noalign{\smallskip}\hline
\end{tabular}}
\label{fig:clean_training_cobra}
\end{table}

\noindent\textbf{Anomaly Score}
Instead of utilizing our proposed Anomaly Score, we explored alternative approaches while keeping other components unchanged. Specifically, our default anomaly score is based on the similarity between a test sample and normal training samples. We substituted this anomaly score with logits provided by our binary classifier head. Specifically, the COBRA results reported in the main paper are based on the \(A(x)\) anomaly score, which is defined as:

\[
A(x) = -\max_{x^i \in D_{\text{train}}} \left\{ \langle \mathcal{G}(\mathcal{F}(x)), \mathcal{G}(\mathcal{F}(x^i)) \rangle \right\},
\]

Instead, we replaced that with \(A'(x) = p(\mathcal{H}(\mathcal{F}(x))|y=1)\), which denotes the probability of belonging to the pseudo-anomaly class as assigned by the binary classifier head. Additionally, we considered \(A(x)+A'(x)\) as another alternative. As the results presented in Table \ref{tab:score_function_ablation_} demonstrate, all strategies achieve significant performance with minor differences.
\begin{table}[!h]
    \centering
    \small
    \centering
 	\caption{Ablation study on different score functions. Results are AUROC (\%)}
    \renewcommand{\arraystretch}{0.5} 
    \resizebox{0.7\linewidth}{!}{\begin{tabular}{l*{4}{C{2.5cm}}} 
        \toprule
        \multicolumn{1}{c}{Dataset} & \multicolumn{3}{c}{Anomaly Score } \\
        \cmidrule(lr){2-4}

        & $A(x)+A'(x)$ & $A'(x)$  & $A(x)$  (default) \\
        
        \noalign{\smallskip}\hline\noalign{\smallskip}

        MVTecAD & \graytext{88.4} / 74.6 & \graytext{83.1} / 72.8 & \graytext{89.1} / 75.1 \\
        
        \cmidrule(lr){1-1}\cmidrule(lr){2-4} 

         CIFAR10 & \graytext{81.9} / 60.2 & \graytext{76.1} / 55.7 &  \graytext{ 83.7} / 62.3   \\
        
        \cmidrule(lr){1-1}\cmidrule(lr){2-4}

        FMNIST & \graytext{89.6} / 87.4 & \graytext{88.5} / 84.8 &  \graytext{ 93.1 } / 89.6  \\
        
        \cmidrule(lr){1-1}\cmidrule(lr){2-4}

        CIFAR100 & \graytext{72.6} / 51.2 & \graytext{70.4} / 48.3 &  \graytext{ 76.9 } / 51.7 \\
        
        \bottomrule
	\end{tabular}}
	\label{tab:score_function_ablation_}
\end{table}


        
   
			
			
			

			
			
			

\noindent\textbf{Ablation Study on Distribution-Aware Hard Transformation}\\
In this section, we utilize the feature extractor \(C\) as an anomaly detector to demonstrate that its learned features from the normal training set are meaningful and significantly surpass random detection, highlighting its role in crafting effective pseudo-anomaly samples. Results presented in Table \ref{table:C_cobra}. \\

\begin{table}[!t]
\caption{\(C\)'s performance in the AD task across various datasets.}
\resizebox{\linewidth}{!}{\begin{tabular}{*{11}{C{1.8cm}}} 
    \hline\noalign{\smallskip}
    \multicolumn{1}{c}{Methods} &\multicolumn{10}{c}{Datasets} 
    \\ \cmidrule(lr){2-11}   
    & CIFAR10 & CIFAR100 & MNIST & FMnist & SVHN & \scriptsize ImagenNet30 & MVTechAD & VisA & DAGM & ISIC2018 \\
    \cmidrule(lr){1-1}\cmidrule(lr){2-11}\noalign{\smallskip}
    COBRA & 77.1 & 72.4 & 86.0 & 83.6 & 74.7 & 80.3 & 75.2 & 64.9 & 78.5 & 73.2 \\
\noalign{\smallskip}\hline
\end{tabular}}
\label{table:C_cobra}
\end{table}

\noindent\textbf{Ablation Study on Thresholding}\\
Here, while keeping all components of COBRA constant, we skip the thresholding strategy and instead use a random subset of hard transformations for crafting pseudo-anomaly samples. The results, indicated in Table \ref{fig:no_threshold_training}, suggest that this approach leads to decreased performance due to the failure to filter incorrect pseudo-anomaly samples (those that still belong to the normal set).
\begin{table}[!h]
\caption{ Impact of Skipping Thresholding on Pseudo-Anomaly Generation in COBRA}
\resizebox{\linewidth}{!}{\begin{tabular}{*{11}{C{1.8cm}}} 
    \hline\noalign{\smallskip}
    \multicolumn{1}{c}{Methods} &\multicolumn{10}{c}{Datasets} 
    \\ \cmidrule(lr){2-11}   
    & CIFAR10 & CIFAR100 & MNIST & FMnist & SVHN & \scriptsize ImagenNet30 & MVTechAD & VisA & DAGM & ISIC2018 \\
    \cmidrule(lr){1-1}\cmidrule(lr){2-11}\noalign{\smallskip}
    COBRA &  \graytext{78.6} / 58.1 & \graytext{65.4} / 46.5 & \graytext{94.3} / 91.5 & \graytext{90.1} / 86.5 & \graytext{84.8} / 53.2 & \graytext{72.5} / 55.7 & \graytext{76.0} / 68.2 & \graytext{78.4} / 47.9 & \graytext{74.6} / 52.0 & \graytext{73.8} / 51.4 \\
\noalign{\smallskip}\hline
\end{tabular}}
\label{fig:no_threshold_training}
\end{table}

\section{Analyzing COBRA's Stability and Effectiveness} \label{Appendix_COBRA_Loss_Stability}
\begin{figure}[h]
  \centering
  \begin{minipage}{\textwidth}
  \centering
  \subfloat[Loss]{\includegraphics[width=0.49\linewidth]{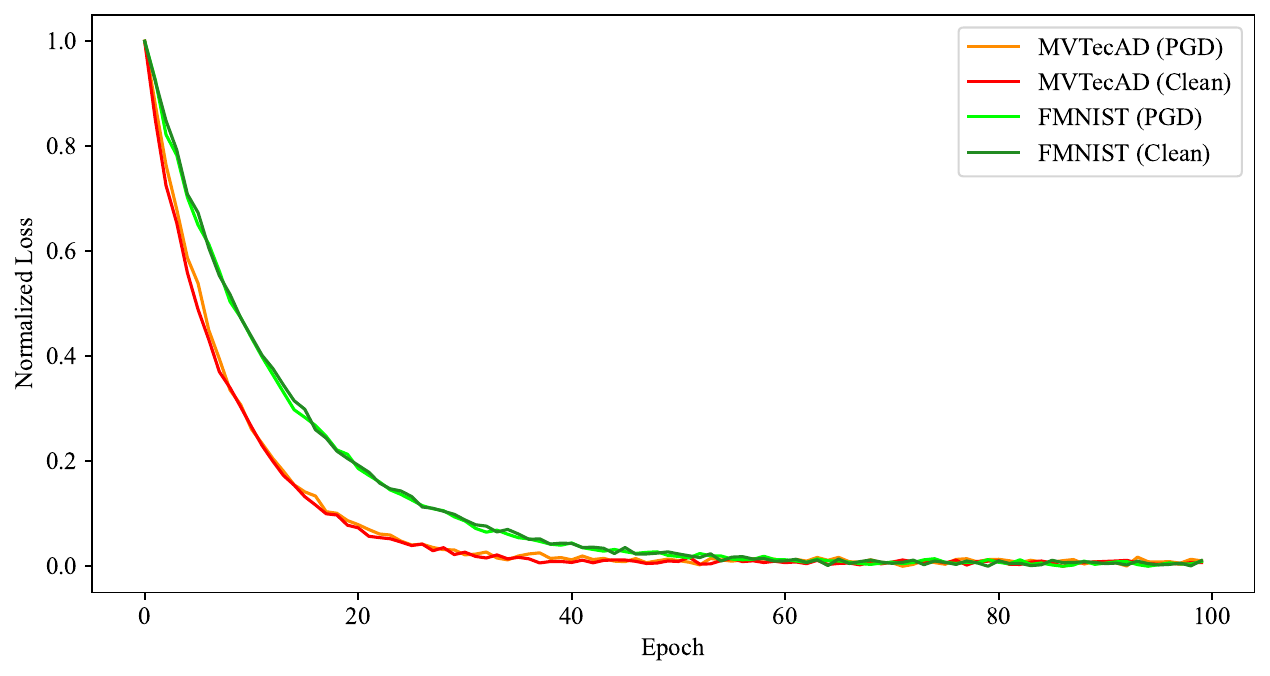}\vspace{5pt}\label{fig:loss_cobra_appendix}}
  \subfloat[AUC]{\includegraphics[width=0.49\linewidth]{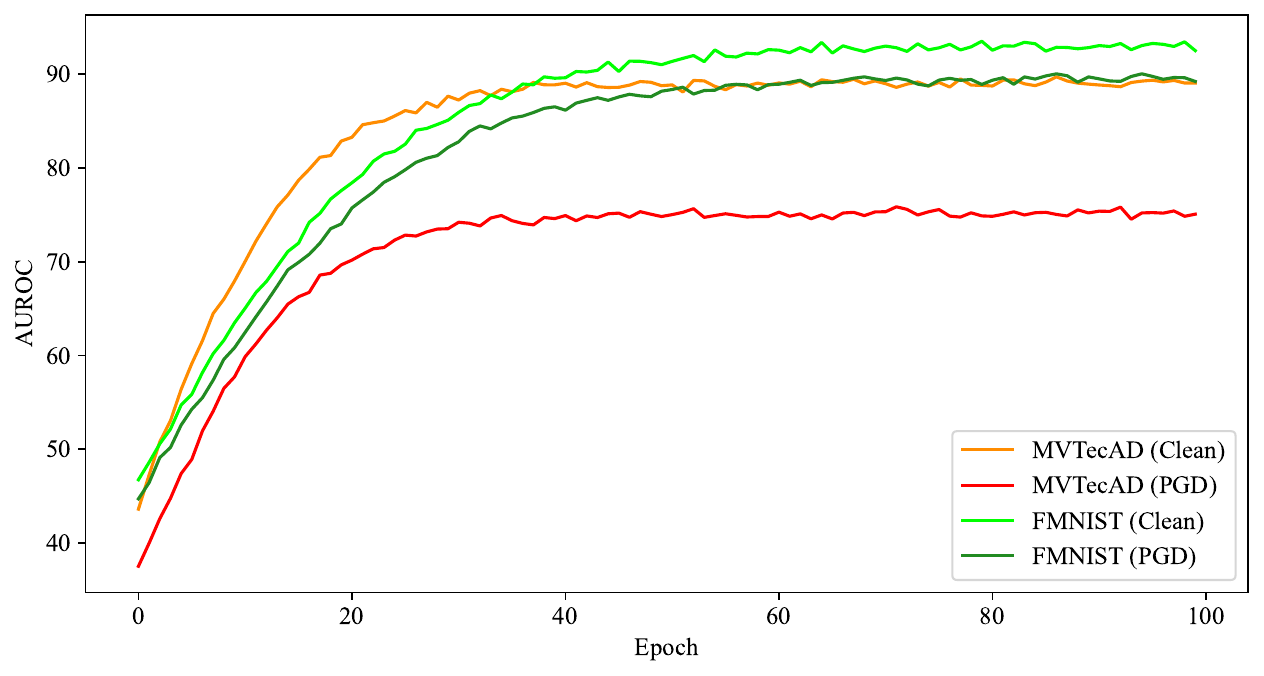}\vspace{5pt}\label{fig:auc_cobra_appendix}}
  \caption{This figure depicts COBRA's loss values and detection performance for each epoch of training on both clean data and data under PGD attack, demonstrating the stability of COBRA's loss function. The experiment was carried out in a one-class setup using the MVETEC-AD and FMNIST datasets.}
  \label{fig:loss_auc_cobra_appendix}
  \end{minipage}
\end{figure}

Figure \ref{fig:loss_auc_cobra_appendix} represents COBRA's loss values and its detection performance at each epoch of training for both clean data and data subjected to PGD attack, demonstrating the stability of COBRA's loss function. The experiment was conducted in a one-class setup using the MVETEC-AD and FMNIST datasets. Note that the loss values have been normalized between 0 and 1, and the loss values for the PGD data are higher than those for the clean data. Despite this, the figure underscores the stability of our loss function, which remains consistent across different training conditions.


\section{$A^3$ to All Methods}\label{Appendix_AutoAttack}
\begin{table*}[!htbp]
\caption{ AUROC (\%) of various methods for One Class AD methods under AutoAttack.  }
\footnotesize{$^*$These models are trained to be adversarially robust.}

\resizebox{\linewidth}{!}{\begin{tabular}{l*{10}{C{1.9cm}}} 

\hline\noalign{\smallskip}			
        
    \multicolumn{1}{c}{Dataset} &\multicolumn{10}{c}{Method } \\
    
    \cmidrule(lr){1-1}\cmidrule(lr){2-11}  
        
    &\multirow{2}{*}{DeepSVDD}&\multirow{2}{*}{CSI}&\multirow{2}{*}{DN2}&\multirow{2}{*}{PANDA}&\multirow{2}{*}{MSAD} &\multirow{2}{*}{Transformaly}&\multirow{2}{*}{PatchCore} & \multirow{2}{*}{PrincipaLS$^*$} & \multirow{2}{*}{OCSDF$^*$} &\multirow{2}{*}{APAE$^*$}  \\
     & & & &  & &  &   &   &  & \\
        
    \noalign{\smallskip}\hline\noalign{\smallskip}
        
    CIFAR10 &  16.6 & 3.3 & 2.5 & 1.2 & 0.7 & 2.4 & 2.5 &  29.6  & 26.9 & 2.0  \\
    
    \cmidrule(lr){1-1} \cmidrule(lr){2-11}\noalign{\smallskip}
        
    CIFAR100 &  14.3 & 1.2 & 0.7 & 1.1 & 10.7 & 7.3 & 3.5 &  24.6  & 19.8 & 0.9 \\

    \cmidrule(lr){1-1} \cmidrule(lr){2-11}\noalign{\smallskip}
        
    MNIST&  15.4 & 1.7 & 1.0 & 0.7 & 14.1 & 11.6 & 2.4 &  77.3  & 66.2 & 28.6  \\
    
    \cmidrule(lr){1-1} \cmidrule(lr){2-11}\noalign{\smallskip}
    
    Fashion-MNIST  & 48.1 & 8.5 & 2.2 & 9.7 & 3.7 & 2.8 & 2.8 &  67.5  & 59.6 & 12.3 \\
    \cmidrule(lr){1-1} \cmidrule(lr){2-11}\noalign{\smallskip}
  
    SVHN   &4.3 & 0.8& 0.2&  0.4&  0.1&0.2 &1.1&  8.9 &  6.1&  0.3 \\    
    \cmidrule(lr){1-1} \cmidrule(lr){2-11}\noalign{\smallskip}
    
    ImageNet30  & 12.4& 2.7& 1.2& 0.5& 1.4& 1.8& 2.3 &  23.8 & 20.6& 1.1 \\
    
    \noalign{\smallskip}\hline\noalign{\smallskip}
\end{tabular}}
\label{Table1:Novelty_Detection_AutoAttack}
\end{table*}

Here, we present more detailed results of previous anomaly detection (AD) works and their performance under $A^3$  (See Table \ref{Table1:Novelty_Detection_AutoAttack}). Their respective performance against PGD-1000 attacks has been provided in the main paper.

\section{evaluating our model under various attacks with diverse epsilon values}\label{appendix:linf_diverse_epsilon}
In this section, we evaluate our model's performance under adversarial attacks, specifically focusing on $\ell_\infty$ and $\ell_2$ PGD attacks across various epsilon values. Table \ref{linf_varied_epsilon_table} presents the results for the $\ell_\infty$ PGD-1000 attacks, indicating that the model maintains strong accuracy for lower epsilon values, with some decline in performance observed as epsilon increases. Similarly, Table \ref{l2_varied_epsilon_table} illustrates the model's robustness under $\ell_2$ PGD attacks, demonstrating consistent accuracy across lower epsilon values, with a gradual trend observed at higher epsilon values. These results highlight the model's resilience to adversarial attacks while also indicating areas for potential improvement.

\begin{table}[h]  
\centering
\caption{Evaluation of our model under various $\ell_{\infty}$ PGD-1000 attacks across different $\epsilon$ values. Although the model was trained using $\ell_{\infty}$ PGD-10 with $\epsilon = 2/255$ for high-resolution images and $\epsilon = 4/255$ for low-resolution images, the evaluation settings were modified to test its robustness against stronger attacks.}
\label{linf_varied_epsilon_table}
\resizebox{\linewidth}{!}{\begin{tabular}{lcccccccc} 
\toprule
\textbf{Epsilon} & \textbf{MVTec} & \textbf{VisA} & \textbf{ImageNet} & \textbf{CityScapes} & \textbf{ISIC2018} & \textbf{CIFAR10} & \textbf{FMNIST} & \textbf{Avg.} \\
\midrule
\centering 0       & 89.1 & 75.2 & 85.2 & 81.7 & 81.3 & 83.7 & 93.1 & 84.2 \\
1/255   & 81.6 & 74.6 & 68.1 & 65.7 & 62.8 & 77.0 & 91.1 & 74.4 \\
2/255   & 75.1 & 73.8 & 57.0 & 56.2 & 56.1 & 72.9 & 90.2 & 68.7 \\
3/255   & 64.7 & 67.2 & 54.1 & 50.9 & 51.8 & 69.3 & 89.9 & 64.0 \\
4/255   & 56.1 & 52.9 & 45.2 & 47.1 & 49.3 & 62.3 & 89.6 & 57.5 \\
5/255   & 49.5 & 46.3 & 43.7 & 40.9 & 45.5 & 60.8 & 89.2 & 53.7 \\
6/255   & 42.8 & 40.7 & 39.4 & 38.6 & 42.1 & 56.3 & 89.2 & 49.9 \\
7/255   & 38.1 & 33.9 & 34.5 & 35.2 & 39.6 & 51.8 & 89.1 & 46.0 \\
8/255   & 34.2 & 30.7 & 28.6 & 32.8 & 35.5 & 47.4 & 89.1 & 42.6 \\
\bottomrule
\end{tabular}}
\end{table}

\begin{table}[h]
\centering
\caption{Evaluation of our model under various $\ell_2$ PGD attacks across different $\epsilon$ values. Although the model was trained using $\ell_{\infty}$ PGD-10 with $\epsilon = 2/255$ for high-resolution images and $\epsilon = 4/255$ for low-resolution images, the evaluation settings were modified to test its robustness under stronger $\ell_2$-norm adversarial attacks.}
\label{l2_varied_epsilon_table}
\resizebox{\linewidth}{!}{\begin{tabular}{lcccccccc} 
\toprule
\textbf{Epsilon} & \textbf{MVTec} & \textbf{VisA} & \textbf{ImageNet} & \textbf{CityScapes} & \textbf{ISIC2018} & \textbf{CIFAR10} & \textbf{FMNIST} & \textbf{Avg.} \\
\midrule
0      & 89.1 & 75.2 & 85.2 & 81.7 & 81.3 & 83.7 & 93.1 & 84.2 \\
16/255   & 87.1 & 74.9 & 77.8 & 77.9 & 76.8 & 73.8 & 91.0 & 79.9 \\
32/255   & 85.4 & 74.8 & 75.1 & 77.8 & 74.2 & 64.1 & 89.9 & 77.3 \\
64/255   & 83.8 & 74.2 & 71.0 & 72.6 & 71.3 & 55.4 & 88.9 & 73.9 \\
128/256    & 79.6 & 73.1 & 65.0 & 66.7 & 60.5 & 48.9 & 87.6 & 68.8 \\
\bottomrule
\end{tabular}}
\end{table}

\section{Implementation Details}\label{Appendix_Implementation_Details}
We employ ResNet-18 as the foundational encoder network ($f_\theta$), accompanied by an auxiliary head ($g_\phi$) consisting of a 2-layer multi-layer perceptron with a 128-dimensional embedding. For optimization, COBRA is trained for 100 epochs using the LARS optimizer, with a weight decay of $1 \times 10^{-6}$ and a momentum of 0.9.
To schedule the learning rate, we adopt a linear warmup for the initial 10 epochs, gradually increasing the learning rate to 1.0. Subsequently, we use a cosine decay schedule without restarts. The batch size for COBRA is set to 128. Our experiments were conducted using NVIDIA GeForce RTX 3090 GPUs (24GB).

\begin{figure}[h]
  \centering
  \includegraphics[width=1.0\linewidth]{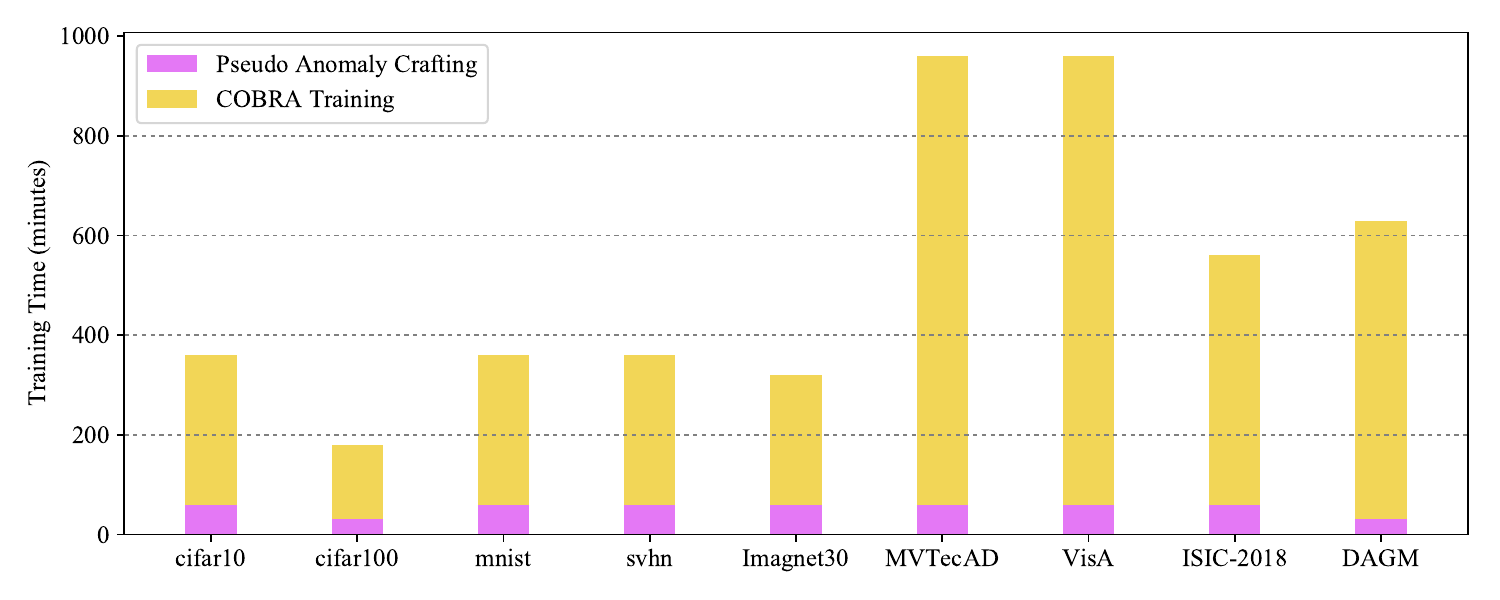}
  \caption{Computation cost}
  \label{fig:computation_cost_fig}
\end{figure}

\noindent\textbf{Training Computational Cost.}\ \ COBRA comprises two main steps: (i) generating pseudo-anomaly samples from the normal training set, and (ii) adversarially training a model using both normal and crafted pseudo-anomalies. In this section, we analyze the complexity of these steps. Figure \ref{fig:computation_cost_fig} illustrates that COBRA achieves significant performance with low complexity in terms of time. Please note that the training time shown in the figure is calculated for a single class of the dataset.\\
In Table \ref{tab:training_time}, we compare the time efficiency of our method, both with and without adversarial training, against other state-of-the-art methods across multiple datasets.

\begin{table}[h]
\centering
\caption{Training Time Comparison (in hours) Across Different Methods and Datasets}
\resizebox{0.8\linewidth}{!}{\begin{tabular}{lcccc}
\toprule
\textbf{Method} & \textbf{MVTecAD} & \textbf{VisA} & \textbf{CityScape} & \textbf{ImageNet} \\
\midrule
Transformaly & 30h & 70h & 6h & 200h \\
ReContrast & 25h & 60h & 5h & 180h \\
Ours (with adv. training) & 95h & 250h & 20h & 500h \\
Ours (without adv. training) & 9h & 23h & 2h & 70h \\
\bottomrule
\end{tabular}}
\label{tab:training_time}
\end{table}

\noindent\textbf{Evaluation of Computational Cost.}\ \ After training, we freeze our model and extract features from the training samples to create an embedding bank, which maps images from high-dimensional spaces (e.g., (1000, 3, 224, 224)) to a lower-dimensional space (e.g., (1000, 256)). During inference, for each test sample, we compute its features using the frozen model and compare its similarity to the precomputed embedding bank. This process is computationally efficient since it operates with a frozen model and in a low-dimensional space. To the best of our knowledge, using similarity-based methods, such as k-NN in embedding spaces for detection tasks, is common and well-established in the literature (e.g., MSAD). Furthermore, we provide detailed computational time information, excluding the time required for feature bank creation as it can be precomputed. The anomaly score computation times on a 3090 GPU, in comparison to other methods, are presented in Table \ref{tab:evaluation_time}, with results averaged over 100,000 inferences. For our model, embedding extraction takes 1-3 ms, matrix multiplication 1-2 ms, and finding the maximum value takes less than 1 ms.

\begin{table}[h]
\centering
\caption{Per-Image Evaluation Time Comparison (in milliseconds) Across Different Methods}
\resizebox{0.8\linewidth}{!}{\begin{tabular}{lcccccc}
\toprule
\textbf{Metric} & \textbf{CSI} & \textbf{MSAD} & \textbf{Transformaly} & \textbf{ReContrast} & \textbf{OCSDF} & \textbf{COBRA (Ours)} \\
\midrule
\textbf{Per-Image Eval Time} & 4 ms & 4 ms & 5 ms & 5 ms & 3 ms & 4 ms \\
\bottomrule
\end{tabular}}
\label{tab:evaluation_time}
\end{table}

\section{Per-Class Results}\label{Appendix_Per_Class_Results}
Anomaly detection   evaluation scenarios can be categorized into one-class anomaly detection and unlabeled multi-class setups, as mentioned in the experiments section. In this section, we provide details of the one-class classification setup for COBRA on the reported dataset, as presented in Tables \ref{per_class_1}, \ref{per_class_2}.

\begin{table*}[h]
\caption{The detailed AUROC scores of the class-wise experiments for One-Class Anomaly Detection setting with PGD-1000 $\epsilon = \frac{4}{255}$ in CIFAR10, CIFAR100, MNIST, Fashion-MNIST, SVHN datasets.}
\label{per_class_1}

\begin{subtable}{1\textwidth}
\subcaption{ MNIST }
\label{}
\resizebox{\linewidth}{!}{\begin{tabular}{llr*{13}{C{1.2cm}}} 
\hline\noalign{\smallskip}

   \multicolumn{1}{c}{Method} & \multicolumn{1}{c}{Attack} & &\multicolumn{9}{c}{Class } &\multicolumn{1}{c}{ Average }  
     \\      \cmidrule(lr){1-1} \cmidrule(lr){2-2} 
    \cmidrule(lr){3-12}  
    \cmidrule(lr){13-13}   
   
  & &\multirow{1}{*}{0}&\multirow{1}{*}{1}&\multirow{1}{*}{2}&\multirow{1}{*}{3}&\multirow{1}{*}{4} &\multirow{1}{*}{5}&\multirow{1}{*}{6} & \multirow{1}{*}{7} & \multirow{1}{*}{8} &\multirow{1}{*}{9}  \\
  
  \cmidrule(lr){1-1} \cmidrule(lr){2-2} 
  \cmidrule(lr){3-12}\cmidrule(lr){13-13}

\noalign{\smallskip}
\multirow{4}{*}{Ours} 
 &Clean & 97.6 & 39.9 & 99.6 & 99.2 & 98.9 & 98.3 & 99.6 & 95.6 & 99.7 & 99.6 & 92.8
\\ & BlackBox& 92.4 & 99.6 & 98.9 & 95.1 & 97.5 & 94.6 & 99.1 & 96.9 & 96 & 97.2 & 96.8
\\ & PGD-100 & 92.4 & 99.3 & 98.3 & 95.1 & 96.9 & 94.4 & 98.7 & 96.7 & 96 & 96.6 & 96.4
\\ & $A^3$ & 91.7 & 98.3 & 97.8 & 94.3 & 94.5 & 94 & 97.3 & 94.3 & 95.1 & 94.4 & 95.2\\

\noalign{\smallskip}

\hline
\end{tabular}}
\end{subtable}

 \bigskip

\begin{subtable}{1\textwidth}
\subcaption{ Fashion-MNIST }
\label{}
\resizebox{\linewidth}{!}{\begin{tabular}{llr*{13}{C{1.2cm}}} 
\hline\noalign{\smallskip}

   \multicolumn{1}{c}{Method} & \multicolumn{1}{c}{Attack} & &\multicolumn{9}{c}{Class } &\multicolumn{1}{c}{ Average }  
     \\      \cmidrule(lr){1-1} \cmidrule(lr){2-2} 
    \cmidrule(lr){3-12}  
    \cmidrule(lr){13-13}   
   
  & &\multirow{1}{*}{0}&\multirow{1}{*}{1}&\multirow{1}{*}{2}&\multirow{1}{*}{3}&\multirow{1}{*}{4} &\multirow{1}{*}{5}&\multirow{1}{*}{6} & \multirow{1}{*}{7} & \multirow{1}{*}{8} &\multirow{1}{*}{9}  \\
  
  \cmidrule(lr){1-1} \cmidrule(lr){2-2} 
  \cmidrule(lr){3-12}\cmidrule(lr){13-13}

\noalign{\smallskip}
\multirow{4}{*}{ COBRA } 
&Clean & 89.5 & 99.6 & 92.7 & 91.4 & 86.7 & 97.1 & 82.5 & 97 & 96.2 & 98.4 & 93.1
\\ & BlackBox & 88.7 & 99.5 & 85.8 & 85.7 & 83.3 & 95.3 & 82.6 & 97.9 & 92.6 & 97.8 & 90.9
 \\ & PGD-100 & 85.4 & 99 & 85.1 & 84.6 & 83 & 95.3 & 80.9 & 95.1 & 90.7 & 97.4 & 89.6
\\ & $A^3$  & 83.7 & 96.5 & 85 & 83.9 & 81.7 & 93.5 & 77.2 & 91.5 & 87 & 93.6 & 87.4\\
\noalign{\smallskip}
\hline
\end{tabular}}
\end{subtable}

\bigskip

\begin{subtable}{1\textwidth}
\subcaption{ CIFAR10 }
\label{}
\resizebox{\linewidth}{!}{\begin{tabular}{llr*{13}{C{1.2cm}}} 
\hline\noalign{\smallskip}

   \multicolumn{1}{c}{Method} & \multicolumn{1}{c}{Attack} & &\multicolumn{9}{c}{Class } &\multicolumn{1}{c}{ Average }  
     \\      \cmidrule(lr){1-1} \cmidrule(lr){2-2} 
    \cmidrule(lr){3-12}  
    \cmidrule(lr){13-13}   
   
  & &\multirow{1}{*}{0}&\multirow{1}{*}{1}&\multirow{1}{*}{2}&\multirow{1}{*}{3}&\multirow{1}{*}{4} &\multirow{1}{*}{5}&\multirow{1}{*}{6} & \multirow{1}{*}{7} & \multirow{1}{*}{8} &\multirow{1}{*}{9}  \\
  
  \cmidrule(lr){1-1} \cmidrule(lr){2-2} 
  \cmidrule(lr){3-12}\cmidrule(lr){13-13}

\noalign{\smallskip}
\multirow{4}{*}{Ours} &Clean& 79.4 & 96.3 & 75.4 & 71.3 & 76 & 84.3 & 76.1 & 94.5 & 92.8 & 91.1 & 83.7
\\ & BlackBox & 75.9 & 96.1 & 75.2 & 68.4 & 73.4 & 84 & 74.2 & 92 & 89.9 & 89.8 & 81.8
\\ & PGD-100& 64.7 & 80.4 & 47.9 & 39.9 & 49.5 & 60.7 & 52.9 & 74.8 & 80.3 & 71.7 & 62.3
\\ & $A^3$ & 64.5 & 77.7 & 46.6 & 39.8 & 48.8 & 58.3 & 50.1 & 72.9 & 80.3 & 68.7 & 60.7 \\
\noalign{\smallskip}

\hline
\end{tabular}}
\end{subtable}

  \bigskip

\begin{subtable}{1\textwidth}
\subcaption{SVHN}
\label{}
\resizebox{\linewidth}{!}{\begin{tabular}{llr*{13}{C{1.2cm}}}
\hline\noalign{\smallskip}
   
   \multicolumn{1}{c}{Method} & \multicolumn{1}{c}{Attack} & &\multicolumn{9}{c}{Class } &\multicolumn{1}{c}{ Average }  
     \\      \cmidrule(lr){1-1} \cmidrule(lr){2-2} 
    \cmidrule(lr){3-12}  
    \cmidrule(lr){13-13}   
   
  & &\multirow{1}{*}{0}&\multirow{1}{*}{1}&\multirow{1}{*}{2}&\multirow{1}{*}{3}&\multirow{1}{*}{4} &\multirow{1}{*}{5}&\multirow{1}{*}{6} & \multirow{1}{*}{7} & \multirow{1}{*}{8} &\multirow{1}{*}{9}  \\
  
  \cmidrule(lr){1-1} \cmidrule(lr){2-2} 
  \cmidrule(lr){3-12}\cmidrule(lr){13-13}

\noalign{\smallskip}
\multirow{4}{*}{Ours} 
&Clean& 793.7 & 89.6 & 88.8 & 79.3 & 88.3 & 88.9 & 89.8 & 93.1 & 90.4 & 90.9 & 89.3
\\ & BlackBox & 88.8 & 88.9 & 85.7 & 75.2 & 83.7 & 84.1 & 86.9 & 90.5 & 89.8 & 89.9 & 86.4
\\ & PGD-100&61.2 & 61.8 & 54.5 & 45.4 & 59.4 & 55.2 & 61 & 67.7 & 54.4 & 61.4 & 58.2
\\ & $A^3$ & 60.1 & 61.6 & 52.4 & 43.6 & 56.6 & 55.6 & 59.7 & 65.3 & 52.4 & 58.7 & 56.7
  \\

\noalign{\smallskip}

\hline
\end{tabular}}
\end{subtable}

 \bigskip

\begin{subtable}{1\textwidth}
\subcaption{ CIFAR100 }
\label{}
\resizebox{\linewidth}{!}{\begin{tabular}{llr*{23}{C{0.9cm}}} 
\hline\noalign{\smallskip}

   \multicolumn{1}{c}{Method} & \multicolumn{1}{c}{Attack} & &\multicolumn{19}{c}{Class } &\multicolumn{1}{c}{ Average }  
     \\      \cmidrule(lr){1-1} \cmidrule(lr){2-2} 
    \cmidrule(lr){3-22}  
    \cmidrule(lr){23-23}   

  & &\multirow{1}{*}{0}&\multirow{1}{*}{1}&\multirow{1}{*}{2}&\multirow{1}{*}{3}&\multirow{1}{*}{4} &\multirow{1}{*}{5}&\multirow{1}{*}{6} & \multirow{1}{*}{7} & \multirow{1}{*}{8} &\multirow{1}{*}{9} &\multirow{1}{*}{10} &\multirow{1}{*}{11}&\multirow{1}{*}{12}&\multirow{1}{*}{13}&\multirow{1}{*}{14} &\multirow{1}{*}{15}&\multirow{1}{*}{16} & \multirow{1}{*}{17} & \multirow{1}{*}{18} &\multirow{1}{*}{19}  \\
  
  \cmidrule(lr){1-1} \cmidrule(lr){2-2} 
  \cmidrule(lr){3-22}\cmidrule(lr){23-23}

\noalign{\smallskip}
\multirow{4}{*}{Ours} &Clean & 68.9 & 73.1 & 79.4 & 71.9  & 84.24 & 70.1 & 79.5 & 72.3 & 71.7 & 85.4 & 83.34& 78.8 & 80 & 57.5 & 80.4 & 69.6 & 62.5 & 92.4 & 88.9 & 88.5 & 76.9
\\ & BlackBox & 65.6 & 72.7 & 79 & 70.2 & 80.7 & 68.2 & 75.4 & 72.1 & 71.3 & 84.1 & 79.7 & 76.5 & 78.6 & 54.4 & 76.7 & 65.4 & 60.7 & 91.3 & 85 & 87.4 & 74.6
\\ & PGD-100 & 43.6 & 51.2 & 47.8 & 51.2 & 57.2 & 44.5 & 55.3 & 38.5 & 40.2 & 70.4 & 72.4 & 50.2 & 48.7 & 29.6 & 50.7 & 39.9 & 36 & 75.5 & 63.5 & 69.3 & 51.7
\\ & $A^3$  & 
 42.6 & 48.1 & 47.4 & 51.2 & 56.6 & 42.9 & 52.9 & 35.8 & 40.1 & 69.5 & 72.4 & 48.3 & 47.3 & 27.3 & 49.3 & 37.4 & 32.4 & 72.5 & 62.3 & 66.7 & 60.7
  \\
\noalign{\smallskip}
\hline
\end{tabular}}
\end{subtable}

\end{table*}

\begin{table*}[h]
\caption{The detailed AUROC scores of the class-specific experiments for One-Class Anomaly Detection setting with  PGD-1000 $\epsilon = \frac{4}{255}$ in Imagenet30 dataset.}
\label{per_class_2}

\begin{subtable}{1\textwidth}
\subcaption{ ImageNet30 }
\label{}
\resizebox{\linewidth}{!}{\begin{tabular}{llr*{13}{C{1.2cm}}} 
\hline\noalign{\smallskip}

   \multicolumn{1}{c}{Method} & \multicolumn{1}{c}{Attack} & &\multicolumn{9}{c}{Class } &\multicolumn{1}{c}{ Average }  
     \\      \cmidrule(lr){1-1} \cmidrule(lr){2-2} 
    \cmidrule(lr){3-12}  
    \cmidrule(lr){13-13}   
   
  & &\multirow{1}{*}{0}&\multirow{1}{*}{1}&\multirow{1}{*}{2}&\multirow{1}{*}{3}&\multirow{1}{*}{4} &\multirow{1}{*}{5}&\multirow{1}{*}{6} & \multirow{1}{*}{7} & \multirow{1}{*}{8} &\multirow{1}{*}{9}  \\
  
  \cmidrule(lr){1-1} \cmidrule(lr){2-2} 
  \cmidrule(lr){3-12}\cmidrule(lr){13-13}

\noalign{\smallskip}
\multirow{4}{*}{Ours} 
 &Clean & 74.9 & 98.1 & 99.7 & 73.5 & 83.4 & 96.7 & 93.2 & 83.3 & 82.7& 74 & 85.2
\\ & BlackBox& 72.3 & 93.3 & 95.6 & 71.1 & 80.1 & 96 & 89.1 & 82.2 & 78.8 & 70.3 & 82.4
\\ & PGD-100 & 35.8 & 90.9 & 96.7 & 46.4 & 36.3 & 86.3 & 60.9 & 55.7 & 38.3 & 24.4 & 57
\\ & $A^3$ &  34.9 & 89.8 & 94.6 & 44.2 & 32.7 & 82.6 & 56.4 & 52.6 & 35.8 & 22.6 & 54.7 \\

\noalign{\smallskip}

\hline
\end{tabular}}
\end{subtable}

\bigskip

\begin{subtable}{1\textwidth}
\subcaption{ ImageNet30 }
\label{}
\resizebox{\linewidth}{!}{\begin{tabular}{llr*{13}{C{1.2cm}}} 
\hline\noalign{\smallskip}

   \multicolumn{1}{c}{Method} & \multicolumn{1}{c}{Attack} & &\multicolumn{9}{c}{Class } &\multicolumn{1}{c}{ Average }  
     \\      \cmidrule(lr){1-1} \cmidrule(lr){2-2} 
    \cmidrule(lr){3-12}  
    \cmidrule(lr){13-13}   
   
  & &\multirow{1}{*}{10}&\multirow{1}{*}{11}&\multirow{1}{*}{12}&\multirow{1}{*}{13}&\multirow{1}{*}{14} &\multirow{1}{*}{15}&\multirow{1}{*}{16} & \multirow{1}{*}{17} & \multirow{1}{*}{18} &\multirow{1}{*}{19}  \\
  
  \cmidrule(lr){1-1} \cmidrule(lr){2-2} 
  \cmidrule(lr){3-12}\cmidrule(lr){13-13}

\noalign{\smallskip}
\multirow{4}{*}{Ours} 
 &Clean &96.9 & 90.1 & 93.3 & 82.1 & 94.6 & 62.8 & 98.2 & 58.5 & 89.4 & 54.7 & 85.2
\\ & BlackBox& 96 & 87.1 & 91.8 & 78.3 & 90.3 & 62.5 & 95.4 & 55.5 & 88.7 & 51.9 & 82.4
\\ & PGD-100 & 71.3 & 62.8 & 75.5 & 36.6 & 76.6 & 16.4 & 88.1 & 17 & 58.5 & 24 & 57
\\ & $A^3$ & 67.1 & 61.6 & 70.9 & 35.7 & 72.6 & 13.8 & 87.1 & 15.9 & 55.7 & 22.2 & 54.7\\

\noalign{\smallskip}

\hline
\end{tabular}}
\end{subtable}

\bigskip

\begin{subtable}{1\textwidth}
\subcaption{ ImageNet30 }
\label{}
\resizebox{\linewidth}{!}{\begin{tabular}{llr*{13}{C{1.2cm}}} 
\hline\noalign{\smallskip}

   \multicolumn{1}{c}{Method} & \multicolumn{1}{c}{Attack} & &\multicolumn{9}{c}{Class } &\multicolumn{1}{c}{ Average }  
     \\      \cmidrule(lr){1-1} \cmidrule(lr){2-2} 
    \cmidrule(lr){3-12}  
    \cmidrule(lr){13-13}   
   
  & &\multirow{1}{*}{20}&\multirow{1}{*}{21}&\multirow{1}{*}{22}&\multirow{1}{*}{23}&\multirow{1}{*}{24} &\multirow{1}{*}{25}&\multirow{1}{*}{26} & \multirow{1}{*}{27} & \multirow{1}{*}{28} &\multirow{1}{*}{29}  \\
  
  \cmidrule(lr){1-1} \cmidrule(lr){2-2} 
  \cmidrule(lr){3-12}\cmidrule(lr){13-13}

\noalign{\smallskip}
\multirow{4}{*}{Ours} 
 &Clean & 95 & 88.1 & 97.2 & 96.5 & 80.2 & 69.2 & 83.5 & 92.9 & 78.3 & 93.7 & 85.2
\\ & BlackBox& 92.3 & 83.9 & 91.5 & 96.3 & 75.1 & 65.3 & 82.4 & 89.9 & 76.7 & 90.8 & 82.4
\\ & PGD-100 & 77.9 & 44.8 & 90.8 & 79.1 & 39.2 & 40.6 & 32.2 & 78.1 & 41.3 & 86.5 & 57
\\ & $A^3$ & 74.7 & 44.9 & 89.8 & 75.2 & 35.7 & 38.4 & 31.8 & 77.3 & 38.8 & 84.4 & 54.7\\

\noalign{\smallskip}

\hline
\end{tabular}}
\end{subtable}

\end{table*}

.
\clearpage  

\section{Adversarial Attacks Adaptation}\label{Appendix_Adv__Adaptation}
We evaluated COBRA's robustness against a variety of powerful attacks, including BlackBox, FGSM, CAA, AutoAttack, $A^{3}$, and PGD-1000. These attacks, originally designed to compromise classification tasks by exploiting the cross-entropy loss, were adapted for anomaly detection (AD) tasks, focusing on the anomaly scores of detector models. The aim was to generate perturbations that increase the anomaly score for normal test samples and decrease it for anomalous ones. As discussed in the preliminaries section, adapting AutoAttack (AA) \cite{croce2020reliable} for AD tasks was particularly challenging. AutoAttack is an ensemble of different attack methods, such as FAB, multi-targeted FAB, Square Attack, APGDT, APGD with cross-entropy loss, and APGD with DLR loss. The main challenge in adaptation arises because attacks based on DLR loss assume the model's output includes at least three elements, an assumption valid for classification tasks on datasets with three or more classes but not applicable to AD tasks. Consequently, we replaced the DLR loss component in AutoAttack with a PGD attack. However, for the other attacks under consideration, no adjustments were necessary.

\section{Datasets Details}\label{Appendix_Dataset_Details}
The MVTecAD is an industrial defect detection dataset used to evaluate AD methods. It consists of 4,096 normal and 1,258 anomaly samples, encompassing various types of texture defects. MVTecAD is under the CC-BY-NC-SA 4.0 license. VisA is another challenging dataset for industrial defect detection, comprising 9,621 normal and 1,200 anomaly samples. VisA is under the CC-BY 4.0 license. DAGM is a synthetic dataset created for defect detection on textured surfaces. To broaden the scope beyond traditional industrial scenarios, the Cityscapes dataset offers stereo videos from 50 cities, each meticulously annotated for 30 classes such as roads and buildings. We leverage this dataset by extracting 256x256 patches from its images to construct an anomaly detection dataset, focusing on the presence of anomaly objects within these patches. Anomaly classes encompass motorcycles, persons, riders, traffic signs, traffic lights, and bicycles, while other classes from Cityscapes are considered normal. Their code is released under the MIT license. Additionally, ISIC2018 is a skin disease dataset, available as task 3 of the ISIC2018 challenge. It contains seven classes. NV (nevus) is taken as the normal class, and the rest of the classes are taken as anomalies, following. The training set contains 6,705 normal images. The ISIC dataset is available under CC-BY-NC license. 
Furthermore, we utilize ImageNet30 \cite{hendrycks2019using}, an anomaly detection benchmark that selects 30 classes from ImageNet and employs a one-versus-rest setup for anomaly detection. This dataset is freely available to researchers for non-commercial use.

\section{Detailed Results} \label{appendix:detailed results}
The figure presents the standard deviation and mean of COBRA (a specific algorithm or method) calculated over 5 separate runs for each experiment. This comprehensive data collection and analysis underscore the reliability and consistency of our experimental results. By reporting both the mean and standard deviation, we provide a clear depiction of the average performance and the variability, ensuring that the performance of COBRA is not only robust but also consistently reproducible across multiple trials.

\begin{table*}[h]
\caption{The standard deviation and mean of COBRA across 5 runs of each experiment are reported, demonstrating the consistency of our results.}
\label{Table 1:Novelty Detection}
\resizebox{ \linewidth}{!}{\begin{tabular}{ll*{10}{c}} 
\specialrule{1.5pt}{\aboverulesep}{\belowrulesep}

\multicolumn{1}{c}{Statistics} & \multicolumn{1}{c}{Eval Type} &\multicolumn{10}{c}{Datasets }  \\
\cmidrule(lr){1-1} \cmidrule(lr){2-2}\cmidrule(lr){3-12}  
			
& & \multirow{2}{*}{CIFAR10} & \multirow{2}{*}{CIFAR100} & \multirow{2}{*}{MNIST} & \multirow{2}{*}{FMNIST} & \multirow{2}{*}{SVHN} &\multirow{2}{*}{ImageNet} & \multirow{2}{*}{VisA} & \multirow{2}{*}{CityScapes} & \multirow{2}{*}{DAGM} & \multirow{2}{*}{ISIC2018} \\

\noalign{\vskip 3pt}
\specialrule{1.5pt}{\aboverulesep}{\belowrulesep}
\noalign{\vskip 3pt}

\multirow{2}{*}{Mean $\pm$ STD} 
& Clean & 83.7 $\pm$ 0.52 & 76.9 $\pm$ 0.81 & 92.8 $\pm$ 0.47 & 93.1 $\pm$ 0.56 & 89.3 $\pm$ 0.39 & 85.2 $\pm$ 1.21 & 75.2 $\pm$ 1.03 & 81.7 $\pm$ 0.97 & 82.4 $\pm$ 0.87 & 81.3 $\pm$ 0.74 \\

& Adv & 62.3 $\pm$ 0.73 & 51.7 $\pm$ 0.91 & 96.4$\pm$ 0.86 & 89.6 $\pm$ 0.91 & 58.2 $\pm$ 0.62 & 57.0 $\pm$ 1.41 & 73.8 $\pm$ 1.17 & 56.2 $\pm$ 1.31 & 56.8 $\pm$ & 56.1 $\pm$ 1.42\\

\specialrule{1.5pt}{\aboverulesep}{\belowrulesep}

\end{tabular}}
\end{table*}

\section{Supplementary Metrics for COBRA Assessment} \label{appendix_Supplementary_Metrics}
We found the Area Under the Receiver Operating Characteristic curve (AUROC) to be the most widely accepted and utilized metric. To further our exploration, we have provided additional results using two supplementary metrics—AUPR and FPR95\%—which have been utilized in some previous works \cite{hendrycks2019oe}. In the table \ref{Metrics_Evaluation}, we compare COBRA against TRANSFORMALY and ZARND, a recent detection method, using these metrics. FPR95\% represents the false positive rate when 95\% of the outliers are correctly detected; a lower FPR95\% signifies better performance. Both AUROC and AUPR summarize a detection method's performance over various thresholds. Specifically, AUROC indicates the likelihood that an outlier is ranked higher in anomaly score compared to an in-distribution sample. Hence, higher AUROC and AUPR values denote superior performance, with an uninformative detector scoring an AUROC of 50\%. To address the reviewer's concerns, we will consider including AUPR and FPR95\% along with AUROC in our final manuscript.

\begin{table}[h]
\centering
\caption{Methods and Metrics Comparison Across Different Datasets}
\resizebox{\linewidth}{!}{\begin{tabular}{lcccccccc}
\toprule
\textbf{Methods} & \textbf{Metric} & \textbf{MVTec} & \textbf{VisA} & \textbf{ImageNet} & \textbf{CityScapes} & \textbf{ISIC2018} & \textbf{CIFAR10} & \textbf{FMNIST} \\ 
\midrule
\multirow{3}{*}{Transformaly} & AUROC ↑ & 88.5/2.2 & 85.5/0.0 & 99.0/2.9 & 87.4/4.5 & 86.6/3.9 & 98.3/3.7 & 94.4/7.4 \\ 
 & AUPR ↑ & 85.6/3.9 & 83.7/4.1 & 94.1/2.3 & 86.0/2.6 & 83.5/2.8 & 99.6/1.4 & 98.2/0.0 \\ 
 & FPR95\% ↓ & 41.6/99.7 & 26.7/98.9 & 2.6/99.1 & 35.7/97.9 & 28.4/98.0 & 8.4/99.6 & 9.5/95.4 \\ 
\midrule
\multirow{3}{*}{ZARND} & AUROC ↑ & 71.6/30.1 & 71.8/24.9 & 96.4/27.4 & 75.9/28.6 & 70.2/14.6 & 89.7/56.0 & 95.0/82.3 \\ 
 & AUPR ↑ & 73.5/28.7 & 69.8/20.1 & 90.1/29.8 & 72.1/26.5 & 73.6/16.4 & 85.5/52.3 & 92.5/80.0 \\ 
 & FPR95\% ↓ & 41.2/69.2 & 44.1/73.9 & 7.3/64.8 & 40.1/66.4 & 26.8/75.3 & 26.4/55.9 & 9.6/23.3 \\ 
\midrule
\multirow{3}{*}{COBRA (Ours)} & AUROC ↑ & 89.1/75.1 & 75.2/73.8 & 85.2/57.0 & 81.7/56.2 & 81.3/56.1 & 83.7/62.3 & 93.1/89.6 \\ 
 & AUPR ↑ & 91.7/71.9 & 78.6/70.3 & 88.9/61.2 & 85.6/62.1 & 87.6/59.8 & 84.7/65.7 & 95.1/89.8 \\ 
 & FPR95\% ↓ & 18.5/36.8 & 35.8/38.7 & 24.9/54.7 & 29.3/55.0 & 30.1/52.7 & 26.4/43.9 & 6.7/17.3 \\ 
\bottomrule
\end{tabular}}
\label{Metrics_Evaluation}
\end{table}

\section{Social Impacts}\label{appendix:social impacts}
Robust anomaly detection is crucial across many safety-critical domains like security, healthcare, finance, and manufacturing to identify potential threats, diseases, fraud or system faults before they cause harm. However, existing machine learning-based anomaly detectors are vulnerable to adversarial attacks that can make them miss anomalies or falsely flag normal data.
Our work on COBRA presents a significant step towards developing reliable and robust anomaly detection systems resilient to adversarial conditions. By learning representations inherently robust to input perturbations and distribution shifts, COBRA enables safer anomaly detection deployment in security-sensitive areas where an adversary may attempt evasion. Notably, COBRA achieves high robustness without requiring anomaly data during training, valuable when such data is limited due to privacy/safety concerns.
Overall, we believe COBRA importantly enhances the safety and reliability of anomaly detection systems. However, this powerful technology must be responsibly developed and deployed with technical safeguards, policy measures and institutional controls to maximize societal benefit while mitigating potential misuse risks. We advocate future work exploring robust machine learning trustworthy real-world deployment.

\section{Limitations} \label{appendix_limitation}
\textbf{Scope of Application} In this study, we primarily focus on anomaly detection for texture-based defects, which are common in real-world applications such as industrial defect detection and medical image diagnosis. Specifically, while our experiments include one-class classification (semantic anomaly detection), our performance is more pronounced in texture-based anomaly detection. In semantic anomaly detection, normal samples and anomalous samples are semantically different.

\textbf{Clean Performance} This study aims to improve the adversarial detection performance of anomaly detection tasks. Despite significant improvements in adversarial detection, our clean performance lags behind existing state-of-the-art detection methods. The trade-off between clean and adversarial test performance is well-documented in the literature \cite{tsipras2018robustness,zhang2019theoretically,madry2017towards,schmidt2018adversarially,raghunathan2020understanding}. Our work is also subject to such trade-offs. However, we have also provided results for scenarios where adversarial training is not performed.





\end{document}